%% file: main.tex
\documentclass{article}

\usepackage[main,final]{neurips_2025}
\input{header}
\usepackage{bbm}
\usepackage{graphicx}
\usepackage{subcaption}
\usepackage{booktabs}
\usepackage[utf8]{inputenc} 
\usepackage[T1]{fontenc}    
\usepackage{hyperref}       
\usepackage{url}            
\usepackage{amsfonts}       
\usepackage{nicefrac}       
\usepackage{microtype}      
\usepackage{xcolor}         

\usepackage{multirow}
\usepackage{algorithm}
\usepackage{algpseudocode}
\usepackage{subcaption}

\usepackage{xcolor}
\usepackage{bm}
\definecolor{mycyan}{RGB}{0,204,204}
\newcommand{\DD}[1]{{\color{mycyan}[DD: #1]}}
\definecolor{myorange}{RGB}{252, 105, 16}
\definecolor{myblue}{RGB}{32, 101, 164}

\usepackage{mwe}
\usepackage{booktabs}
\usepackage{array}

\usepackage{color}
\usepackage{colorspace}
\usepackage{caption}
\usepackage{wrapfig}
\definecolor{mutedred}  {RGB}{129,62,62}
\definecolor{mutedgreen}{RGB}{110,122,115}
\definecolor{sectioncolor}{rgb}{0.309803922, 0.443137255, 0.745098039} 
\definecolor{footcolor}   {rgb}{0.309803922, 0.443137255, 0.745098039} 

\definecolor{mygreen}       {rgb} {0.10,0.50,0.10}
\definecolor{mylightgreen}  {RGB} {50,168,82}
\definecolor{lightred}      {rgb} {1.00,0.80,0.80}
\definecolor{lightblue}     {rgb} {0.80,0.80,1.00}
\definecolor{mediumred}     {rgb} {1.00,0.60,0.60}
\definecolor{mediumblue}    {rgb} {0.60,0.60,1.00}

\definecolor{pennblue}{cmyk}{1,0.65,0,0.30}
\definecolor{pennred}{cmyk}{0,1,0.65,0.34}

\definecolor{myred}{RGB}{199,107,102}

\input{mysymbol.sty}

\title{Composition and Alignment of Diffusion Models using Constrained Learning}

\newcommand{\SK}[1]{{\color{myred}[SK: #1]}}

\newcommand{\IH}[1]{{\color{blue}[IH: #1]}}

\author{%
  Shervin~Khalafi\thanks{Corresponding authors.} \\
  University of Pennsylvania\\
  \texttt{shervink@seas.upenn.edu} \\
  \And
  Ignacio~Hounie \\
  University of Pennsylvania \\
  \texttt{ihounie@seas.upenn.edu} \\
  \And
  Dongsheng~Ding\footnotemark[\value{footnote}] \\
  University of Tennessee, Knoxville \\
  \texttt{dongshed@utk.edu}\\
  \And
  Alejandro~Ribeiro \\
  University of Pennsylvania \\
  \texttt{aribeiro@seas.upenn.edu}\\
}

\begin{document}

\maketitle

\begin{abstract}
   Diffusion models have become prevalent in generative modeling due to their ability to sample from complex distributions. To improve the quality of generated samples and their compliance with user requirements, two commonly used methods are: (i) Alignment, which involves finetuning a diffusion model to align it with a reward; and (ii) Composition, which combines several pretrained diffusion models together, each emphasizing a desirable attribute in the generated outputs. However, trade-offs often arise when optimizing for multiple rewards or combining multiple models, as they can often represent competing properties. Existing methods cannot guarantee that the resulting model faithfully generates samples with all the desired properties. To address this gap, we propose a constrained optimization framework that unifies alignment and composition of diffusion models by enforcing that the aligned model satisfies reward constraints and/or remains close to each pretrained model. We provide a theoretical characterization of the solutions to the constrained alignment and composition problems and develop a Lagrangian-based primal-dual training algorithm to approximate these solutions. Empirically, we demonstrate our proposed approach in image generation, applying it to alignment and composition, and show that our aligned or composed model satisfies constraints effectively. Our implementation can be found at: \href{https://github.com/shervinkhalafi/constrained_comp_align}{https://github.com/shervinkhalafi/constrained\_comp\_align}
\end{abstract}

\input{sec1_introduction}

\input{sec2_constrained_opt_alignment_composition}
\input{sec3_alignment}

\input{sec4_composition}

\input{sec5_experiments}

\input{sec6_conclusion}
\newpage
\bibliography{dd-bib}
\bibliographystyle{abbrv} %

\newpage
\section*{NeurIPS Paper Checklist}

\begin{enumerate}

\item {\bf Claims}
    \item[] Question: Do the main claims made in the abstract and introduction accurately reflect the paper's contributions and scope?
    \item[] Answer: \answerYes{} 
    \item[] Justification: There are references in the introduction to sections of the paper were we dicuss in depth the claims made.
    \item[] Guidelines:
    \begin{itemize}
        \item The answer NA means that the abstract and introduction do not include the claims made in the paper.
        \item The abstract and/or introduction should clearly state the claims made, including the contributions made in the paper and important assumptions and limitations. A No or NA answer to this question will not be perceived well by the reviewers. 
        \item The claims made should match theoretical and experimental results, and reflect how much the results can be expected to generalize to other settings. 
        \item It is fine to include aspirational goals as motivation as long as it is clear that these goals are not attained by the paper. 
    \end{itemize}

\item {\bf Limitations}
    \item[] Question: Does the paper discuss the limitations of the work performed by the authors?
    \item[] Answer: \answerYes{} 
    \item[] Justification: We discuss the limitations briefly in the conclusion, and more thoroughly in Appendix~\ref{app: limitations}.
    \item[] Guidelines:
    \begin{itemize}
        \item The answer NA means that the paper has no limitation while the answer No means that the paper has limitations, but those are not discussed in the paper. 
        \item The authors are encouraged to create a separate "Limitations" section in their paper.
        \item The paper should point out any strong assumptions and how robust the results are to violations of these assumptions (e.g., independence assumptions, noiseless settings, model well-specification, asymptotic approximations only holding locally). The authors should reflect on how these assumptions might be violated in practice and what the implications would be.
        \item The authors should reflect on the scope of the claims made, e.g., if the approach was only tested on a few datasets or with a few runs. In general, empirical results often depend on implicit assumptions, which should be articulated.
        \item The authors should reflect on the factors that influence the performance of the approach. For example, a facial recognition algorithm may perform poorly when image resolution is low or images are taken in low lighting. Or a speech-to-text system might not be used reliably to provide closed captions for online lectures because it fails to handle technical jargon.
        \item The authors should discuss the computational efficiency of the proposed algorithms and how they scale with dataset size.
        \item If applicable, the authors should discuss possible limitations of their approach to address problems of privacy and fairness.
        \item While the authors might fear that complete honesty about limitations might be used by reviewers as grounds for rejection, a worse outcome might be that reviewers discover limitations that aren't acknowledged in the paper. The authors should use their best judgment and recognize that individual actions in favor of transparency play an important role in developing norms that preserve the integrity of the community. Reviewers will be specifically instructed to not penalize honesty concerning limitations.
    \end{itemize}

\item {\bf Theory assumptions and proofs}
    \item[] Question: For each theoretical result, does the paper provide the full set of assumptions and a complete (and correct) proof?
    \item[] Answer: \answerYes{} 
    \item[] Justification: Yes, the full proofs for the theoretical results are provided in Appendix~\ref{app: proofs}.
    \item[] Guidelines:
    \begin{itemize}
        \item The answer NA means that the paper does not include theoretical results. 
        \item All the theorems, formulas, and proofs in the paper should be numbered and cross-referenced.
        \item All assumptions should be clearly stated or referenced in the statement of any theorems.
        \item The proofs can either appear in the main paper or the supplemental material, but if they appear in the supplemental material, the authors are encouraged to provide a short proof sketch to provide intuition. 
        \item Inversely, any informal proof provided in the core of the paper should be complemented by formal proofs provided in appendix or supplemental material.
        \item Theorems and Lemmas that the proof relies upon should be properly referenced. 
    \end{itemize}

    \item {\bf Experimental result reproducibility}
    \item[] Question: Does the paper fully disclose all the information needed to reproduce the main experimental results of the paper to the extent that it affects the main claims and/or conclusions of the paper (regardless of whether the code and data are provided or not)?
    \item[] Answer: \answerYes{} 
    \item[] Justification: We provide details for reproducing the experiments in the paper in Appendix~\ref{app: experiments}. We will also provide the code used for all of the experiments upon the paper's publication.
    \item[] Guidelines:
    \begin{itemize}
        \item The answer NA means that the paper does not include experiments.
        \item If the paper includes experiments, a No answer to this question will not be perceived well by the reviewers: Making the paper reproducible is important, regardless of whether the code and data are provided or not.
        \item If the contribution is a dataset and/or model, the authors should describe the steps taken to make their results reproducible or verifiable. 
        \item Depending on the contribution, reproducibility can be accomplished in various ways. For example, if the contribution is a novel architecture, describing the architecture fully might suffice, or if the contribution is a specific model and empirical evaluation, it may be necessary to either make it possible for others to replicate the model with the same dataset, or provide access to the model. In general. releasing code and data is often one good way to accomplish this, but reproducibility can also be provided via detailed instructions for how to replicate the results, access to a hosted model (e.g., in the case of a large language model), releasing of a model checkpoint, or other means that are appropriate to the research performed.
        \item While NeurIPS does not require releasing code, the conference does require all submissions to provide some reasonable avenue for reproducibility, which may depend on the nature of the contribution. For example
        \begin{enumerate}
            \item If the contribution is primarily a new algorithm, the paper should make it clear how to reproduce that algorithm.
            \item If the contribution is primarily a new model architecture, the paper should describe the architecture clearly and fully.
            \item If the contribution is a new model (e.g., a large language model), then there should either be a way to access this model for reproducing the results or a way to reproduce the model (e.g., with an open-source dataset or instructions for how to construct the dataset).
            \item We recognize that reproducibility may be tricky in some cases, in which case authors are welcome to describe the particular way they provide for reproducibility. In the case of closed-source models, it may be that access to the model is limited in some way (e.g., to registered users), but it should be possible for other researchers to have some path to reproducing or verifying the results.
        \end{enumerate}
    \end{itemize}

\item {\bf Open access to data and code}
    \item[] Question: Does the paper provide open access to the data and code, with sufficient instructions to faithfully reproduce the main experimental results, as described in supplemental material?
    \item[] Answer: \answerYes{} 
    \item[] Justification:  We will make public the repository with our code and implementations used for all of the experiments upon the paper's publication and include a link to it in the paper. Instructions and implementation details are provided in Appendix~\ref{app: experiments}. Anonymized code for implementing some of the experiments will also be provided with the supplementary material.
    \item[] Guidelines:
    \begin{itemize}
        \item The answer NA means that paper does not include experiments requiring code.
        \item Please see the NeurIPS code and data submission guidelines (\url{https://nips.cc/public/guides/CodeSubmissionPolicy}) for more details.
        \item While we encourage the release of code and data, we understand that this might not be possible, so “No” is an acceptable answer. Papers cannot be rejected simply for not including code, unless this is central to the contribution (e.g., for a new open-source benchmark).
        \item The instructions should contain the exact command and environment needed to run to reproduce the results. See the NeurIPS code and data submission guidelines (\url{https://nips.cc/public/guides/CodeSubmissionPolicy}) for more details.
        \item The authors should provide instructions on data access and preparation, including how to access the raw data, preprocessed data, intermediate data, and generated data, etc.
        \item The authors should provide scripts to reproduce all experimental results for the new proposed method and baselines. If only a subset of experiments are reproducible, they should state which ones are omitted from the script and why.
        \item At submission time, to preserve anonymity, the authors should release anonymized versions (if applicable).
        \item Providing as much information as possible in supplemental material (appended to the paper) is recommended, but including URLs to data and code is permitted.
    \end{itemize}

\item {\bf Experimental setting/details}
    \item[] Question: Does the paper specify all the training and test details (e.g., data splits, hyperparameters, how they were chosen, type of optimizer, etc.) necessary to understand the results?
    \item[] Answer: \answerYes{} 
    \item[] Justification: These details are provided in Appendix~\ref{app: experiments}.
    \item[] Guidelines:
    \begin{itemize}
        \item The answer NA means that the paper does not include experiments.
        \item The experimental setting should be presented in the core of the paper to a level of detail that is necessary to appreciate the results and make sense of them.
        \item The full details can be provided either with the code, in appendix, or as supplemental material.
    \end{itemize}

\item {\bf Experiment statistical significance}
    \item[] Question: Does the paper report error bars suitably and correctly defined or other appropriate information about the statistical significance of the experiments?
    \item[] Answer: \answerYes{} 
    \item[] Justification: Most of the plots in the main paper include error bars. Some don't for visual clarity. More details on statistical significance of the results are provided in Appendix~\ref{app: experiments}.
    \item[] Guidelines:
    \begin{itemize}
        \item The answer NA means that the paper does not include experiments.
        \item The authors should answer "Yes" if the results are accompanied by error bars, confidence intervals, or statistical significance tests, at least for the experiments that support the main claims of the paper.
        \item The factors of variability that the error bars are capturing should be clearly stated (for example, train/test split, initialization, random drawing of some parameter, or overall run with given experimental conditions).
        \item The method for calculating the error bars should be explained (closed form formula, call to a library function, bootstrap, etc.)
        \item The assumptions made should be given (e.g., Normally distributed errors).
        \item It should be clear whether the error bar is the standard deviation or the standard error of the mean.
        \item It is OK to report 1-sigma error bars, but one should state it. The authors should preferably report a 2-sigma error bar than state that they have a 96\% CI, if the hypothesis of Normality of errors is not verified.
        \item For asymmetric distributions, the authors should be careful not to show in tables or figures symmetric error bars that would yield results that are out of range (e.g. negative error rates).
        \item If error bars are reported in tables or plots, The authors should explain in the text how they were calculated and reference the corresponding figures or tables in the text.
    \end{itemize}

\item {\bf Experiments compute resources}
    \item[] Question: For each experiment, does the paper provide sufficient information on the computer resources (type of compute workers, memory, time of execution) needed to reproduce the experiments?
    \item[] Answer: \answerYes{} 
    \item[] Justification: We ran the experiments on a system with 2 NVIDIA RTX A6000 GPUs with 48 GB of GPU memory each. More details can be found in Appendix~\ref{app: experiments}.
    \item[] Guidelines:
    \begin{itemize}
        \item The answer NA means that the paper does not include experiments.
        \item The paper should indicate the type of compute workers CPU or GPU, internal cluster, or cloud provider, including relevant memory and storage.
        \item The paper should provide the amount of compute required for each of the individual experimental runs as well as estimate the total compute. 
        \item The paper should disclose whether the full research project required more compute than the experiments reported in the paper (e.g., preliminary or failed experiments that didn't make it into the paper). 
    \end{itemize}
    
\item {\bf Code of ethics}
    \item[] Question: Does the research conducted in the paper conform, in every respect, with the NeurIPS Code of Ethics \url{https://neurips.cc/public/EthicsGuidelines}?
    \item[] Answer: \answerYes{} 
    \item[] Justification: Yes.
    \item[] Guidelines:
    \begin{itemize}
        \item The answer NA means that the authors have not reviewed the NeurIPS Code of Ethics.
        \item If the authors answer No, they should explain the special circumstances that require a deviation from the Code of Ethics.
        \item The authors should make sure to preserve anonymity (e.g., if there is a special consideration due to laws or regulations in their jurisdiction).
    \end{itemize}

\item {\bf Broader impacts}
    \item[] Question: Does the paper discuss both potential positive societal impacts and negative societal impacts of the work performed?
    \item[] Answer: \answerYes{} 
    \item[] Justification: These impacts are discussed in Appendix~\ref{app: limitations}.
    \item[] Guidelines:
    \begin{itemize}
        \item The answer NA means that there is no societal impact of the work performed.
        \item If the authors answer NA or No, they should explain why their work has no societal impact or why the paper does not address societal impact.
        \item Examples of negative societal impacts include potential malicious or unintended uses (e.g., disinformation, generating fake profiles, surveillance), fairness considerations (e.g., deployment of technologies that could make decisions that unfairly impact specific groups), privacy considerations, and security considerations.
        \item The conference expects that many papers will be foundational research and not tied to particular applications, let alone deployments. However, if there is a direct path to any negative applications, the authors should point it out. For example, it is legitimate to point out that an improvement in the quality of generative models could be used to generate deepfakes for disinformation. On the other hand, it is not needed to point out that a generic algorithm for optimizing neural networks could enable people to train models that generate Deepfakes faster.
        \item The authors should consider possible harms that could arise when the technology is being used as intended and functioning correctly, harms that could arise when the technology is being used as intended but gives incorrect results, and harms following from (intentional or unintentional) misuse of the technology.
        \item If there are negative societal impacts, the authors could also discuss possible mitigation strategies (e.g., gated release of models, providing defenses in addition to attacks, mechanisms for monitoring misuse, mechanisms to monitor how a system learns from feedback over time, improving the efficiency and accessibility of ML).
    \end{itemize}
    
\item {\bf Safeguards}
    \item[] Question: Does the paper describe safeguards that have been put in place for responsible release of data or models that have a high risk for misuse (e.g., pretrained language models, image generators, or scraped datasets)?
    \item[] Answer: \answerNA{} 
    \item[] Justification: We use publicly available pretrained models.
    \item[] Guidelines:
    \begin{itemize}
        \item The answer NA means that the paper poses no such risks.
        \item Released models that have a high risk for misuse or dual-use should be released with necessary safeguards to allow for controlled use of the model, for example by requiring that users adhere to usage guidelines or restrictions to access the model or implementing safety filters. 
        \item Datasets that have been scraped from the Internet could pose safety risks. The authors should describe how they avoided releasing unsafe images.
        \item We recognize that providing effective safeguards is challenging, and many papers do not require this, but we encourage authors to take this into account and make a best faith effort.
    \end{itemize}

\item {\bf Licenses for existing assets}
    \item[] Question: Are the creators or original owners of assets (e.g., code, data, models), used in the paper, properly credited and are the license and terms of use explicitly mentioned and properly respected?
    \item[] Answer: \answerYes{} 
    \item[] Justification: The models and code used have been properly cited.
    \item[] Guidelines:
    \begin{itemize}
        \item The answer NA means that the paper does not use existing assets.
        \item The authors should cite the original paper that produced the code package or dataset.
        \item The authors should state which version of the asset is used and, if possible, include a URL.
        \item The name of the license (e.g., CC-BY 4.0) should be included for each asset.
        \item For scraped data from a particular source (e.g., website), the copyright and terms of service of that source should be provided.
        \item If assets are released, the license, copyright information, and terms of use in the package should be provided. For popular datasets, \url{paperswithcode.com/datasets} has curated licenses for some datasets. Their licensing guide can help determine the license of a dataset.
        \item For existing datasets that are re-packaged, both the original license and the license of the derived asset (if it has changed) should be provided.
        \item If this information is not available online, the authors are encouraged to reach out to the asset's creators.
    \end{itemize}

\item {\bf New assets}
    \item[] Question: Are new assets introduced in the paper well documented and is the documentation provided alongside the assets?
    \item[] Answer: \answerNA{} 
    \item[] Justification: -
    \item[] Guidelines:
    \begin{itemize}
        \item The answer NA means that the paper does not release new assets.
        \item Researchers should communicate the details of the dataset/code/model as part of their submissions via structured templates. This includes details about training, license, limitations, etc. 
        \item The paper should discuss whether and how consent was obtained from people whose asset is used.
        \item At submission time, remember to anonymize your assets (if applicable). You can either create an anonymized URL or include an anonymized zip file.
    \end{itemize}

\item {\bf Crowdsourcing and research with human subjects}
    \item[] Question: For crowdsourcing experiments and research with human subjects, does the paper include the full text of instructions given to participants and screenshots, if applicable, as well as details about compensation (if any)? 
    \item[] Answer: \answerNA{} 
    \item[] Justification: -
    \item[] Guidelines:
    \begin{itemize}
        \item The answer NA means that the paper does not involve crowdsourcing nor research with human subjects.
        \item Including this information in the supplemental material is fine, but if the main contribution of the paper involves human subjects, then as much detail as possible should be included in the main paper. 
        \item According to the NeurIPS Code of Ethics, workers involved in data collection, curation, or other labor should be paid at least the minimum wage in the country of the data collector. 
    \end{itemize}

\item {\bf Institutional review board (IRB) approvals or equivalent for research with human subjects}
    \item[] Question: Does the paper describe potential risks incurred by study participants, whether such risks were disclosed to the subjects, and whether Institutional Review Board (IRB) approvals (or an equivalent approval/review based on the requirements of your country or institution) were obtained?
    \item[] Answer: \answerNA{} 
    \item[] Justification: -
    \item[] Guidelines:
    \begin{itemize}
        \item The answer NA means that the paper does not involve crowdsourcing nor research with human subjects.
        \item Depending on the country in which research is conducted, IRB approval (or equivalent) may be required for any human subjects research. If you obtained IRB approval, you should clearly state this in the paper. 
        \item We recognize that the procedures for this may vary significantly between institutions and locations, and we expect authors to adhere to the NeurIPS Code of Ethics and the guidelines for their institution. 
        \item For initial submissions, do not include any information that would break anonymity (if applicable), such as the institution conducting the review.
    \end{itemize}

\item {\bf Declaration of LLM usage}
    \item[] Question: Does the paper describe the usage of LLMs if it is an important, original, or non-standard component of the core methods in this research? Note that if the LLM is used only for writing, editing, or formatting purposes and does not impact the core methodology, scientific rigorousness, or originality of the research, declaration is not required.
    \item[] Answer: \answerNA{} 
    \item[] Justification: -
    \item[] Guidelines:
    \begin{itemize}
        \item The answer NA means that the core method development in this research does not involve LLMs as any important, original, or non-standard components.
        \item Please refer to our LLM policy (\url{https://neurips.cc/Conferences/2025/LLM}) for what should or should not be described.
    \end{itemize}

\end{enumerate}

\newpage
\clearpage

~\\
\centerline{{\fontsize{14}{14}\selectfont \textbf{Supplementary Materials for }}}

\vspace{6pt}
\centerline{\fontsize{13.5}{13.5}\selectfont \textbf{
	``Composition and Alignment of Diffusion Models using}}

\vspace{6pt}
\centerline{\fontsize{13.5}{13.5}\selectfont \textbf{
	 Constrained Learning''}}

\vspace{6pt}

\appendix
\input{app1_proofs}

\input{app2_mixture_with_rewards}
\input{app4_mixture_algorithm}

\input{app5_negation}
\input{app6_regularization_correction}
\input{app7_continuous_time_analysis}
\input{app8_algorithms}


\end{document}

%% file: header.tex
\usepackage{amsthm}
\usepackage{amsmath}
\usepackage{amssymb}
\usepackage{graphicx}
\usepackage{mathtools}
\usepackage{enumerate}
\usepackage{enumitem}
\usepackage{footnote}
\usepackage{float}
\usepackage{xspace}
\usepackage{multirow}
\usepackage{nicefrac}
\usepackage{wrapfig}
\usepackage{framed}
\usepackage{url}
\usepackage{units}
\usepackage[colorlinks=true, linkcolor=blue, citecolor=blue]{hyperref}
\usepackage{blkarray}
\PassOptionsToPackage{round}{natbib}

\usepackage{tikz}
\usetikzlibrary{arrows,chains,matrix,positioning,scopes}

\let\hat\widehat
\let\tilde\widetilde
\let\check\widecheck

\newtheorem{lemma}{Lemma}

\newtheorem{theorem}{Theorem}
\newtheorem{corollary}{Corollary}
\newtheorem{remark}{Remark}

\newtheorem{assumption}{Assumption}

\DeclareMathOperator*{\minimize}{minimize}
\DeclareMathOperator*{\maximize}{maximize}
\DeclareMathOperator*{\subject}{subject~to}
\newcommand{\DefinedAs}[0]{\mathrel{\mathop:}=}

\newcommand{\one}{\boldsymbol{1}}

\DeclareMathOperator*{\argmin}{argmin}
\DeclareMathOperator*{\argmax}{argmax}

\newcommand{\norm}[1]{\left\|{#1}\right\|}

\DeclareFontFamily{OMX}{MnSymbolE}{}
\DeclareFontShape{OMX}{MnSymbolE}{m}{n}{
    <-6>  MnSymbolE5
   <6-7>  MnSymbolE6
   <7-8>  MnSymbolE7
   <8-9>  MnSymbolE8
   <9-10> MnSymbolE9
  <10-12> MnSymbolE10
  <12->   MnSymbolE12}{}
\DeclareSymbolFont{mnlargesymbols}{OMX}{MnSymbolE}{m}{n}
\SetSymbolFont{mnlargesymbols}{bold}{OMX}{MnSymbolE}{b}{n}
\DeclareMathDelimiter{\llangle}{\mathopen}{mnlargesymbols}{'164}{mnlargesymbols}{'164}
\DeclareMathDelimiter{\rrangle}{\mathclose}{mnlargesymbols}{'171}{mnlargesymbols}{'171}

\usepackage{prettyref}

\newcommand{\savehyperref}[2]{\texorpdfstring{\hyperref[#1]{#2}}{#2}}
\newrefformat{eq}{\savehyperref{#1}{\textup{(\ref*{#1})}}}
\newrefformat{eqn}{\savehyperref{#1}{Equation~\eqref{#1}}}
\newrefformat{lem}{\savehyperref{#1}{Lemma~\ref*{#1}}}
\newrefformat{def}{\savehyperref{#1}{Definition~\ref*{#1}}}
\newrefformat{line}{\savehyperref{#1}{line~\ref*{#1}}}
\newrefformat{thm}{\savehyperref{#1}{Theorem~\ref*{#1}}}
\newrefformat{corr}{\savehyperref{#1}{Corollary~\ref*{#1}}}
\newrefformat{cor}{\savehyperref{#1}{Corollary~\ref*{#1}}}
\newrefformat{sec}{\savehyperref{#1}{Section~\ref*{#1}}}
\newrefformat{app}{\savehyperref{#1}{Appendix~\ref*{#1}}}
\newrefformat{ap}{\savehyperref{#1}{Appendix~\ref*{#1}}}
\newrefformat{assum}{\savehyperref{#1}{Assumption~\ref*{#1}}}
\newrefformat{as}{\savehyperref{#1}{Assumption~\ref*{#1}}}
\newrefformat{ex}{\savehyperref{#1}{Example~\ref*{#1}}}
\newrefformat{fig}{\savehyperref{#1}{Figure~\ref*{#1}}}
\newrefformat{alg}{\savehyperref{#1}{Algorithm~\ref*{#1}}}
\newrefformat{rem}{\savehyperref{#1}{Remark~\ref*{#1}}}
\newrefformat{conj}{\savehyperref{#1}{Conjecture~\ref*{#1}}}
\newrefformat{prop}{\savehyperref{#1}{Proposition~\ref*{#1}}}
\newrefformat{proto}{\savehyperref{#1}{Protocol~\ref*{#1}}}
\newrefformat{prob}{\savehyperref{#1}{Problem~\ref*{#1}}}
\newrefformat{claim}{\savehyperref{#1}{Claim~\ref*{#1}}}
\newrefformat{que}{\savehyperref{#1}{Question~\ref*{#1}}}

%% file: sec1_introduction.tex
    


    
    
    
    
    


    


\section{Introduction}

Diffusion models have emerged as the tool of choice for generative modeling in a variety of settings \cite{saharia2022photorealistic, blattmann2023align, wang2025diffusion, chi2023diffusion}, image generation being most prominent among them \cite{rombach2022highresolutionimagesynthesislatent}. Users of these diffusion models would like to adapt them to their  specific preferences, but this aspiration is hindered by the often enormous cost and complexity of their training~\cite{ulhaq2022efficient,yan2024diffusion}. For this reason, \emph{alignment} and \emph{composition} of what, in this context, become \emph{pretrained} models, has become popular~\cite{liu2024alignment,liu2022compositional}. 

Regardless of whether the goal is alignment or composition, we want to balance what are most likely conflicting requirements. In alignment, we want to stay close to the pretrained model while deviating sufficiently so as to affect some rewards of interest~\cite{fan2023dpokreinforcementlearningfinetuning, domingoenrich2025adjointmatchingfinetuningflow}. In composition, given several pretrained models, our goal is to sample from the union or intersection of their distributions~\cite{du2024reducereuserecyclecompositional, biggs2024diffusion}. The standard approach to balance these requirements involves the use of \emph{weighted averages}. This can be a linear combination of score functions in composition~\cite{du2024reducereuserecyclecompositional, biggs2024diffusion} or may involve a loss given by a linear combination of a Kullback-Leibler (KL) divergence and a reward~\cite{fan2023dpokreinforcementlearningfinetuning} in the case of alignment. In practice, weight-based methods are often designed in an ad hoc manner, with the weights treated as tunable hyperparameters, which makes the approach notoriously difficult to optimize and generalize.

In this work, we propose a unified view of alignment and composition via the lens of constrained learning~\cite{chamon2021probablyapproximatelycorrectconstrained, chamon2022constrained}. As their names indicate, constrained alignment and constrained composition problems balance conflicting requirements using \emph{constraints} instead of \emph{weights}. Learning with constraints and learning with weights are related problems -- indeed, we will train constrained diffusion models in their Lagrangian forms. Yet, they are also fundamentally different. In the constrained formulation, the hyperparameter tuning spaces are more interpretable (see Section~\ref{sec_alignment}), and in some cases -- such as the constrained composition formulation -- hyperparameter tuning can even be avoided entirely (see Section~\ref{sec: composition reverse KL}). These advantages are particularly evident in constrained problems, as discussed in Sections~\ref{sec_alignment} and~\ref{sec: composition reverse KL}. We summarize our key contributions in three aspects below. 

\begin{enumerate}
    \item[(i)] Problem Formulation 
    \begin{itemize}
        \item  For alignment, we formulate a reverse KL divergence-constrained distribution optimization problem that minimizes the reverse KL divergence to a pretrained model, subject to expected reward constraints with user-specified thresholds.
        
        \item For composition, we propose using KL divergence constraints to ensure the closeness to each pretrained model. It is important to distinguish composition with \emph{reverse} KL and \emph{forward} KL constraints as they lead to a weighted product or weighted mixture~\cite{khalafi2024constrained} of the individual distributions, respectively. In this work, we focus on composition with reverse KL constraints, and discuss forward KL constraints in Appendix~\ref{app: mixture comp}. 
    \end{itemize}
    \item[(ii)] Theoretical Analysis 
    \begin{itemize}
        \item In Section~\ref{sec_alignment}, we characterize the solution of the alignment problem as the pretrained model distribution scaled by an exponential function of a weighted sum of reward functions. In Section~\ref{sec: composition reverse KL}, we characterize the solution of the constrained optimization problem with \emph{reverse} KL divergence constraints as a tilted product of the individual distributions. We establish strong duality for both problems, which enables us to use a dual-based approach to develop primal-dual training algorithms for solving them.
        \item We illustrate the distinction between the KL divergence between diffusion trajectories (path-wise), and the KL divergence between the final distributions (point-wise) in Section~\ref{subsec: diffusion models}. We also propose a new method to evaluate the point-wise KL divergence.
    \end{itemize}
    \item[(iii)] Empirical Results 
    \begin{itemize}
        \item For alignment, we demonstrate the difference between constrained and weighted alignment through experiments in Section~\ref{subsec: alignment experiments}. The constrained approach scales naturally to finetuning with multiple rewards, eliminating the need for extensive hyperparameter searches to determine suitable weights. Moreover, specifying reward thresholds is often more intuitive than choosing regularization weights. Without constraints, however, the model can easily overfit to one or several rewards and diverge substantially from the pretrained model. In contrast, our method identifies the model closest to the pretrained one that still satisfies the desired reward constraints (see Figure~\ref{fig: all rewards alignment}).

        \item For composition, we show the properties of constrained composition of diffusion models through experiments in Section~\ref{subsec: composition experiments}. We see that when the composition weights are not chosen properly, the resulting model can become biased towards certain individual models while neglecting others. Constrained composition addresses this issue by finding optimal weights that preserve closeness to each individual model. Particularly, when composing multiple text-to-image models each finetuned on a different reward, imposing constraints yields weights that enable the composed model to achieve higher performance across all rewards, compared to composition with equal weights.
    \end{itemize}
\end{enumerate}

%% file: sec2_constrained_opt_alignment_composition.tex
\section{Composition and Alignment of Diffusion Models}\label{sec_composition_and_alignement}

We introduce constrained distribution problems for alignment and composition in Section~\ref{subsec_composition_and_alignement}, and characterize the reverse and forward KL divergences for diffusion models in Section~\ref{subsec: diffusion models}. 

\subsection{Composition and alignment in distribution space}\label{subsec_composition_and_alignement}

We formulate~{U}nparameterized constrained distribution optimization problems using {R}everse or {F}orward KL divergence, for {A}lignment and {C}omposition as illustrated in~\eqref{eq: constrained alignment reverse KL},~\eqref{eq: constrained composition reverse KL}, and~\eqref{eq: constrained composition forward KL}. 

\textbf{Reward alignment:} Given a pretrained model $q$ and a set of $m$ rewards $\{r_i(x)\}_{i\,=\,1}^m$ that can be evaluated on a sample $x$, we consider the \emph{reverse} KL divergence $D_{\text{KL}}(p\,\Vert\, q) \DefinedAs \int p(x) \log(p(x)/q(x))dx$ that measures the difference  between a distribution $p$ and the pretrained model $q$. Additionally, for each reward $r_i$, we define a constant $b_i$ standing for requirement for reward $r_i$. We formulate a constrained alignment problem that minimizes a reverse KL divergence subject to $m$ constraints:
\begin{align}\label{eq: constrained alignment reverse KL}\tag{UR-A}
    p^\star ~=~ 
    \argmin_{p} 
        ~ D_{\text{KL}} \big(\, p\,\Vert\, q \,\big) \qquad
    \subject      
        ~ \mathbb{E}_{x\,\sim\,p} \big[ \,r_i(x)\, \big]\; \geq \; b_i \,\text{ for}\; i = 1,\ldots,m.
\end{align}
As per \eqref{eq: constrained alignment reverse KL}, the constrained alignment problem is solved by the distribution $p^\star$ that is closest to the pretrained one $q$ as measured by the reverse KL divergence $D_{\text{KL}}(p\,\Vert\, q)$ among those whose expected rewards $\mathbb{E}_{x\,\sim\,p} [r_i(x)]$ accumulate to at least $b_i$. By `pretrained model' we refer to a sampling process that produces samples, not the underlying distribution. Let the primal value be $P_{\text{ALI}}^\star\DefinedAs D_{\text{KL}}(p^\star\,\Vert\,q)$.



\begin{figure}[t]
  \centering
  \begin{subfigure}[b]{0.3\textwidth}
    \centering
    \includegraphics[width=\linewidth]{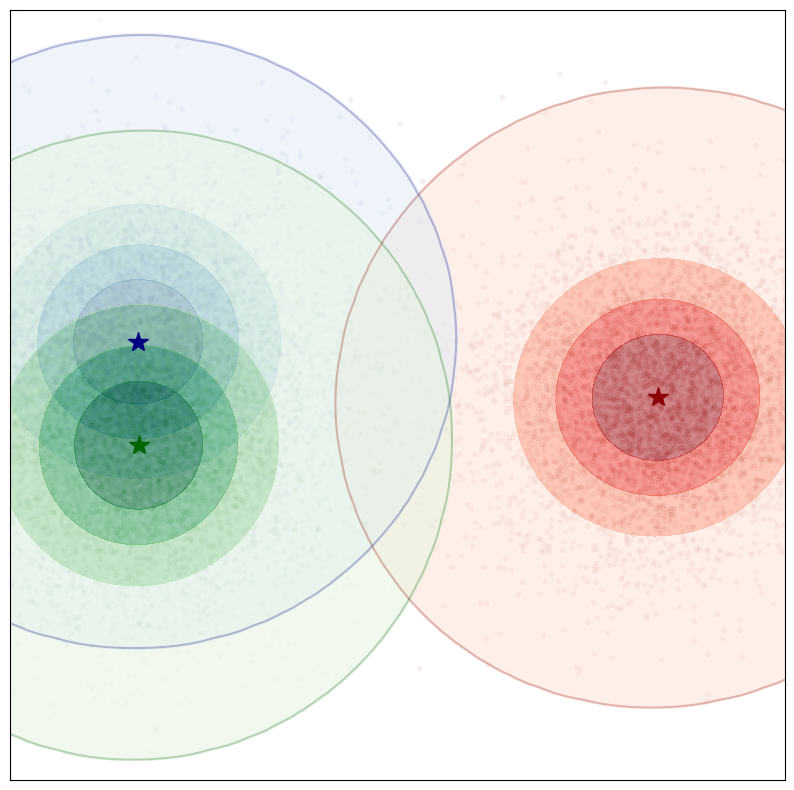}
    \label{fig:sub2}
  \end{subfigure}\hfill
  \begin{subfigure}[b]{0.3\textwidth}
    \centering
    \includegraphics[width=\linewidth]{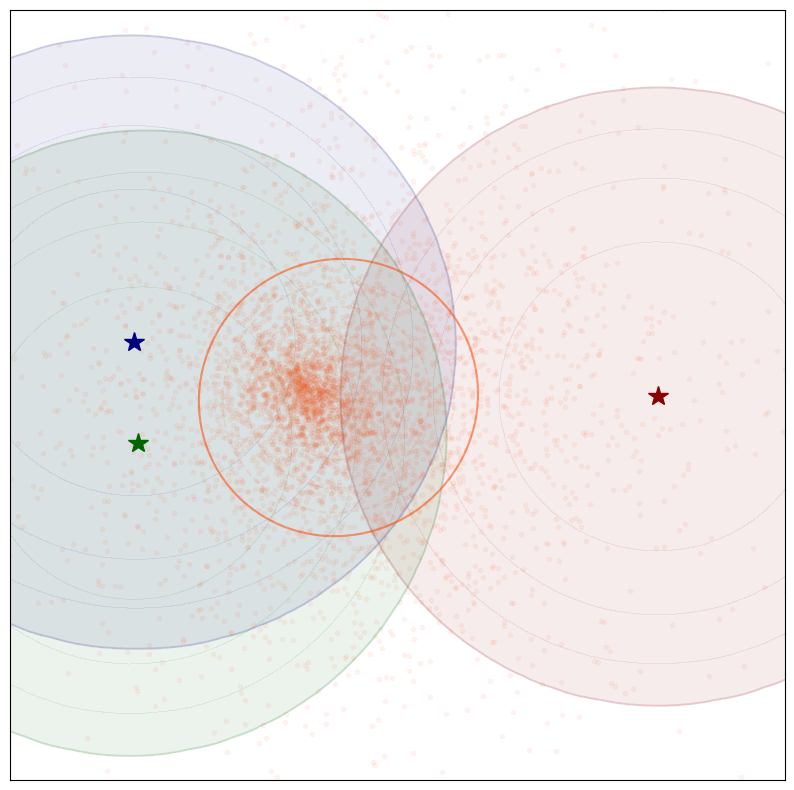}
    \label{fig:sub1}
  \end{subfigure}\hfill
  \begin{subfigure}[b]{0.3\textwidth}
    \centering
    \includegraphics[width=\linewidth]{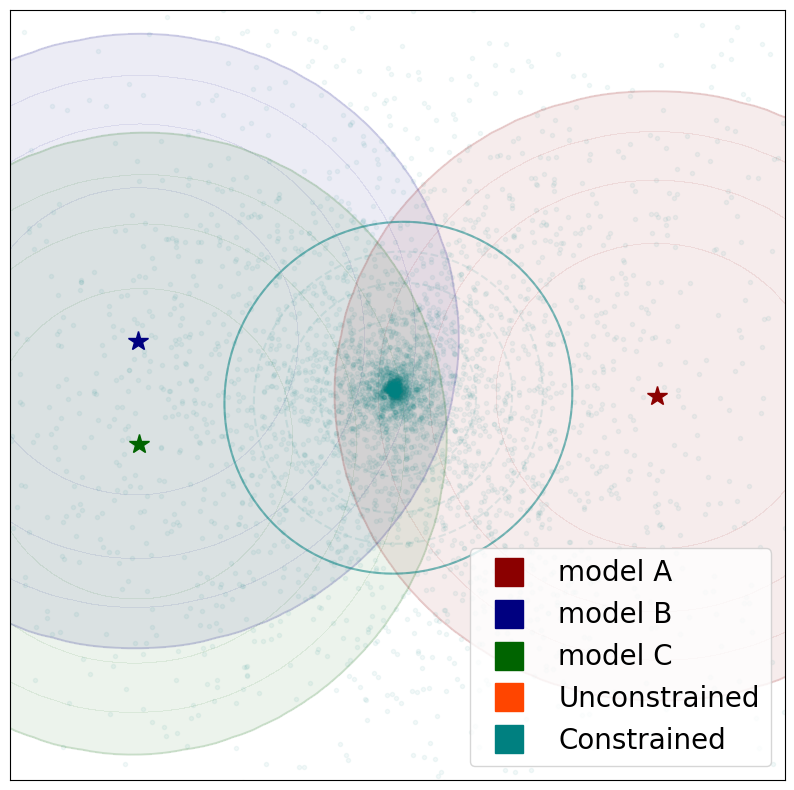}
    \label{fig:sub2}
  \end{subfigure}
  \caption{Product composition (AND). Three Gaussian distributions being composed (Left). Composition using equal weights (Middle), and with constraints (Right). The constrained model samples from the intersection of the three models.}
  \label{fig: gaussians AND}
\end{figure}

\textbf{Product composition (AND)}: Given a set of $m$ pretrained models $\{q^i\}_{i\,=\,1}^m$, we formulate a constrained composition problem that solves a reverse KL-constrained optimization problem:
\begin{align}\label{eq: constrained composition reverse KL}\tag{UR-C}
    (p^\star,u^\star) ~=~ 
    \argmin_{p,\, u} 
        ~ u \qquad
    \subject      
        ~ D_{\text{KL}} \big(\, p\,\Vert\, q^i \,\big) \leq u \,\text{ for}\; i = 1,\ldots,m.
\end{align}
In \eqref{eq: constrained composition reverse KL}, the decision variable $u$ serves as an upper bound on the $m$ KL divergences between a distribution $p$ and $m$ pretrained models $\{q^i\}_{i\,=\,1}^m$. Partial minimization over $u$ allows us to search for a distribution $p$ that minimizes this common upper bound. Hence, the optimal solution $p^\star$ minimizes the maximum KL divergence among $m$ terms, each computed between $p$ and a pretrained model $q_i$. Let the primal value be $P_{\text{AND}}^\star\DefinedAs u^\star$. The epigraph formulation~\eqref{eq: constrained composition reverse KL} is practical, as the constraint threshold $u$ can be updated dynamically during training. In contrast, Figure~\ref{fig: gaussians AND} shows that the model composed with equal weights is biased toward the two most similar distributions.

\begin{figure}[t]
  \centering
  \begin{subfigure}[b]{0.3\textwidth}
    \centering
    \includegraphics[width=\linewidth]{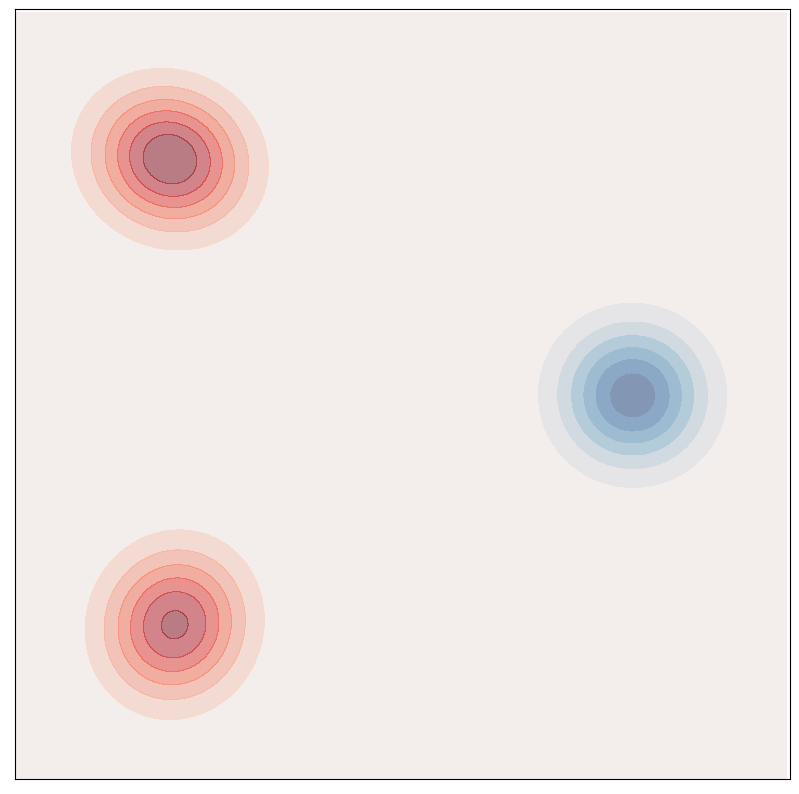}
    \label{fig:sub1}
  \end{subfigure}\hfill
  \begin{subfigure}[b]{0.3\textwidth}
    \centering
    \includegraphics[width=\linewidth]{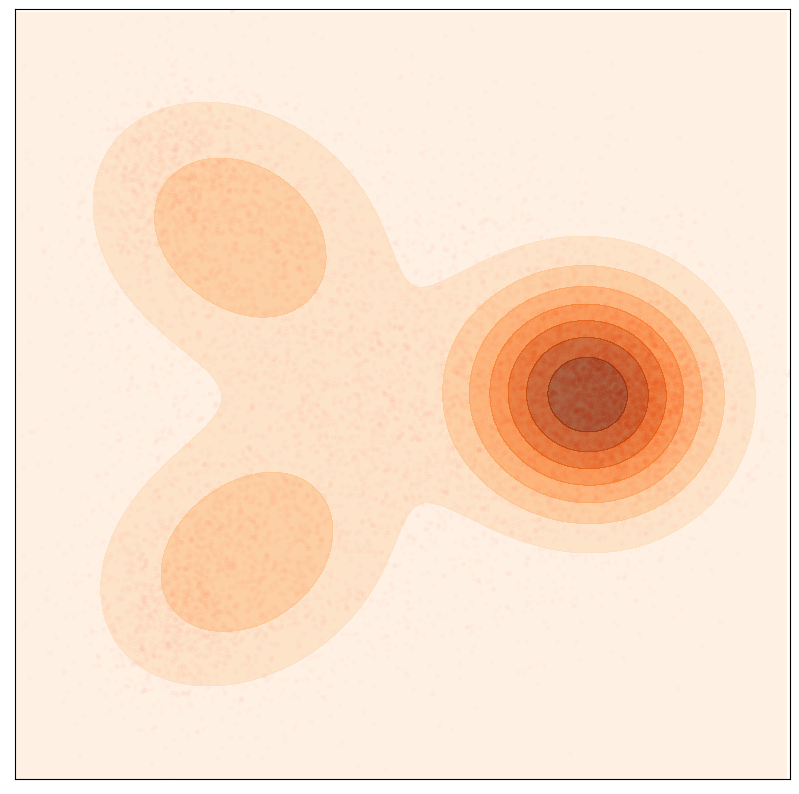}
    \label{fig:sub2}
  \end{subfigure}\hfill
  \begin{subfigure}[b]{0.3\textwidth}
    \centering
    \includegraphics[width=\linewidth]{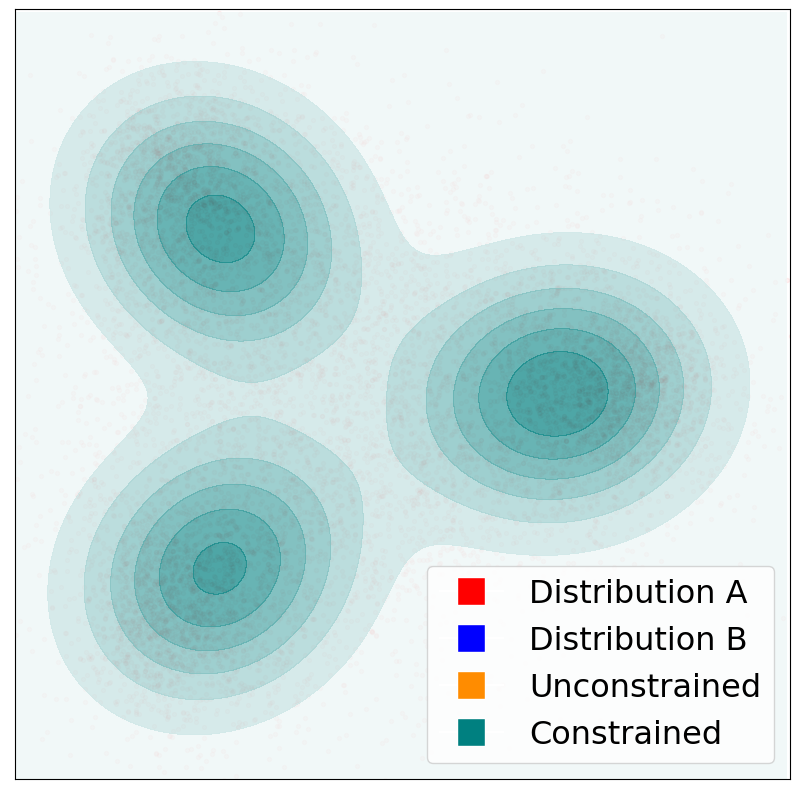}
    \label{fig:sub2}
  \end{subfigure}
  \caption{Mixture composition (OR). Two of Gaussian mixtures being composed (Left). One has two modes and the other has only a single mode. Composition using equal weights (Middle), and with constraints (Right). }
  \label{fig:gaussians OR}
\end{figure}

\textbf{Mixture composition (OR)}: A different composition modality that also fits within our constrained framework is the \emph{forward} KL-constrained composition problem. We obtain this formulation by replacing the \emph{reverse} divergence $D_{\text{KL}}(p\,\Vert\, q_i)$ in \eqref{eq: constrained composition reverse KL} with the \emph{forward} KL divergence $D_{\text{KL}}(q_i\,\Vert\, p)$:
\begin{align}\label{eq: constrained composition forward KL}\tag{UF-C}
    (p^\star,u^\star) ~=~ 
    \argmin_{p,\, u} 
        ~ u \qquad
    \subject      
        ~ D_{\text{KL}} \big(\, q_i\,\Vert\, p \,\big) \leq u \,\text{ for}\; i = 1,\ldots,m.
\end{align}
Mixture composition was studied in a related but slightly different constrained setting~\cite{khalafi2024constrained}. In fact, the solution of the constrained problem~\eqref{eq: constrained composition forward KL} learns to sample from each distribution in proportion to its entropy; see~\cite[Theorem~2]{khalafi2024constrained}. As shown in Figure~\ref{fig:gaussians OR}, the constrained model samples more frequently from the higher-entropy distribution with two models, whereas the equally weighted composition samples equally from both distributions, leading to unbalanced sampling. Since the algorithmic design and analysis for~\eqref{eq: constrained composition forward KL} follow those in~\cite{khalafi2024constrained}, mixture composition is not the main focus of this work. For completeness, we compare it with product composition in Appendix~\ref{app: mixture comp}. 


The reverse KL-based composition~\eqref{eq: constrained composition reverse KL} tends to sample from the intersection of the pretrained models $\{q_i\}_{i\,=\,1}^m$, whereas the forward KL-based composition~\eqref{eq: constrained composition forward KL} tends to sample from their union. Thus, product composition enforces a conjunction (logical AND) across pretrained models, while mixture composition corresponds to a disjunction (logical OR). We emphasize that Problems~\eqref{eq: constrained alignment reverse KL}, \eqref{eq: constrained composition reverse KL}, and \eqref{eq: constrained composition forward KL} should serve as canonical formulations; the proposed constrained methods can be readily adapted to their variants, e.g., mixture composition with reward constraints.

\subsection{KL divergence for diffusion models}\label{subsec: diffusion models}
A generative diffusion model consists of forward and backward processes. In the forward process, we add Gaussian noise $\epsilon_t$ to a clean sample $\bar X_0 \sim \bar p_0$ over $T$ time steps as follows
\begin{equation}\label{eq: forward process}
    \bar X_t 
    \; = \;
    \frac{\alpha_t}{\alpha_{t - 1}}\, \bar X_{t - 1} 
    \, + \,
    \sqrt{1 - \frac{\alpha_t}{\alpha_{t - 1}}}\, \epsilon_t, \; \text{ for } t = 1, \cdots, T
\end{equation}
where $\epsilon_t \sim \mathcal{N}(0, I)$ is the standard Gaussian noise, and $\{ \alpha_{t} \}_{t\,=\,1}^T$ is a decreasing sequence of coefficients called the noise schedule. We denote the marginal density of $\bar X_t$ at time $t$ as $\bar p_t(\cdot)$. Given a $d$-dimensional score predictor function $s(x,t)$: $\mathbb{R}^d\times \{1, \cdots,T\} \rightarrow \mathbb{R}^d$, we introduce a backward denoising diffusion implicit model (DDIM) process~\cite{song2022denoisingdiffusionimplicitmodels} as follows
\begin{equation}
X_{t-1} 
\; = \;
\sqrt{\frac{\alpha_{t-1}}{{\alpha}_t}} X_t 
\, + \,
\beta_t \, s(X_t, t)
\, + \,
\sigma_t \epsilon_t
\label{eq:ddim discrete time}
\end{equation}
where $\epsilon_t \sim \mathcal{N}(0, I)$ is the standard Gaussian noise, and $\{\sigma_t^2\}_{t\,=\,1}^T$ is the variance schedule that determines the level of randomness in the backward process (e.g., $\sigma_t =0$ reduces to deterministic trajectories), and $\beta_t \DefinedAs \sqrt{\frac{\alpha_{t-1}}{{\alpha}_t}} \sqrt{(1-{\alpha}_t) (1 - \bar \alpha_t)} - \sqrt{(1 - {\alpha}_{t-1} - \sigma_t^2)(1-\bar{\alpha}_t)}$ is determined by the variance schedule $\sigma_t$ and the noise schedule $\alpha_t$. Here, we use the equivalence between the score-matching and denoising formulations of diffusion model to replace the denoising predictor in~\cite{song2022denoisingdiffusionimplicitmodels} by the score function. Given a score function $s(x,t)$, we denote the marginal density of $X_t$ as $ p_t(\cdot\,; s)$ and the joint distribution over the entire process as $p_{0:T}(x_{0:T}; s)$.


In the score-matching formulation~\cite{song2021scorebasedgenerativemodelingstochastic}, a denoising score-matching objective is minimized to obtain a function $s^\star$ that approximates the true score function of the forward process, i.e., $s^\star(x,t) \approx \nabla \log \bar p_t(x)$. Then, the marginal densities of the backward process~\eqref{eq:ddim discrete time} match those of the forward process~\eqref{eq: forward process}, i.e., $p_t(\cdot\,; s^\star) = \bar p_t(\cdot)$ for all $t$. Thus we can run the backward process to generate samples {$x_0 \sim p_0$} that resemble samples from the original data distribution $\bar x_0 \sim \bar p_0$.

We denote the KL divergence between two joint distributions $p$, $q$ over the backward process by $D_{\text{\normalfont KL}} (p_{0:T}(\cdot) \,\Vert\, q_{0:T}(\cdot))$, which is known as path-wise KL~\cite{fan2023dpokreinforcementlearningfinetuning,han2024stochastic}. The path-wise KL divergence is often used in alignment to measure the difference between finetuned and pretrained models.
\begin{lemma}[Path-wise KL divergence]\label{lem: reverse KL computation}
    If two backward processes $p_{0:T}(\cdot)$ and $q_{0:T}(\cdot)$ have the same variance schedule $\sigma_t$ and noise schedule $\alpha_t$, then the reverse KL divergence between them is given by
\begin{equation}
    \label{eq:path KL}
    D_{\text{\normalfont KL}}(p_{0:T}(\cdot; s_p) \, \Vert\,  q_{0:T}(\cdot; s_q)) 
    \; = \;
    \sum_{t\,=\,1}^{T} \mathbb{E}_{x_t \,\sim\, p_t(\cdot; s_p)}\left[ 
    \,\frac{1}{2 \sigma_t^2} \Vert s_p (x_t, t) - s_q (x_t, t) \Vert^2
    \,\right].
\end{equation}
\end{lemma}
See Appendix~\ref{app: proofs_reverse kl diffusion} for the proof. When the two backward processes differ in their variance and noise schedules, the path-wise KL divergence remains tractable, and we omit it for simplicity. While the path-wise KL divergence is a useful regularizer for alignment, when composing multiple models, the point-wise KL  divergence $D_{\text{\normalfont KL}}(p_{0}(\cdot) \, \Vert\,  q_{0}(\cdot))$ is a more natural measure of the closeness between two diffusion models. This is because we mainly care about the closeness of the final sampling distributions: $p_0 (\cdot)$, $q_0(\cdot)$, and not the underlying processes: $p_{0:T} (\cdot)$, $q_{0:T}(\cdot)$. However, since our proposed approach to compute the point-wise KL is intractable for alignment, we adopt the path-wise KL for alignment and retain the point-wise KL for composition; see more discussion in Section~\ref{subsec:Product composition of diffusion models}.

However, it is not obvious how to compute the point-wise KL divergence, as evaluating the marginal densities is intractable. We next establish a similar formula as~\eqref{eq:path KL} by limiting the score function class.

\begin{lemma}[Point-wise KL divergence]\label{lem: marginal kl computation}
    Assume two score functions $s_p(x,t) = \nabla \log \bar p_t(x)$, $s_q (x,t) = \nabla \log \bar q_t(x)$, where $\bar p_t$, $\bar q_t$ are two marginal densities induced by two forward diffusion processes, with the same noise schedule, starting from initial distributions $\bar p_0$ and $\bar q_0$, respectively. Then, the point-wise KL divergence between two distributions of the samples generated by running DDIM with $s_p$ and $s_q$ is given by
    \begin{equation}\label{eq: marginal kl computation}
        D_{\text{\normalfont KL}} (p_0(\cdot\,; s_p) \,\Vert\, q_0(\cdot\,; s_q))
        \; = \;
        \sum_{t \,=\, 0}^T \tilde \omega_t \, \mathbb{E}_{x\,\sim\, p_t(\cdot\,; s_p)} \left[ \,
        \norm{s_p(x, t) - s_q(x, t)}_2^2
        \,\right] 
        \,+\, \epsilon_T
    \end{equation}
    where $\tilde \omega_t$ is a time-dependent constant, and $\epsilon_T$ is a discretization error that depends on the total number of diffusion time steps $T$.
\end{lemma} 
See Appendix~\ref{app: proof of lemma marginal kl} for proof. The key intuition behind Lemma~\ref{lem: marginal kl computation} is that if two diffusion processes are close, and their starting distributions are the same (e.g., $\mathcal{N}(0, I)$ at time $t = T$), then the end points (i.e., the distributions at $t = 0$) must also be close. The sum on the right-hand side of~\eqref{eq: marginal kl computation} can be viewed as the difference of the two processes over time steps, up to a discretization error. 

%% file: sec3_alignment.tex
\section{Aligning Pretrained Model with Multiple Reward Constraints}\label{sec_alignment}

We provide a characterization of the solution to Problem~\eqref{eq: constrained alignment reverse KL} in Section~\ref{subsec: Reward alignment in distribution space}, and establish strong duality for diffusion models in Section~\ref{subsec: Reward alignment for diffusion models}, together with a dual-based training algorithm.


\subsection{Reward alignment in distribution space}
\label{subsec: Reward alignment in distribution space}

To apply Problem~\eqref{eq: constrained alignment reverse KL} to diffusion models, we first employ Lagrangian duality to derive its solution in distribution space. Alignment with constraints is related but fundamentally different from the standard approach of minimizing a weighted average of the KL divergence and rewards~\cite{fan2023dpokreinforcementlearningfinetuning}. They are related because the Lagrangian for Problem~\eqref{eq: constrained alignment reverse KL} is precisely the weighted average:
\begin{align}\label{eqn_alignement_lagrangian}
    L_{\text{ALI}}(p ,\lambda) 
        ~ = ~ D_{\text{KL}} \big(\, p\,\Vert\, q \,\big)
        ~ - ~ \lambda^\top 
                  \left ( \, \mathbb{E}_{x\,\sim\,p} 
                              [ \,r(x)\, ] - b \,\right).
\end{align}
where we use shorthand $b \DefinedAs [\, b_1, \ldots, b_m \,]^\top$, $r \DefinedAs [\, r_1, \ldots, r_m \,]^\top$, and $\lambda \DefinedAs [\lambda_1,\ldots\,\lambda_m]^\top$ is the Lagrangian multiplier or dual variable. Let the dual function be  $D_{\text{ALI}}(\lambda) \DefinedAs \minimize_{p} L_{\text{ALI}}(p,\lambda)$ and an optimal dual variable be $\lambda^\star\in\argmax_{\lambda\,\geq\,0} D_{\text{ALI}}(\lambda)$. Denote $D_{\text{ALI}}^\star \DefinedAs D_{\text{ALI}}(\lambda^\star)$. For any $\lambda > 0$, we define the reward weighted distribution $q_{\text{rw}}^{(\lambda)}$ (subscript $\text{rw}$ for \emph{reward weighted}):
\begin{equation}\label{eq: conjunction}
    q_{\text{rw}}^{(\lambda)} (\cdot)
    \; \DefinedAs \;
    \frac{1}{ Z_{\text{\normalfont 
    rw}}(\lambda)}
    q(\cdot) {\rm e}^{\lambda^\top r(\cdot)}
\end{equation}
where $Z_{\text{\normalfont 
    rw}}(\lambda) = \int q(x) {\rm e}^{\lambda^\top r(x)} dx$ is the normalizing constant.

In the distribution space, Problem~\eqref{eq: constrained alignment reverse KL} is a convex optimization problem, since the KL divergence is strongly convex and the reward constraints are linear in $p$. Thus, we can apply strong duality in convex optimization~\cite{boyd2004convex} to characterize the solution to Problem~\eqref{eq: constrained alignment reverse KL} in Theorem~\ref{thm: alignment}. Moreover, it is ready to formulate the constrained alignment problem~\eqref{eq: constrained alignment reverse KL} as an unconstrained problem by specializing the dual variable to a solution to the dual problem. 

\begin{assumption}[Feasibility]\label{as:feasibility alginment}
    There exists a model $p$ such that $\mathbb{E}_{x\,\sim\,p}[ \,r_i(x)\,] > b_i$ for all $i = 1, \ldots, n$.
\end{assumption}

\begin{theorem}[Reward alignment]\label{thm: alignment}
    Let Assumption~\ref{as:feasibility alginment} hold. 
    Then, Problem~\eqref{eq: constrained alignment reverse KL} is strongly dual, i.e., $P_{\text{\normalfont ALI}}^\star = D_{\text{\normalfont ALI}}^\star$. Moreover, Problem~\eqref{eq: constrained alignment reverse KL} is equivalent to
     \begin{equation}\label{eq: unconstrained alignment reverse KL}
    \minimize_{p}
    \;
    D_{\text{\normalfont KL}}
    \big(\,
        p
        \,\Vert\,
        q_{\text{\normalfont rw}}^{(\lambda^\star)}
    \,\big)
    \end{equation}
    where $\lambda^\star$ is an optimal dual variable, and the dual function has the explicit form:
    $D_{\text{\normalfont ALI}}(\lambda) = -\log Z_{\text{\normalfont rw }} (\lambda)$. 
    Furthermore, an optimal solution of~\eqref{eq: constrained alignment reverse KL} is given by
    \begin{equation}\label{eq: unconstrained alignment reverse KL solution}
        p^\star 
        \; = \;
        q_{\text{\normalfont rw}}^{(\lambda^\star)}.
    \end{equation}
\end{theorem}
See Appendix~\ref{app: proof of alignment} for the proof. Theorem~\ref{thm: alignment} provides a closed-form solution to the constrained alignment problem~\eqref{eq: constrained alignment reverse KL}, i.e., $q_{\text{\normalfont rw}}^{(\lambda^\star)}$. This solution generalizes the reward-tilted distribution~\cite{domingoenrich2025adjointmatchingfinetuningflow}, which corresponds to finetuning a model with an expected reward regularizer. In Problem~\eqref{eq: constrained alignment reverse KL}, the optimal dual variable $\lambda^\star$ assigns weights to the rewards such that all the constraints are satisfied optimally, while remaining as close as possible to the pretrained model.

\subsection{Reward alignment of diffusion models}
\label{subsec: Reward alignment for diffusion models}

We introduce diffusion models into Problem~\eqref{eq: constrained alignment reverse KL} by representing $p$ and $q$ as two diffusion models: $p_{0:T}(\cdot; s_p)$ and $q_{0:T}(\cdot; s_q)$, with score functions $s_p$ and $s_q$, respectively. The path-wise KL divergence has been widely used in diffusion model alignment to capture the difference between two diffusion models~\cite{uehara2024understanding}. Hence, we instantiate Problem~\eqref{eq: constrained alignment reverse KL} in a space of score functions as follows
\begin{equation}\label{eq: constrained alignment reverse KL diffusion}\tag{SR-A}
    \begin{array}{rl}
        \displaystyle\minimize_{ s_p \,\in\, \mathcal{S}} & 
        D_{\text{KL}} \big(\, p_{0:T}(\cdot; s_p)\,\Vert\, q_{0:T}(\cdot; s_q) \,\big)
        \\[0.2cm]
        \subject &  \mathbb{E}_{x_0\,\sim\,p_0(\cdot; s_p)} \big[ \,r_i(x_0)\, \big]\; \geq \; b_i \;\;\;\; \text{ for } i = 1, \ldots, m.
    \end{array}
\end{equation}
We define the Lagrangian for Problem~\eqref{eq: constrained alignment reverse KL diffusion} as $\bar{L}_{\text{ALI}}(s_p,\lambda) \DefinedAs L_{\text{ALI}}(p_{0:T}(\cdot;s_p),\lambda)$. Similarly, we introduce the primal and dual values: $\bar{P}_{\text{ALI}}^\star$ and $\bar{D}_{\text{ALI}}^\star$. In general, Problem~\eqref{eq: constrained alignment reverse KL diffusion} is not guaranteed to be convex, since the path-wise KL divergence~\eqref{eq:path KL} involves an expectation taken over the backward process $p_{0:T}(\cdot)$. Nevertheless, the path-wise KL divergence is convex in the entire path space $\{ p_{0:T}(\cdot)\}$, and constraints are linear. Hence, when the score function class $\mathcal{S}$ is expressive enough to induce any path distribution, we establish strong duality for Problem~\eqref{eq: constrained alignment reverse KL diffusion} in Theorem~\ref{thm:Strong duality alginment}.

\begin{theorem}[Strong duality]\label{thm:Strong duality alginment}
    Let Assumption~\ref{as:feasibility alginment} hold for some $s\in\mathcal{S}$. If any path distribution $p_{0:T}(\cdot)$ can be induced by a score function $s_p\in \mathcal{S}$, then Problem~\eqref{eq: constrained alignment reverse KL diffusion} is strongly dual, i.e., $\bar{P}_{\text{\normalfont ALI}}^\star=\bar{D}_{\text{\normalfont ALI}}^\star$.
\end{theorem}

See Appendix~\ref{app:Strong duality alginment} for the proof. It is mild to assume the score function class is expressive, as diffusion models typically employ overparameterized networks (e.g., U-Nets or transformers) in practice. Motivated by strong duality, we propose a dual-based method for solving Problem~\eqref{eq: constrained alignment reverse KL diffusion}, alternating between minimizing the Lagrangian via gradient descent and maximizing the dual function via dual sub-gradient ascent below.

\textbf{Primal minimization: } At iteration $n$, we obtain a new model $s^{(n+1)}$ via a Lagrangian maximization:
\[
    s^{(n + 1)} 
    \; \in  \;
    \argmin_{s \,\in\, \mathcal{S}}\; \bar{L}_{\text{ALI}}(s_{p}, \lambda^{(n)}).
\]
\textbf{Dual maximization: } Then, we use the model $s^{(n+1)}$ to estimate the constraint violation $\mathbb{E}_{x_0}[ r (x_0)] - b$, denoted as $r(s^{(n+1)}) - b$, and perform a dual sub-gradient ascent step:
\[
    \lambda^{(n + 1)} 
    \; = \;
    \left[
    \, \lambda^{(n)} 
    \, +\,
    \eta\, \left(r(s^{(n + 1)}) - b\right) \,\right]_+.
\]

%% file: sec4_composition.tex
\section{Constrained Composition of Multiple Pretrained Models}\label{sec: composition reverse KL}

We provide a characterization of the solution to Problem~\eqref{eq: constrained composition reverse KL} in Section~\ref{subsec:Composition in distribution space}, and establish strong duality for diffusion models in Section~\ref{subsec:Product composition of diffusion models}, together with a dual-based training algorithm.

\subsection{Composition in distribution space}
\label{subsec:Composition in distribution space}


To apply Problem~\eqref{eq: constrained composition reverse KL} to diffusion models, we first employ Lagrangian duality to derive its solution in distribution space.
Let the Lagrangian for Problem~\eqref{eq: constrained composition reverse KL} be 
\begin{equation}
    L_{\text{AND}}(p, u, \lambda)
    \; = \; u + 
    \sum_{i\,=\,1}^m \lambda_i \left(
    D_{\text{KL}}(p\,\Vert\, q^i) - u
    \right),
\end{equation}
and the associated dual function $D_{\text{AND}}(\lambda)$, which is always concave, is defined as
\begin{equation}
      D_{\text{AND}}(\lambda) \;:=\; \max_{u,\ p} \; L_{\text{AND}}(p, u, \lambda).
\end{equation}

Let a solution to Problem~\eqref{eq: constrained composition reverse KL} be $(p^\star, u^\star)$, and let the optimal value of the objective function be $P_{\text{AND}}^\star = u^\star$. Let an optimal dual variable be $\lambda^\star \in \argmax_{\lambda \,\geq\,0} D_{\text{AND}}(\lambda)$, and the optimal value of the dual function be $D_{\text{AND}}^\star \DefinedAs D_{\text{AND}}(\lambda^\star)$. For any $\lambda> 0$, we define the tilted product distribution $q_{\text{AND}}^{(\lambda)}$ as a product of $m$ tilted distributions $\{q^i\}_{i\,=\,1}^m$:
\begin{equation}\label{eq: conjunction}
    q_{\text{AND}}^{(\lambda)} (\cdot)
    \; = \;
    \frac{1}{ Z_{\text{\normalfont 
    AND}}(\lambda)}
    \prod_{i\,=\,1}^m \left( q^i(\cdot) \right)^{\frac{\lambda_i}{\one^\top \lambda}} 
\end{equation}
where $Z_{\text{\normalfont 
    AND}}(\lambda) \DefinedAs \int \prod_{i\,=\,1}^m \left( q^i(x) \right)^{\frac{\lambda_i}{\one^\top \lambda}} dx$ is the normalizing constant.

In the distribution space, Problem~\eqref{eq: constrained composition reverse KL} is a convex optimization problem, since the sub-level set of the KL divergence is convex. Again, we apply strong duality in convex optimization~\cite{boyd2004convex} to characterize the solution to Problem~\eqref{eq: constrained composition reverse KL} in Theorem~\ref{thm: reverse KL general}. Moreover, it is ready to formulate the constrained composition problem~\eqref{eq: constrained composition reverse KL} as an unconstrained problem by specializing the dual variable to a solution to the dual problem. 

\begin{assumption}[Feasibility]\label{as:feasibility product}
    There exists a pair $(p,u)$ such that $D_{\text{\normalfont KL}}(p\,\Vert\,q^i) < u$ for all $i = 1, \ldots, n$.
\end{assumption}

\begin{theorem}[Product composition]\label{thm: reverse KL general}
    Let Assumption~\ref{as:feasibility product} hold. Then, Problem~\eqref{eq: constrained composition reverse KL} is strongly dual, i.e., $P_{\text{\normalfont AND}}^\star = D_{\text{\normalfont AND}}^\star$. Moreover, Problem~\eqref{eq: constrained composition reverse KL} is equivalent to 
    \begin{equation}\label{eq: unconstrained composition reverse KL}
    \minimize_{p}
    \;
    D_{\text{\normalfont KL}}\Big(\,p\,\Vert\,  q_{\text{\normalfont AND}}^{(\lambda^\star)}\,\Big)
    \end{equation}
    where $\lambda^\star$ is the optimal dual variable, and the dual function has the explicit form,
    $D(\lambda) = -\log Z_{\text{\normalfont AND }} (\lambda)$. Furthermore, the optimal solution of~\eqref{eq: unconstrained composition reverse KL} is given by
    \begin{equation}\label{eq: unconstrained alignment reverse KL solution}
        p^\star 
        \; = \;
        q_{\text{\normalfont AND}}^{(\lambda^\star)}.
    \end{equation}
\end{theorem}

See Appendix~\ref{app:proofs_reverse KL general} for proof. The distribution $q_{\text{\normalfont AND}}^{(\lambda)} \propto \prod_{i\,=\,1}^m \left( q^i(\cdot) \right)^{\frac{\lambda_i}{\one^\top \lambda}}$ allows sampling from a weighted product of $m$ distributions $\{q^i \}_{i\,=\,1}^m$, where the parameters $\{ \lambda_i/\one^\top \lambda \}_{i\,=\,1}^m$ weight the importance of each distribution. The geometric mean is a special case when all $\lambda_i$ are equal~\cite{biggs2024diffusion}.

\begin{remark}
    Theorem~\ref{thm: reverse KL general} connects our proposed constrained optimization problem~\eqref{eq: constrained composition reverse KL} to the well-known problem of sampling from a product of multiple distributions~\cite{biggs2024diffusion, du2024reducereuserecyclecompositional}. Furthermore, our constraints enforce that the resulting product is properly weighted to ensure the solution diverges as little as possible from each of the individual distributions (see Figure~\ref{fig: gaussians AND} for illustration).
\end{remark}

\subsection{Product composition of diffusion models}
\label{subsec:Product composition of diffusion models}

We introduce diffusion models into Problem~\eqref{eq: constrained composition reverse KL} by representing $p$ and $q^i$ as two diffusion models: $p_0(x_0; s_p)$ and $q_0^i(x_0; s_q^i)$, with score functions $s_p$ and $s_q^i$, respectively. The point-wise KL divergence naturally measures the closeness of the final sampling distributions we care about. Hence, we instantiate Problem~\eqref{eq: constrained composition reverse KL} in a space of score functions as follows
\begin{equation}\label{eq: diffusion score problem}\tag{SR-C}
    \begin{array}{rl}
        \displaystyle\minimize_{{u \,\geq\, 0} ,\ {s_p \,\in\, \mathcal{S}}} & 
        {u}
        \\[0.2cm]
        \subject &  D_{\text{KL}}(p_0(x_0; s_p)\,\Vert\, q_0(x_0; s_q^i))\; \leq \; {u} \;\;\;\; \text{ for } i = 1, \ldots, m.
    \end{array}
\end{equation}

We define the Lagrangian for Problem~\eqref{eq: diffusion score problem} as $\bar{L}_{\text{AND}}(s_p,u,\lambda) \DefinedAs L_{\text{AND}}(p_0(x_0;s_p),u, \lambda)$. Similarly, we introduce the primal and dual values $\bar{P}_{\text{AND}}^\star$ and $\bar{D}_{\text{AND}}^\star$. Although Problem~\eqref{eq: diffusion score problem} is non-convex, since the point-wise KL divergence~\eqref{eq: marginal kl computation} involves an expectation taken over the backward process $p_{0:T}(\cdot)$. Nevertheless, the point-wise KL divergence is convex in the final distribution space. Hence, when the score function class $\mathcal{S}$ is expressive enough to induce any path distribution (hence any final distribution), we establish strong duality for Problem~\eqref{eq: diffusion score problem} in Theorem~\ref{thm:Strong duality composition}.

\begin{theorem}[Strong duality]\label{thm:Strong duality composition}
    Let Assumption~\ref{as:feasibility product} hold for some $s\in \mathcal{S}$. If any path distribution $p_{0:T}(\cdot)$ can be induced by a score function $s_p\in\mathcal{S}$, then Problem~\eqref{eq: diffusion score problem} is strongly dual, i.e., $\bar{P}_{\text{\normalfont AND}}^\star = \bar{D}_{\text{\normalfont AND}}^\star$.
\end{theorem}

See Appendix~\ref{app:Strong duality composition} for proof. It is mild to assume the score function class is expressive, as diffusion models typically employ overparameterized networks (e.g., U-Nets or transformers) in practice. To solve Problem~\eqref{eq: diffusion score problem}, similar to the one in Section~\ref{subsec: Reward alignment for diffusion models}, we apply a dual-based approach below. 

\textbf{Primal minimization: } At iteration $n$, we obtain a new model $s^{(n+1)}$ via a Lagrangian maximization:
\[
    s^{(n + 1)} 
    \; \in  \;
    \argmin_{s_p \,\in\, \mathcal{S}}\; \bar{L}
    _{\text{AND}}(s_{p}, \lambda^{(n)}).
\]
\textbf{Dual maximization: } Then, we use the model $s^{(n+1)}$ to estimate the constraint violation and perform a dual sub-gradient ascent step: 
\[
    \lambda_i^{(n + 1)} 
    \; = \;
    \left[
    \, \lambda_i^{(n)} 
    \, +\,
    \eta\, \left(D_{\text{KL}}\big(p_0(x_0; s^{(n + 1)})\,\Vert\, q_0^i(x_0; s_q^i)\big) - u \right) \,\right]_+\; \text{ for } i = 1,\ldots,m.
\]
It is nontrivial to compute the point-wise KL divergence in the Lagrangian $\bar{L}
    _{\text{AND}}(s_{p}, \lambda^{(n)})$ and the constraint violations above. Recall that Lemma~\ref{lem: marginal kl computation} gives us a way to compute the point-wise KL: $ D_{\text{KL}}(p_0(x_0; s)\,\Vert\, q_0(x_0; s_q^i))$. However, it requires the functions $s$ and $s_q^i$ each to be a valid score function for some forward process. Indeed, this is the case for $s_q^i$, since it is a pretrained model where it would have been trained to approximate the true score of a forward process. Yet, regarding the function $s$ that we are optimizing over, there is no guarantee that any given $s \in \mathcal{S}$ is a valid score function. To address this issue, we introduce Lemma~\ref{lem: primal true score} that allows us to minimize the Lagrangian.
\begin{lemma}\label{lem: primal true score}
    The Lagrangian for Problem~\eqref{eq: diffusion score problem} is equivalently written as
    \begin{equation}
        L_{\text{\normalfont AND}} (s, \lambda) 
        \; = \; D_{\text{\normalfont KL}}\Big(\,p_0(x_0; s)\,\Vert\, q_{\text{\normalfont AND}}^{(\lambda)}(x_0)\,\Big) - \log Z_{\text{\normalfont AND}}(\lambda).
    \end{equation}
    Furthermore, a Lagrangian minimizer $s^{(\lambda)} \in \argmin_{s} L_{\text{\normalfont AND}}(s, \lambda)$ is given by
    \begin{equation}\label{eq: true score optimization}
        s^{(\lambda)} 
        \; \in \; 
        \argmin_{s \,\in\, \mathcal{S}} \quad \sum_{t \,=\, 0}^T \omega_t \, \mathbb{E}_{x_0 \,\sim\, q_{\text{\normalfont AND}}^{(\lambda)}(\cdot)} \mathbb{E}_{x_t \,\sim\, q(x_t\,|\,x_0)}\left[ \norm{s(x_t, t) - \nabla \log q(x_t \,|\, x_0)}^2 \right]
    \end{equation}
    where $q(x_t \,|\, x_0) \sim \mathcal{N}(\sqrt{\bar \alpha_t} x_0, (1 - \bar \alpha_t) I)$, and $ s^{(\lambda)}(x_t,t) = \nabla \log q^{(\lambda)}_{\text{\normalfont AND}}(x_t)$.
\end{lemma}
See Appendix~\ref{app: proofs} for proof. With Lemma~\ref{lem: primal true score}, as long as we can obtain samples from the distribution $q_{\text{\normalfont AND}}^{(\lambda)}$, we can approximate the expectation in~\eqref{eq: true score optimization} and use gradient-based optimization methods to find a Lagrangian minimizer $s^{(\lambda)}$. To do so, we use annealed Markov Chain Monte Carlo (MCMC) sampling~\cite{du2024reducereuserecyclecompositional}, which requires having access to the scores of a sequence of distributions that interpolate smoothly between $q_{\text{AND}}^{(\lambda)}(x_T)$ and $q_{\text{AND}}^{(\lambda)}(x_0)$: $\nabla \log q_{\text{AND}}^{(\lambda)}(x_t) = \sum_{i \,=\, 1}^m \lambda_i \nabla \log q^i(x_t)$. In alignment, since we don't have these `intermediate' scores, we cannot employ the approach in Lemma~\ref{lem: primal true score}. See Appendix~\ref{app: related work} for sampling details. 

For the dual update, we evaluate the KL divergence $D_{\text{KL}}(p_0(\cdot; s^{(\lambda)})\,\Vert\, p_0(\cdot; s^i))$ between the marginal densities induced by the Lagrangian minimizer $s^{(\lambda)}$ and the individual score functions $s^i$ using Lemma~\ref{lem: marginal kl computation}, since both are valid score functions.

\begin{remark}
    In practice, the primal step only yields an approximate Lagrangian minimizer $s^{(\tilde \lambda)}(x, t) \approx \nabla \log q^{(\lambda)}_{\text{\normalfont AND},\, t} (x)$. This results in two sources of error in evaluating the expectations on the RHS of~\eqref{eq: marginal kl computation}:
    \begin{equation}
        D_{\text{\normalfont KL}} (p_0(\cdot\,; s^{(\lambda)})\,\Vert\, p_0(\cdot\,; s^i)) 
        \; = \;
        \sum_{t \,=\, 0}^T \tilde \omega_t \, \mathbb{E}_{x\,\sim\, p_t(\cdot\,; s^{(\lambda)})} \left[ \,
        \norm{s^{(\lambda)}(x, t) - s^i(x, t)}_2^2
        \,\right] 
        \,+\, \epsilon_T
    \end{equation}
    The first error caused by not using the exact $s^{(\lambda)}$ in $\norm{s^{(\lambda)}(x, t) - s^i(x, t)}_2^2$. The second error introduced by not evaluating the expectation on correct trajectories given by $x\,\sim\, p_t(\cdot\,; s^{(\lambda)})$. However, the second error reduces, if we have a way of sampling from the true product $x_0 \sim q^{(\lambda)}_{\text{\normalfont AND},\, 0}$, because we can get samples from $p_t(\cdot\,; s^{(\lambda)})$ just by adding Gaussian noise to $x_0$. 
\end{remark}

See Appendix~\ref{app: algorithms} for the detailed algorithm of product composition.

%% file: sec5_experiments.tex
\section{Computational Experiments}\label{sec: experiments}

We demonstrate the effectiveness and merits of our constrained alignment and composition in a series of computational experiments in Section~\ref{subsec: alignment experiments} and Section~\ref{subsec: composition experiments}, respectively. 

\subsection{Alignment of diffusion models with multiple rewards}\label{subsec: alignment experiments}

We extend the AlignProp framework~\cite{prabhudesai2024aligningtexttoimagediffusionmodels} to handle multiple rewards as constraints. We finetune Stable Diffusion v1.5\footnote{https://huggingface.co/stable-diffusion-v1-5/stable-diffusion-v1-5} using several widely-used differentiable image quality and aesthetic rewards: aesthetic~\cite{aesthetic},   hps~\cite{hps}, pickscore~\cite{pickscore}, imagereward~\cite{imagereward} and MPS~\cite{mps}. Since these rewards vary substantially in scale, making it difficult to set constraint levels, we normalize each by computing the average and standard deviation over a number of batches. In all experiments, models are finetuned using LoRA~\cite{hu2022lora}. Experimental settings and hyperparameters are provided in Appendix~\ref{app: experiments}. 

\begin{figure}[H]
  \centering
  \begin{subfigure}[b]{0.55\textwidth}
    \centering
    \includegraphics[width=\linewidth]{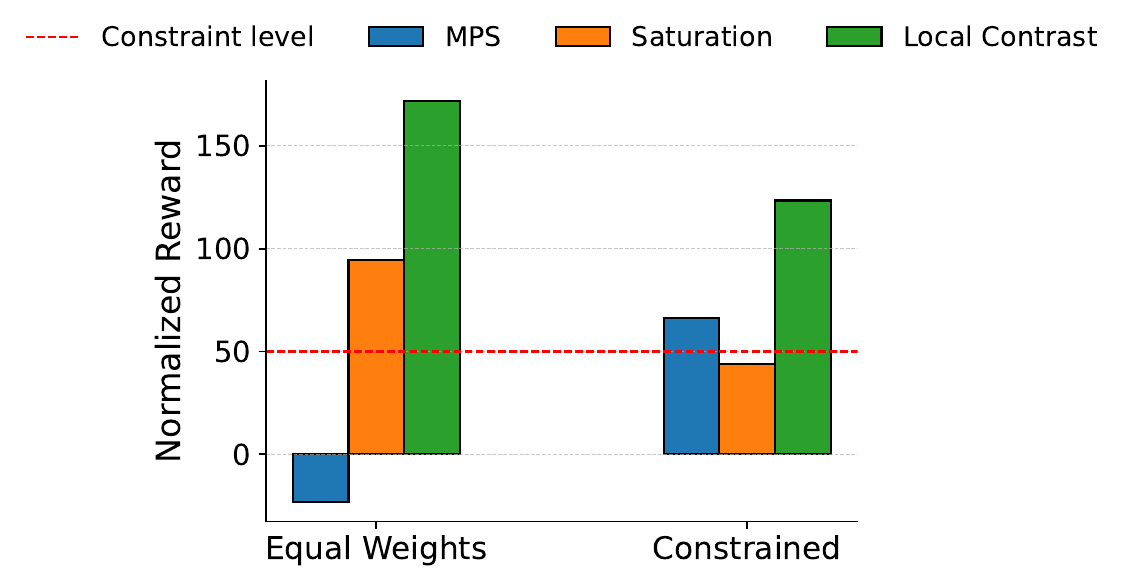}
  \end{subfigure}\hfill
  \begin{subfigure}[b]{0.35\textwidth}
    \centering
    \includegraphics[width=\linewidth]{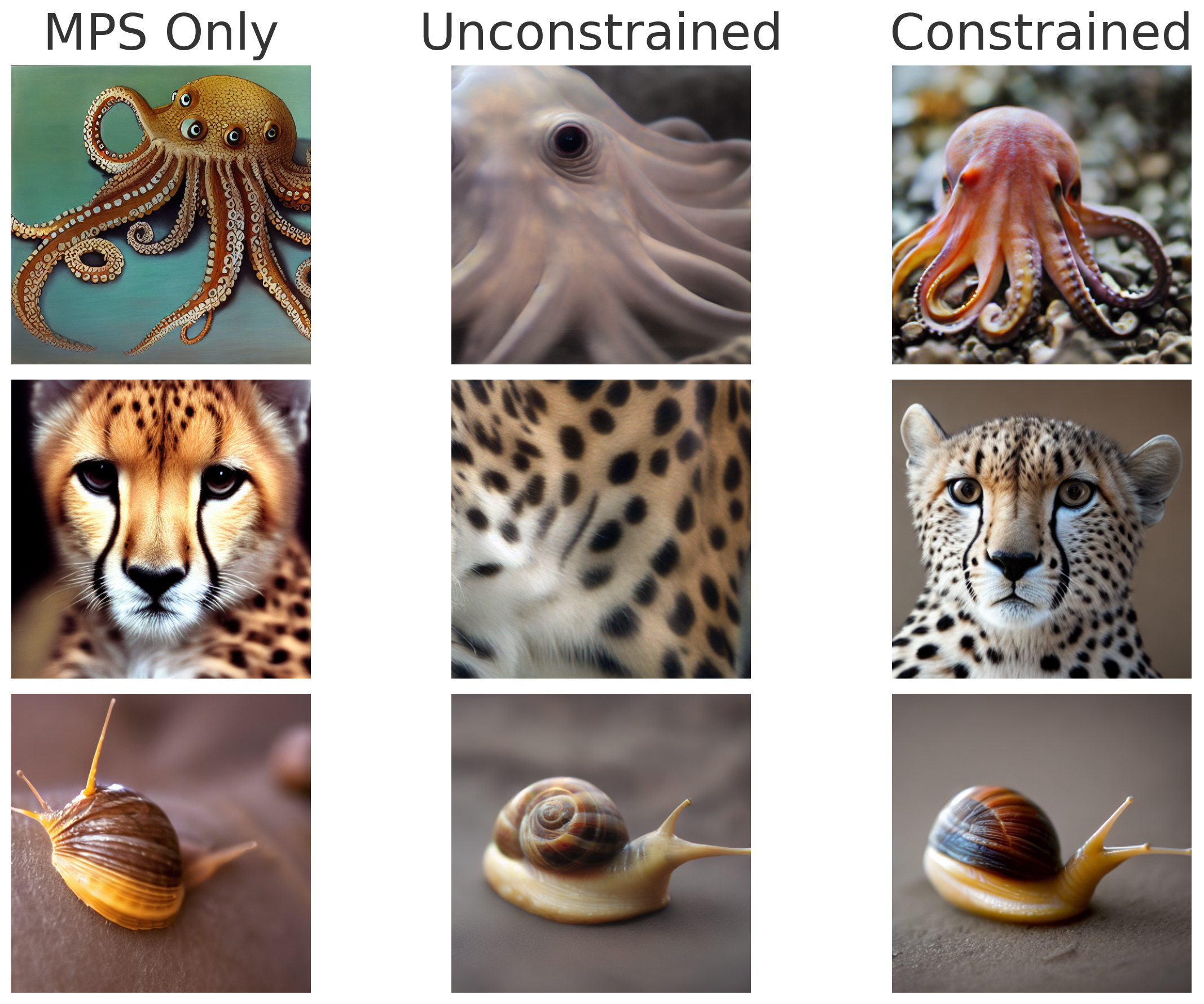}
  \end{subfigure}
  \caption{Reward alignment. Stable diffusion is finetuned using one reward that emphasizes aesthetic quality (MPS), and Saturation and Local Contrast as regularizers. Reward values for the equal weights method and our constrained alignment (Left). Images are sampled from the aligned models (Right), and the model trained solely with MPS reward is used for comparison.}
  \label{fig: mps alignment}
\end{figure}

\textbf{I. MPS, local contrast, and saturation constraints.} A common shortcoming of several off-the-shelf aesthetics, image preference, and quality reward models is their tendency to overfit to specific image characteristics such as saturation and sharp, high-contrast textures; see, for example, images in the first column in Figure~\ref{fig: mps alignment} (Right). To mitigate this issue, we add regularizers to the reward function to explicitly penalize these characteristics. However, if the regularization weight is not carefully tuned, models may overfit to the regularizers instead of optimizing for the intended reward.  As shown in Figure~\ref{fig: mps alignment}, when using equal weights, the MPS reward \emph{decreases} (Left). In contrast, our constrained approach effectively controls multiple undesired artifacts while ensuring none of the rewards are neglected, achieving a near feasible solution at the specified constraint level: a 50\% improvement.

\begin{figure}
  \centering
  \begin{subfigure}[b]{0.33\textwidth}
    \centering
    \includegraphics[width=\linewidth]{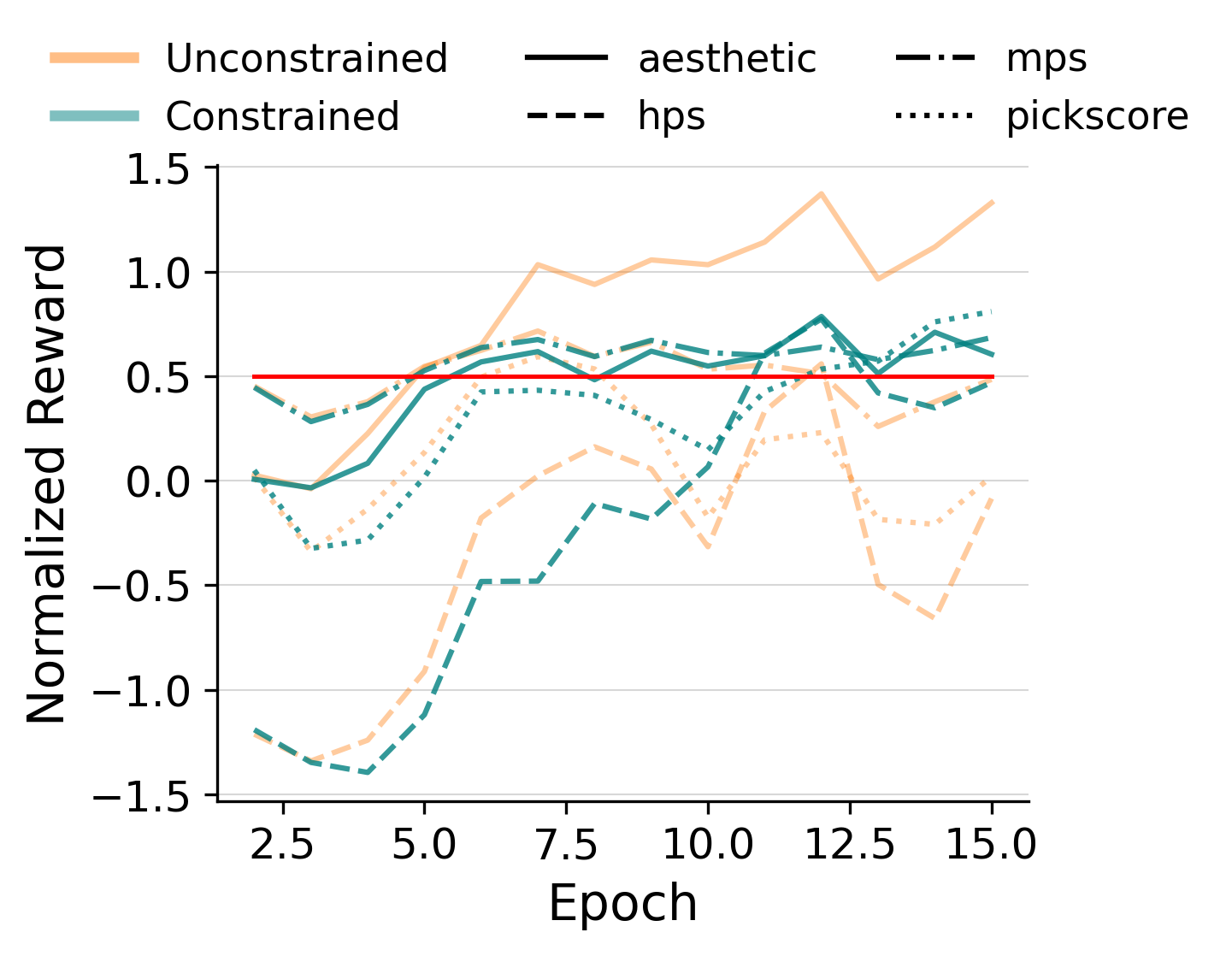}
  \end{subfigure}\hfill
  \begin{subfigure}[b]{0.3\textwidth}
    \centering
    \includegraphics[width=\linewidth]{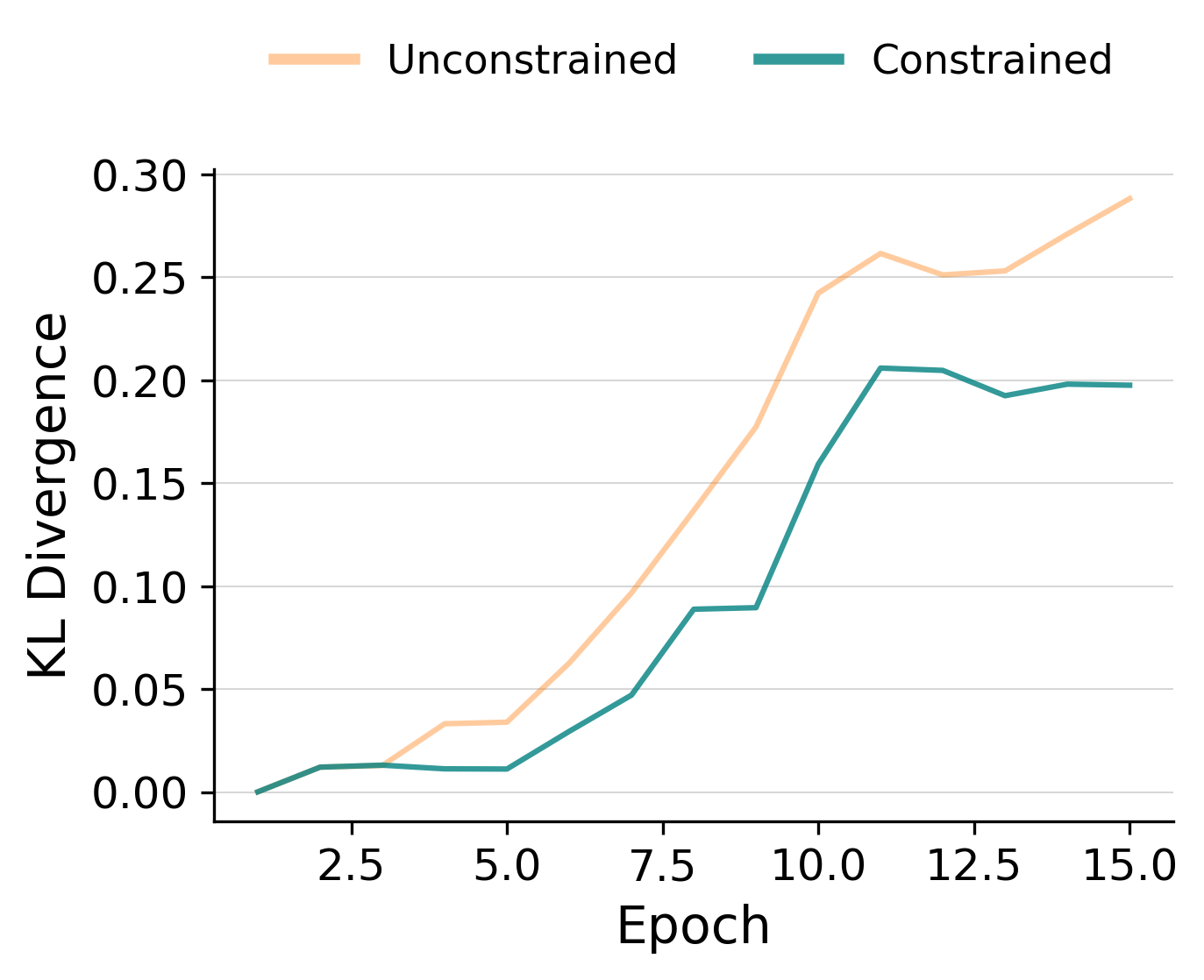}
  \end{subfigure}\hfill
  \begin{subfigure}[b]{0.33\textwidth}
    \centering
    \includegraphics[width=\linewidth]{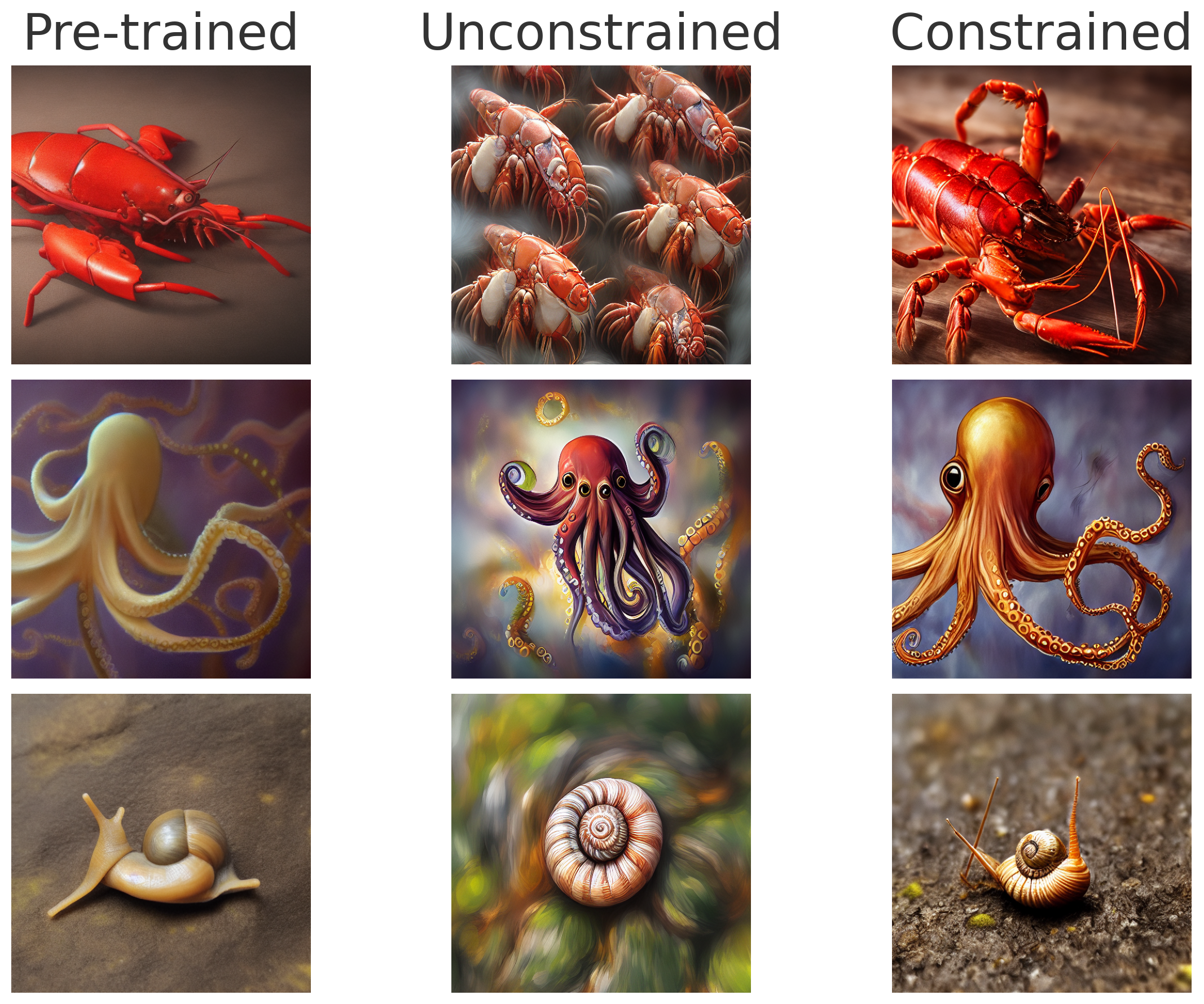}
  \end{subfigure}
  \caption{Reward alignment. Stable diffusion is finetuned using multiple image quality/aesthetic rewards. Reward trajectories for the regularization-based method and our constrained alignment during training (Left). KL divergences to the pretrained model (Middle). Images are sampled from the aligned models (Right), and the pretrained model is used for comparison. }
  \label{fig: all rewards alignment}
\end{figure}

\textbf{II. Multiple aesthetic constraints.} When finetuning with multiple rewards, arbitrarily assigning fixed weights can lead to uneven performance across rewards. As shown in Figure~\ref{fig: all rewards alignment} (Left), the model tends to overfit one reward while neglecting more challenging ones (e.g., hps). In contrast, constraining all rewards enables the model to improve each reward up to its specified level, including the challenging ones. From Figure~\ref{fig: all rewards alignment} (Middle), minimizing the KL divergence subject to these constraints also yields a smaller KL divergence to the pretrained model. Without constraints, overfitting to a subset of rewards causes the model to deviate excessively from the pretrained one, which is undesirable (Right).

\subsection{Product composition of diffusion models}\label{subsec: composition experiments}

In high-dimensional settings such as image generation, obtaining samples from the true product distribution via MCMC and then minimizing the Lagrangian in~\eqref{eq: true score optimization} to estimate the true product score function is prohibitively expensive. To address this, we employ a surrogate for the true score both for sampling and for computing the KL divergence, as detailed in Appendix~\ref{app: experiments}.

\begin{figure}[htbp]
    \centering
    \begin{minipage}[t]{0.48\linewidth}
        \centering
        \includegraphics[width=0.95\linewidth]{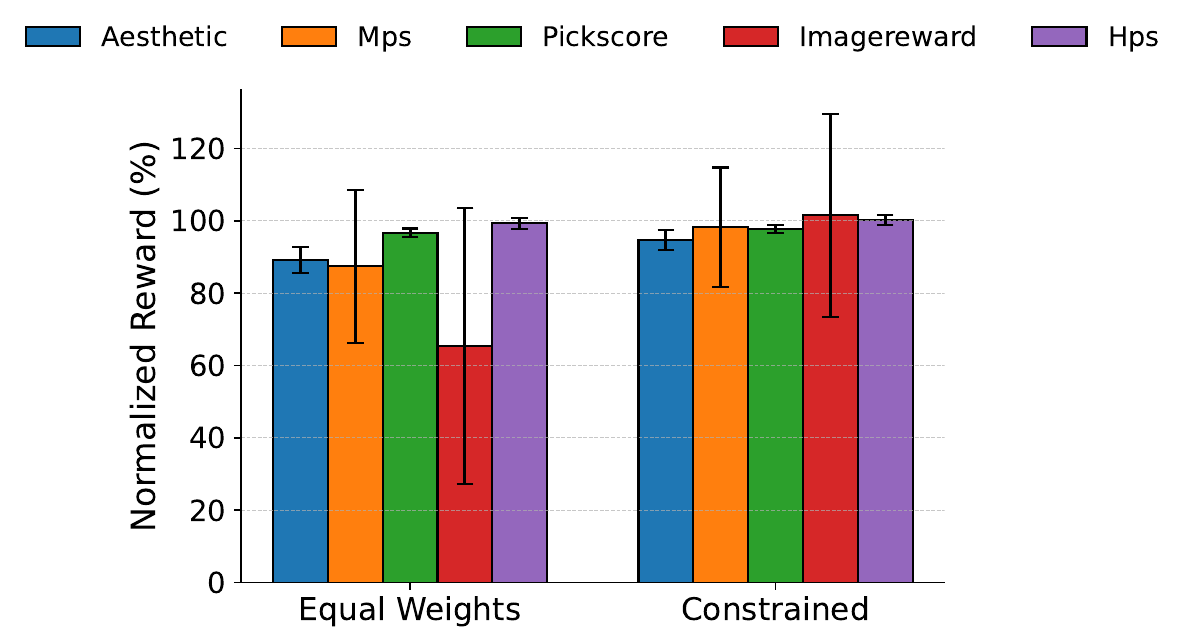}
        \caption{Product composition. Stable diffusion with LoRA is  finetuned using different rewards, for equal weighted and product mixtures. 100\% represents the reward levels attained by models aligned solely with the individual reward. Higher is better.}
        \label{fig:composition}
    \end{minipage}%
    \hfill
    \begin{minipage}[t]{0.48\linewidth}
        \centering
        \includegraphics[width=0.95\linewidth]{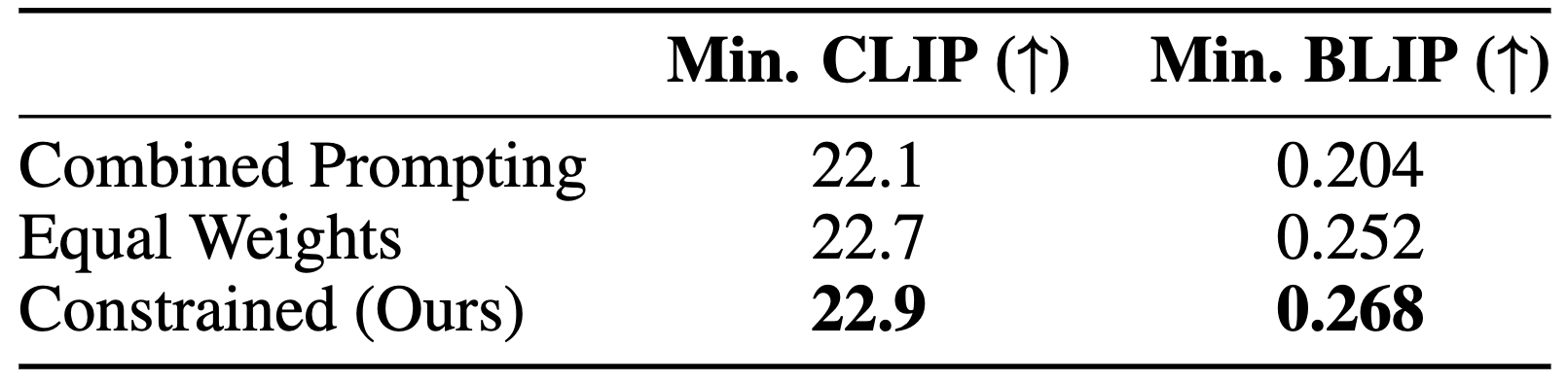}
        \captionof{table}{Product composition. We compare our constrained composition with two baselines using minimum CLIP and BLIP scores. The score is averaged over 50 different prompt pairs that are sampled from a list of simple prompts.} 
        \label{tab:important_table}
    \end{minipage}
\end{figure}

\textbf{I. Composing models finetuned on different rewards.} We investigate the composition of several finetuned variants of the same base model, where each model is trained with LoRA a different reward function. A key challenge is determining appropriate combination weights: arbitrary choices can lead to undesirable trade-offs and underrepresentation of certain models in the mixture, as evidenced in Figure~\ref{fig:composition} by drops in up to 30\% in some rewards. Our constrained composition provides a principled way to select weights that maintain proximity to each model, improving rewards across all models. 

\textbf{II. Concept composition with stable diffusion.} Following the setting in~\cite{skreta2025the}, we compose two text-to-image diffusion models, each conditioned on a different input prompt. We apply the constrained composition~\eqref{eq: diffusion score problem} to determine the optimal weights for composing two models, and compare against the baseline that uses equal weights. Closeness to each model encourages faithful representation of both concepts in the images generated by the composed model, as reflected by improved text-to-image similarity metrics: CLIP~\cite{hessel2022clipscorereferencefreeevaluationmetric} and BLIP~\cite{li2022blipbootstrappinglanguageimagepretraining}, which are reported in Table~\ref{tab:important_table}. We compute similarity scores between the generated images and each of the two prompts and compare their minimum values. We also include a baseline where images are generated from a single combined prompt containing both inputs. Images from all approaches, along with implementation details and additional experimental results, are provided in Appendix~\ref{app: experiments}.

%% file: sec6_conclusion.tex
\section{Conclusion}

We have developed a constrained optimization framework that unifies alignment and composition of diffusion models by enforcing that the aligned model satisfies reward constraints and/or remains close to each pretrained model. Theoretically, we characterize the solutions to the constrained alignment and composition problems and design dual-based training algorithms to approximate these solutions. Empirically, we demonstrate our constrained approach on image generation tasks, showing that the aligned or composed models effectively satisfy the specified constraints.


%% file: app1_proofs.tex
\section{Limitations and Broader Impact}\label{app: limitations}

\noindent\textbf{Limitations}: Despite offering a unified constrained learning framework and demonstrating strong empirical results, further experiments are needed to assess our method’s effectiveness on alignment and composition tasks beyond image generation, under mixed alignment and composition constraints, and in combination with inference-time techniques. Additionally, further theoretical work is needed to understand optimality of non-convex constrained optimization, convergence and sample complexity of primal-dual training algorithms.

\noindent\textbf{Broader impact}: Our method can enhance diffusion models' compliance with diverse requirements, such as realism, safety, fairness, and transparency. By introducing a unified constrained learning framework, our work offers practical guidance for developing more reliable and responsible diffusion model training algorithms, with potential impact across applications such as content generation, robotic control, and scientific discovery.

\section{Related Work}\label{app: related work}

\textbf{Alignment of diffusion models}. Our constrained alignment is related to a line of work on finetuning diffusion models. Standard finetuning typically involves optimizing either a task-specific reward that encodes desired properties, or a weighted sum of this reward and a regularization term that encourages closeness to the pretrained model; see~\cite{fan2023optimizing,xu2023imagereward,lee2023aligning,wu2023human,zhang2023controllable,wu2024deep,black2024training,clark2024directly,zhang2024aligning} for studies using the single reward objective and~\cite{uehara2024fine,zhao2024scores,uehara2024bridging,uehara2024feedback,prabhudesai2024video,fan2023dpokreinforcementlearningfinetuning,han2024stochastic} for those using the weighted sum objective. The former class of single reward-based studies focus exclusively on generating samples with higher rewards, often at the cost of generalization beyond the training data. The latter class introduces a regularization term that regulates the model to be close to the pretrained one, while leaving the trade-off between reward and closeness unspecified; see~\cite{uehara2024understanding} for their typical pros and cons in practice. There are three key drawbacks to using either the single reward or weighted sum objective: (i) the trade-off between reward maximization and leveraging the utility of the pretrained model is often chosen heuristically; (ii) it is unclear whether the reward satisfies the intended constraints; and (iii) multiple constraints are not naturally encoded within a single reward function. In contrast, we formulate alignment as a constrained learning problem: minimizing deviation from the pretrained model subject to reward constraints. This offers a more principled alternative to existing ad hoc approaches~\cite{chen2024towards,giannone2023aligning}. Our new alignment formulation (i) offers a theoretical guarantee of an optimal trade-off between reward satisfaction and proximity to the pretrained model, and (ii) allows for the direct imposition of multiple reward constraints. We also remark that our constrained learning approach generalizes to finetuning of diffusion models with preference~\cite{wallace2024diffusion,yang2024using,li2024aligning}.

\textbf{Composition of diffusion models}. Our constrained composition approach is related to prior work on compositional generation with diffusion models. When composing pretrained diffusion models, two widely used approaches are (i) product composition (or conjunction) and (ii) mixture composition (or disjunction). In product composition, it has been observed that the diffusion process is not compositional, e.g., a weighted sum of diffusion models does not generate samples from the product of the individual target distributions~\cite{du2024reducereuserecyclecompositional,bradley2024classifier,chidambaram2024does}. To address this issue, the weighted sum approach has been shown to be effective when combined with additional assumptions or techniques, such as energy-based models~\cite{liu2022compositional,du2024reducereuserecyclecompositional}, MCMC sampling~\cite{du2024reducereuserecyclecompositional}, diffusion soup~\cite{biggs2024diffusion}, and superposition~\cite{skreta2025the}. However, how to determine optimal weights for the individual models is not yet fully understood. In contrast, we propose a constrained optimization framework for composing diffusion models that explicitly determines the optimal composition weights. Hence, this formulation enables an optimal trade-off among the pretrained diffusion models. Moreover, our constrained composition approach also generalizes to mixture composition, offering advantages over prior work~\cite{liu2022compositional,du2024reducereuserecyclecompositional,biggs2024diffusion,skreta2025the}.

\textbf{Diffusion models under constraints.} Our work is pertinent to a line of research that incorporates constraints into diffusion models. To ensure that generated samples satisfy given constraints, several ad hoc approaches have proposed that train diffusion models under hard constraints, e.g., projected diffusion models~\cite{liang2024multi,christopher2024constrained,liang2025simultaneous}, constrained posterior sampling~\cite{narasimhan2024constrained}, and proximal Langevin dynamics~\cite{zampini2025training}. In contrast, our constrained alignment approach focuses on expected constraints defined via reward functions and provides optimality guarantees through duality theory. A more closely related work considers constrained diffusion models with expected constraints, focusing on mixture composition~\cite{khalafi2024constrained}. In comparison, we develop new constrained diffusion models for reward alignment and product composition.

\section{Proofs}\label{app: proofs}

For conciseness, wherever it is clear from the context we omit the time subscript:
\begin{equation}
    D_{\text{KL}}(p_{0:T}(x_{0:T}; s_p)) = D_{\text{KL}}(p(x_{0:T}; s_p)).
\end{equation}

\subsection{Proof of Lemma~\ref{lem: reverse KL computation}}\label{app: proofs_reverse kl diffusion}

\begin{proof}
    The DDIM process is Markovian in reverse time with the conditional likelihoods given by
    \begin{equation}\label{eq: conditionals}
    p(x_{t - 1} \,|\, x_t; s) \; = \; \mathcal{N}\left(\sqrt{\frac{\alpha_{t-1}}{{\alpha}_t}} x_t 
\, + \,
\beta_t \, s(x_t, t), \,\sigma_t^2 I\right).
\end{equation}

Using~\eqref{eq: conditionals} we expand the path-wise KL:

\[
\begin{array}{rcl}
      &&\!\!\!\!  \!\!\!\!  \!\!\!\!
      \displaystyle
      D_{\text{KL}}(p_{0:T}(\cdot ; s_p)\,\Vert\, q_{0:T}(\cdot ; s_q)) 
      \\[0.2cm]
      &  =  & \mathbb{E}_{x_{0:T}\,\sim\,p} \left[\, \log p(x_{0:T}; s_p) - \log q(x_{0:T}; s_q) \,\right]
     \\[0.2cm]
     &  \overset{(a)}{=}  & \displaystyle
     \mathbb{E}_{x_{T}\,\sim\,p_{T+1}(\cdot), x_{T-1} \,\sim\, p_{T}(\cdot\,\vert\,x_T), \, \ldots \,,  x_0\,\sim\,p_1(\cdot\,\vert\,x_1)} 
     \left[ \sum_{t\,=\,T}^1 \log  \frac{p (x_{t-1}\,\vert\,x_t; s_p)}{q (x_{t-1}\,\vert\,x_t; s_q)} \right]
     \\[0.2cm]
     &  \overset{(b)}{=}  & \displaystyle \sum_{t\,=\,T}^1 
     \mathbb{E}_{x_{T}\,\sim\,p_{T+1}(\cdot), x_{T-1} \,\sim\, p_{T}(\cdot\,\vert\,x_T), \, \ldots \,,  x_0\,\sim\,p_1(\cdot\,\vert\,x_1)} 
     \left[ \log \frac{p(x_{t-1}\,\vert\,x_t; s_p)}{q(x_{t-1}\,\vert\,x_t; s_q)} \right]
     \\[0.2cm]
     &   \overset{(c)}{=}  & \displaystyle \sum_{t\,=\,T}^1 
     \mathbb{E}_{x_{0:T}\,\sim\,p} 
     \left[ D_{\text{KL}}(p(x_{t - 1}\,\vert\,x_t; s_p)\,\Vert\, q(x_{t - 1}\,\vert\,x_t; s_q)) \right]
     \\[0.2cm]
     &   \overset{(d)}{=}  & \displaystyle
     \sum_{t\,=\,T}^1 
     \mathbb{E}_{x_t\,\sim\,p_{t+1}} 
     \left[\frac{\beta_t^2}{2\sigma_t^2} \norm{ s_p(x_t,t) - s_q (x_t, t) }^2\right]
     \\[0.2cm]
     &   \overset{(e)}{=}  & \displaystyle
     \sum_{t\,=\,T}^1 
     \mathbb{E}_{ \{p_t\} } 
     \left[\frac{\beta_t^2}{2\sigma_t^2} \norm{ s_p(x_t,t) - s_q (x_t, t) }^2\right]
\end{array}
\]
where $(a)$ is due to the diffusion process, $(b)$ is due to the exchangeable sum and integration, $(c)$ is the definition of reverse KL divergence at time $t$, $(d)$ is due to the reverse KL divergence between two Guassians with the same covariance and means differing by $\beta_t (s_p(x_t, t) - s_q(x_t, t))$, and in $(e)$ we abbreviate $\mathbb{E}_{x_t\,\sim\,p_{t+1}}$ as $\mathbb{E}_{\{p_t\}}$ that is taken over the randomness of Markov process. 
\end{proof}

\subsection{Proof of Lemma~\ref{lem: marginal kl computation}}\label{app: proof of lemma marginal kl}

The proof for Lemma~\ref{lem: marginal kl computation} is quite involved, thus we have divided it into multiple parts for readability. In Section~\ref{subsec: prelim}, we give a few definitions for continuous time diffusion processes. In Section~\ref{subsec: cont time proof}, we prove an analogue of Lemma~\ref{lem: marginal kl computation} in continuous time. In Section~\ref{subsec: discretization error}, we bound the discretization error $\epsilon_T$ incurred when going from continuous time processes to corresponding discretized processes and thus complete the proof. The proofs for all lemmas presented here can be found in Appendix~\ref{app: Additional Proofs}.

\subsubsection{Continuous time preliminaries}\label{subsec: prelim}

\textbf{Notation Guide:} Throughout the proof, we will be dealing with continuous time forward and reverse diffusion processes and their discretized counterparts. 
\begin{itemize}
    \item We denote the continuous time variable $\tau \in [0, 1]$ to differentiate it from the discrete time indices $t \in \{0, \cdots, T \}$. $t = 0$ corresponds to $\tau = 1$ and $t = T$ corresponds to $\tau = 0$. \footnote{For consistency with other works from whom we will utilize some results in our proofs, namely~\cite{domingoenrich2025adjointmatchingfinetuningflow, lyu2012interpretationgeneralizationscorematching}, the direction of time we consider in continuous time is reversed compared to discrete time. This does not affect our derivations and results beyond a notation change.}

    \item We denote as $\mathfrak{X}_\tau$ the continuous time reverse DDIM process and $X_t$ as the corresponding discrete time process.

    \item The forward processes we denote with an additional bar e.g. $\bar{\mathfrak{X}}_\tau, \bar X_t$ denote the continuous time and discrete time forward processes respectively.

    \item Marginal density of continuos time DDIM process with score predictor $s(x, \tau)$ at time $\tau$ we denote as: $\mathfrak{p}_\tau(x, s)$.

\end{itemize}

Given a function $s(x,\tau):\mathbb{R}^d\times [0, 1] \rightarrow \mathbb{R}^d$, and a noise schedule $\bar\alpha_\tau$ increasing from $\bar \alpha_0 = 0$ to $\bar \alpha_1 = 1$, we define a continuous time reverse DDIM process as
\begin{equation}\label{eq: ddim continuous}
    d\mathfrak{X}_\tau \;=\; \left(\frac{\dot{\bar \alpha}_\tau}{2 \bar \alpha_\tau}\mathfrak{X}_\tau + (\frac{\dot{\bar \alpha}_\tau}{2 \bar \alpha_\tau} + \frac{\sigma_\tau^2}{2})s(\mathfrak{X}_\tau , \tau)\right) dt + \sigma_\tau d\mathfrak{B}_\tau, \;\;\;\mathfrak{X}_0 \sim \mathcal{N}(0, I).
\end{equation}
The variance schedule $\sigma_\tau$ is arbitrary and determines the randomness of the trajectories (e.g. if $\sigma_\tau = 0$ for all $\tau$, then the trajectories will be deterministic). The DDIM generative process~\eqref{eq: ddim continuous} induces marginal densities $\mathfrak{p}_\tau(x, s) $ for $\tau \in [0,1]$.

For reference the Discrete time DDIM process defined in the main paper is
\begin{equation}\label{eq: discrete ddim app}
X_{t-1} 
\; = \;
\sqrt{\frac{\alpha_{t-1}}{{\alpha}_t}} X_t 
\, + \,
\beta_t \, s(X_t, t) 
\, + \,
\sigma_t \epsilon_t.
\end{equation}

Up to first order approximation, the discrete time process~\eqref{eq: discrete ddim app} is the Euler-Maruyama discretization of the continuous time process~\eqref{eq: ddim continuous}. A uniform discretization of time is assumed, i.e., $\tau = 1 - \frac{t}{T}$ (See~\cite[Appendix B.1]{domingoenrich2025adjointmatchingfinetuningflow} for the full derivation).

Given random variables $\bar{\mathfrak{X}}_0 \sim \bar{\mathfrak{p}}_0 = \mathcal{N}(0, I)$ and $\bar{\mathfrak{X}}_1 \sim \bar{\mathfrak{p}}_1$, where $\bar{\mathfrak{p}}_1$ is some probability distribution (e.g., the data distribution), we define a reference flow $\bar{\mathfrak{X}}_\tau$ for ${\tau \in [0, 1]}$ as 
\begin{equation}\label{eq: ref flow}
    \bar{\mathfrak{X}}_\tau \;=\; \alpha_\tau \bar{\mathfrak{X}}_0 + \zeta_\tau \bar{\mathfrak{X}}_1.
\end{equation}

Note that there is no specific process implied by the definition above, since different processes can have the same marginal densities as the reference flow at all times $\tau$. We denote by $\bar{\mathfrak{p}}_t(\cdot)$ the density of $\bar{\mathfrak{X}}_\tau$. As $\alpha_\tau$ decreases from $\alpha_0 = 1$ to $\alpha_1 = 0$, and $\zeta_\tau$ increases from $\zeta_0 = 0$ to $\zeta_1 = 1$ the reference flow gives an interpolation between $\bar{\mathfrak{p}}_0 = \mathcal{N}(0, I)$ and $\bar{\mathfrak{p}}_1$.

If the score predictor $s(x,\tau) = \nabla_x \log \bar{\mathfrak{p}}_\tau(x)$, then the DDIM process~\eqref{eq: ddim continuous} has the same marginals as the reference flow~\eqref{eq: ref flow}, i.e.,  ${\mathfrak{p}}_\tau (x, s) = \bar{\mathfrak{p}}_\tau(x)$ for $\tau \in [0, 1]$. This is assuming proper choice of $\alpha_\tau, \zeta_\tau$, i.e.,
$\alpha_\tau = \sqrt{1 - \bar \alpha_\tau}$, 
$\zeta_\tau = \sqrt{\bar \alpha_\tau}$.

\subsubsection{Proof for continuous time}\label{subsec: cont time proof}

We generalize~\cite[Theorem 1]{lyu2012interpretationgeneralizationscorematching} to characterize how the KL divergence between the marginals of two continuous time forward processes changes with time.
\begin{lemma}\label{lem: kl derivative fisher}
 Consider reference flows defined as $\bar{\mathfrak{X}}_\tau = \alpha_\tau \bar{\mathfrak{X}}_0 + \zeta_\tau \bar{\mathfrak{X}}_1$, \,for $\tau \in [0, 1]$ where $\bar{\mathfrak{X}}_0 \sim \mathcal{N}(0, I)$. Denote by $\bar{\mathfrak{p}}_\tau(\cdot)$, the marginal density of $\bar{\mathfrak{X}}_\tau$ when $\bar{\mathfrak{X}}_1 \sim \bar{\mathfrak{p}}_1$ and similarly $\bar{\mathfrak{q}}_\tau(\cdot)$, the marginal density of $\bar{\mathfrak{X}}_\tau$ when $\bar{\mathfrak{X}}_1 \sim \bar{\mathfrak{q}}_1$. The following then holds: 
    \begin{equation}\label{eq: kl derivative}
        \frac{d}{d\tau} D_{\text{\normalfont KL}} (\bar{\mathfrak{p}}_\tau(\cdot)\,\Vert\,\bar{\mathfrak{q}}_\tau(\cdot)) \;=\; -{\gamma_\tau} \dot \gamma_\tau D_{\text{\normalfont F}} (\bar{\mathfrak{p}}_\tau(\cdot) \,\Vert\, \bar{\mathfrak{q}}_\tau(\cdot))
    \end{equation}
    where $\gamma_\tau = \zeta_\tau/\alpha_\tau$, and $D_{\text{\normalfont F}}(p\,\Vert\,q)$ denotes the Fisher divergence.
\end{lemma}

By integrating the derivative of the KL divergence as given by Lemma~\ref{lem: kl derivative fisher}, we obtain the following continuous-time analogue of Lemma~\ref{lem: marginal kl computation}, which characterizes the point-wise KL divergence of two continuous time diffusion processes.
\begin{lemma}\label{lem: marginal kl computation continuous}
    Consider two score predictors $s_{\mathfrak{p}}(x,\tau) = \nabla_x \log \bar{\mathfrak{p}}_\tau(x)$, $s_{\mathfrak{q}}(x,\tau) = \nabla_x \log \bar{\mathfrak{q}}_\tau(x)$, where $\bar {\mathfrak{p}}_\tau$, $\bar {\mathfrak{q}}_\tau$ are marginal densities of two reference flows, with the same noise schedule, starting from initial distributions $\bar {\mathfrak{p}}_0$ and $\bar {\mathfrak{q}}_0$, respectively. Then, the point-wise KL divergence between two distributions of the samples generated by running continuous time DDIM~\eqref{eq: ddim continuous} with $s_{\mathfrak{p}}$ and $s_{\mathfrak{q}}$ is given by
    \begin{equation}\label{eq: marginal kl computation app}
        D_{\text{\normalfont KL}} ({\mathfrak{p}}_0(\cdot; s_{\mathfrak{p}}) \,\Vert\, {\mathfrak{q}}_0(\cdot; s_{\mathfrak{q}})) 
        \; = \;
        \int_{\tau \,=\, 0}^1 \tilde \omega_\tau \, \mathbb{E}_{x\,\sim\, \mathfrak{p}_\tau(\cdot\,; s_p)} \left[ \,
        \norm{s_p(x, \tau) - s_q(x, \tau)}_2^2
        \,\right] 
    \end{equation}
    where $\tilde \omega_\tau$ is a time-dependent constant
\end{lemma}

\subsubsection{Bounding the discretization error}\label{subsec: discretization error}
We now turn to bridging the gap between continuous and discrete times. In~\cite{mou2019improvedboundsdiscretizationlangevin}, they bound this gap which arises from the discretization of the continuous time diffusion process. We will utilize the main result from this paper with a minor modification in that we consider a time-dependent drift term. This is formalized in Lemma~\ref{lem: discrete continuos mou} which allows us bound the KL divergence between the marginals $p_t(\cdot)$ of the discrete time backward DDIM process and the corresponding marginal $\mathfrak{p}_{t/T}(\cdot)$ of the continuous time backward process. 

\begin{lemma}\label{lem: discrete continuos mou}
(Modification of Theorem 1 from~\cite{mou2019improvedboundsdiscretizationlangevin}.) Under mild assumptions on the score function (outlined in the proof), the KL divergence between the marginals of the discrete time backward process $p_t(\cdot)$ and continuous time backward process $\mathfrak{p}_{t/T}(\cdot)$ can be bounded as follows:
    \begin{equation}
        D_{\text{\normalfont KL}} (p_t(\cdot; s_p) \,\Vert\, \mathfrak{p}_{1 - t/T}(\cdot; s_p)) 
        \; \leq \;
        \frac{c}{T^2}
    \end{equation}
where $c$ is a constant depending on the assumptions.
\end{lemma}

Next we need to characterize the sensitivity of the KL divergence to perturbations in the first and second arguments so that we can apply Lemma~\ref{lem: discrete continuos mou}.

\begin{lemma}\label{lem: kl perturbation}
Assume $M := \max_x\left| \log(\frac{\mathfrak{p}_0(\cdot; s_p)}{\mathfrak{p}_0(\cdot; s_q)}) \right|$ is bounded. Then, the point-wise KL between the continuous time processes approximates the point-wise KL between the discrete time processes up to a discretization error $\epsilon_1(T)$:
\begin{equation}
    \left|D_{\text{\normalfont KL}} (\mathfrak{p}_0(\cdot; s_p) \,\Vert\, \mathfrak{q}_0(\cdot; s_q)) - D_{\text{\normalfont KL}} ({p}_0(\cdot; s_p) \,\Vert\, {q}_0(\cdot; s_q))\right| 
    \;\leq\;
    \epsilon_1(T)
\end{equation}
where $\epsilon_1(T) = O(1/T)$.
\end{lemma}

And lastly, we need to characterize the discretization error in going from a integral over continuous time to a sum over discrete time steps.

\begin{lemma}\label{lem: cont to discrete}
Assume $B_1, B_2$ as defined below are finite:
\begin{equation}
    B_1 \;\DefinedAs\; \sup_{x, \tau} \norm{s_p(x, \tau) - s_q(x, \tau)}_2 
\end{equation}
\begin{equation}
    B_2 \;\DefinedAs\; \sup_{x, \tau} \norm{\frac{d}{d \tau}(s_p(x, \tau) - s_q(x, \tau))}_2.
\end{equation}

    Then the integral from Lemma~\ref{lem: marginal kl computation continuous} giving the point-wise KL in continuous time can be approximated with a discrete time sum as follows
        \begin{equation}
        \left| \int_{\tau \,=\, 0}^1 \tilde \omega_\tau \, \mathbb{E}_{x\,\sim\, \mathfrak{p}_\tau(\cdot\,; s_p)} \left[ \,
        \norm{s_p(x, \tau) - s_q(x, \tau)}_2^2
        \,\right]  - \sum_{t \,=\, 0}^T \frac{1}{T} \tilde \omega_{t/T} \, \mathbb{E}_{x\,\sim\, p_t(\cdot\,; s_p)} \left[ \,
        \norm{s_p(x, t) - s_q(x, t)}_2^2
        \,\right]  \right| \; \leq \; \epsilon_2(T)
    \end{equation}
    where the discretization error is $\epsilon_2(T) = O(1/T)$.
\end{lemma}

It remains to combine Lemmas~\ref{lem: kl perturbation} and~\ref{lem: cont to discrete} to complete the proof of Lemma~\ref{lem: marginal kl computation}:
\begin{equation}\label{eq: final final}
    D_{\text{\normalfont KL}} (p_0(\cdot\,; s_p) \,\Vert\, q_0(\cdot\,; s_q)) 
    \; = \;
    \sum_{t \,=\, 0}^T \tilde \omega_t \, \mathbb{E}_{x\,\sim\, p_t(\cdot\,; s_p)} \left[ \,
    \norm{s_p(x, t) - s_q(x, t)}_2^2
    \,\right] 
    \,+\, \epsilon_T
\end{equation}
where $|\epsilon_T| \leq \epsilon_1(T) + \epsilon_2(T) = O(1/T)$. (We abuse notation to denote $\frac{1}{T} \tilde \omega_{t/T}$ as $\tilde \omega_t$ in~\eqref{eq: final final} and in the main paper.)

\subsection{Proof of Theorem~\ref{thm: alignment}}\label{app: proof of alignment}
\begin{proof}
    For any $\lambda\geq 0$, the optimal solution $p^\star(\cdot;\lambda)$ is uniquely determined by solving a partial minimization problem,
    \[
    \minimize_{p\,\in\,\mathcal{P}} \; L_{\text{ALI}}(p,\lambda).
    \]
    Application of Donsker and Varadhan's variational formula yields the optimal solution 
    \[p^\star(\cdot; \lambda) 
    \;
    \propto
    \;  
    q(\cdot) {\rm e}^{\lambda^\top r(\cdot)}.
    \]
    
    Since the strong duality holds for Problem~\eqref{eq: constrained alignment reverse KL}, its optimal solution is given by $p^\star(\cdot;\lambda)$ evaluated at $\lambda = \lambda^\star$. 

    It is straightforward to evaluate the dual function by the definition $D(\lambda) = L(p^\star(\cdot;\lambda), \lambda)$. 
\end{proof}

\subsection{Proof of Theorem~\ref{thm:Strong duality alginment}} 
\label{app:Strong duality alginment}

\begin{proof}
    We first consider the constrained alignment~\eqref{eq: constrained alignment reverse KL diffusion} in the entire path space $\{p_{0:T}(\cdot)\}$. Since the path-wise KL divergence is convex in the path space and the constraints are linear, the strong duality holds in the path space, i.e., there exists a pair $(p^\star_{0:T}(\cdot),\lambda^\star)$ such that
    \[
    \bar{P}^\star_{\text{ALI}} 
    \; \DefinedAs \; D_{\text{KL}}(p_{0:T}^\star(\cdot)\,\Vert\,q_{0:T}(\cdot;s_q))
    \; = \;  
    \bar{D}_{\text{ALI}}(\lambda^\star)
    \; \DefinedAs \; \bar{D}_{\text{ALI}}^\star.
    \]
    Equivalently, $(p^\star_{0:T}(\cdot),\lambda^\star)$ is a saddle point of the Lagrangian $L_{\text{ALI}}(p_{0:T}(\cdot),\lambda)$, 
    \[
    L_{\text{ALI}}(p_{0:T}^\star(\cdot),\lambda)
    \; \leq \;
    L_{\text{ALI}}(p_{0:T}^\star(\cdot),\lambda^\star) 
    \; \leq \;
    L_{\text{ALI}}(p_{0:T}(\cdot),\lambda^\star)
    \; \text{ for all } p_{0:T}(\cdot) \text{ and } \lambda\geq0.
    \]
    Since the score function class $\mathcal{S}$ is expressive enough, any path distribution $p_{0:T}(\cdot)$ can be represented as $p_{0:T}(\cdot; s_p)$ with some $s_p\in\mathcal{S}$; and vice versa. Thus, we can express $p_{0:T}^\star(\cdot)$ as $p_{0:T}(\cdot; s_p^\star)$ with some $s_p^\star\in\mathcal{S}$. We also note that the dual functions $\bar{D}_{\text{ALI}}(\lambda)$ in the path and score function spaces are the same. Hence, the dual value for Problem~\eqref{eq: constrained alignment reverse KL diffusion} remains to be $\bar{D}_{\text{ALI}}(\lambda^\star)$. Thus, $(s_p^\star,\lambda^\star)$ is a saddle point of the Lagrangian $\bar{L}_{\text{ALI}}(s_p,\lambda) \DefinedAs L_{\text{ALI}}(p_{0:T}(\cdot;s_p),\lambda)$, 
    \[
    \bar{L}_{\text{ALI}}(s_p^\star,\lambda)
    \; \leq \;
    \bar{L}_{\text{ALI}}( s_p^\star,\lambda^\star) 
    \; \leq \;
    \bar{L}_{\text{ALI}}(s_p,\lambda^\star)
    \; \text{ for all } s_p \in \mathcal{S} \text{ and } \lambda\geq0.
    \]
    Therefore,
    the strong duality holds for Problem~\eqref{eq: constrained alignment reverse KL diffusion} in the score function space $\mathcal{S}$. 
\end{proof}

\subsection{Proof of Theorem~\ref{thm: reverse KL general}}\label{app:proofs_reverse KL general}

\begin{proof}
    By the definition,
    \[
    \begin{array}{rcl}
         L_{\text{AND}}(p,u; \lambda)
         & = & 
         \displaystyle
         u
        +
        \sum_{i\,=\,1}^m \lambda_i \left(
         D_{\text{KL}}(p\,\Vert\, q^i)
         -
         u
        \right)
         \\[0.2cm]
         & = & 
         \displaystyle
         u
         -
         u\lambda^\top \one 
        +
        \sum_{i\,=\,1}^m \left(
        \lambda_i \mathbb{E}_{x\,\sim\,p} \left[\,\log p(x)\,\right] 
        - \lambda_i \mathbb{E}_{x\,\sim\,p}\left[\,\log q^i(x)\,\right] 
        \right)
        \\[0.2cm]
         & = & \displaystyle
         u
         -
         u\lambda^\top \one 
        + 
        \sum_{i\,=\,1}^m 
        \lambda_i \mathbb{E}_{x\,\sim\,p} \left[\,\log p(x)\,\right] 
        - \mathbb{E}_{x\,\sim\,p}\left[\,\log \prod_{i\,=\,1}^m \left(q^i(x)\right)^{\lambda_i}\,\right] 
        \\[0.2cm]
         & = &
         \displaystyle
         u
         -
         u\lambda^\top \one 
        \\[0.2cm]
         &  & \displaystyle
         + 
        \sum_{i\,=\,1}^m 
        \lambda_i  \left(
        \mathbb{E}_{x\,\sim\,p} \left[\,\log p(x)\,\right] 
        -\mathbb{E}_{x\,\sim\,p}\left[\,\log \prod_{i\,=\,1}^m \left(q^i(x)\right)^{\frac{\lambda_i}{\one^\top \lambda}}\,\right]
        \right)
        \\[0.2cm]
         & = &
        \displaystyle
         u
        +
        \sum_{i\,=\,1}^m \lambda_i \left(
        D_{\text{KL}}(p\,\Vert\, q_{\text{AND}}^{(\lambda)}) -u
        \right)
        - 
        \one^\top \lambda \log Z_{\text{\normalfont 
    AND}}(\lambda).
    \end{array}
\]
By taking $\lambda = \lambda^\star$, we obtain a primal problem: $\maximize_{p\,\in\,\mathcal{P}, u\,\geq\,0} L_{\text{AND}}(p ,u; \lambda^\star)$, which solves the constrained alignment problem~\eqref{eq: constrained alignment reverse KL} because of the strong duality. By the varational optimality, maximization of $L_{\text{AND}}(p,u ; \lambda^\star)$ over $p$ and $u$ is at a unique maximizer, 
    \[
    p^\star (\cdot; \lambda^\star) 
    \; \propto \;
    q_{\text{AND}}^{( \lambda^\star)}(\cdot)   
    \]
    and $u^\star = 0$ if $1 - \one^\top\lambda^\star \geq 0$ and $u^\star = \infty$ otherwise.
    This gives the optimal model $p^\star(\cdot) = p^\star (\cdot; \lambda^\star)$.

    Meanwhile, for any $\lambda\geq0$, the primal problem: $\maximize_{p\,\in\,\mathcal{P}, u\,\geq\,0} L_{\text{AND}}(p, u; \lambda)$ defines the dual function $D_{\text{AND}}(\lambda)$. By the varational optimality, maximization of $L_{\text{AND}}(p, u; \lambda)$ over $p$ and $u$ is at a unique maximizer,  
    \[
    p^\star (\cdot; \lambda,\mu) 
    \; \propto \;
     q_{\text{AND}}^{(\lambda)}(\cdot)
    \]
    and $u^\star(\lambda) = 0$ if $1 - \one^\top\lambda \geq 0$ and $u^\star(\lambda) = \infty$ otherwise.
    This defines the dual function,
    \[
    \begin{array}{rcl}
         D_{\text{AND}}(\lambda) & = &L_{\text{AND}}(p^\star(\cdot;\lambda), u^\star(\lambda); \lambda)
         \\[0.2cm]
         & = &
        \displaystyle
         u^\star(\lambda)
        +
        \sum_{i\,=\,1}^m \lambda_i \left(
        D_{\text{KL}}(p^\star(\cdot;\lambda)\,\Vert\, q_{\text{AND}}^{(\lambda)}(\cdot))
        -
        u^\star(\lambda)\right)
        - 
        \one^\top \lambda \log Z_{\text{\normalfont 
    AND}}(\lambda)
    \\[0.2cm]
         & = &
        \displaystyle
        (1-\one^\top \lambda) u^\star(\lambda)
        - 
        \one^\top \lambda
         \log Z_{\text{\normalfont 
    AND}}(\lambda)
    \end{array}
    \]
    which completes the proof by following the definition of the dual problem and the dual constraint $\one^\top\lambda \leq 1$.
\end{proof}

\subsection{Proof of Theorem~\ref{thm:Strong duality composition}}
\label{app:Strong duality composition}

\begin{proof}
    Similar to the proof of Theorem~\ref{thm:Strong duality alginment}, we can establish a saddle point condition for the Lagrangian $\bar{L}_{\text{AND}}(s_p, u, \lambda)$ by leveraging the expressiveness of the function class $\mathcal{S}$ which represents the path space $\{p_{0:T}(\cdot) \}$. As the proof follows similar steps, we omit the detail.
\end{proof}

\subsection{Proof of Lemma~\ref{lem: primal true score}}
\begin{proof}
From section~\ref{app:proofs_reverse KL general}, we recall
    \begin{equation}\label{eq: 1}
        L_{\text{AND}}(p,u; \lambda) 
        \; = \;
        u
        +
        \sum_{i\,=\,1}^m \lambda_i \left(
        D_{\text{KL}}(p\,\Vert\, q_{\text{AND}}^{(\lambda)}) -u
        \right)
        - 
        \one^\top \lambda \log Z_{\text{\normalfont 
    AND}}(\lambda).
    \end{equation}
Since in the diffusion formulation of the problem~\eqref{eq: constrained alignment reverse KL diffusion} we have $p = p_0(x_0; s)$, $q^i = p_0(x_0; s^i)$, we can derive similarly to~\eqref{eq: 1} that:
\begin{equation}
    L_{\text{AND}}(p_0(\cdot; s),u; \lambda) 
    \; = \;
    u
        +
        \sum_{i\,=\,1}^m \lambda_i \left(
        D_{\text{KL}}(p_0(\cdot; s)\,\Vert\, q_{\text{AND}, 0}^{(\lambda)}(\cdot)) -u
        \right)
        - 
        \one^\top \lambda \log Z_{\text{\normalfont 
    AND}}(\lambda).
\end{equation}
Since minimizing over $u$ would trivially give $\min_u L_{\text{AND}}(p,u; \lambda) = -\infty$ unless $\lambda^\top 1 = 1$, we next consider the Lagrangian in the non-trivial case: $\lambda^\top 1 = 1$. Then we have
\begin{equation}
    L_{\text{AND}}(p_0(\cdot; s); \lambda) 
    \; = \;
    L_{\text{AND}}(s, \lambda) 
    \; = \; 
        D_{\text{KL}}(p_0(\cdot; s)\,\Vert\, q_{\text{AND},0}^{(\lambda)} ) - \log Z_{\text{\normalfont 
    AND}}(\lambda).
\end{equation}
The second term $\log Z_{\text{\normalfont AND}}(\lambda)$ does not depend on the function $s$, thus it suffices to minimize $D_{\text{KL}}(p_0(\cdot; s)\,\Vert\, q_{\text{AND},0}^{(\lambda)} )$ to find the Lagrangian minimizer which we call $s^{(\lambda)}$. The KL is minimized when $p_0(\cdot; s^{(\lambda)}) = q_{\text{AND},0}^{(\lambda)}$. If we have access to samples from $q_{\text{AND},0}^{(\lambda)}$, we can fit a score-like function $s$  to $q_{\text{AND},0}^{(\lambda)}$ by optimizing the denoising score matching objective similar to~\cite[Equation~(1)]{song2021scorebasedgenerativemodelingstochastic}:
\begin{equation}
    L_{\text{sm}}(s, \lambda) 
    \; = \;
    \sum_{t \,=\, 0}^T \omega_t \, \mathbb{E}_{x_0 \,\sim\, q_{\text{\normalfont AND}}^{(\lambda)}(\cdot)} \mathbb{E}_{x_t \,\sim\, q(x_t\,|\,x_0)}\left[ \norm{s(x_t, t) - \nabla \log q(x_t \,|\, x_0)}^2 \right].
\end{equation}
Given sufficient data and function capacity~\cite{song2021scorebasedgenerativemodelingstochastic}, we have $\argmin_s L_{\text{sm}}(s, \lambda) \simeq q_{\text{AND},0}^{(\lambda)}$, which concludes the proof.
\end{proof}

\section{Additional Proofs}
\label{app: Additional Proofs}

We provide detailed proofs for all lemmas in Section~\ref{app: proof of lemma marginal kl}.

\subsection{Proof of Lemma~\ref{lem: kl derivative fisher}}
\begin{proof}
We start by defining $\bar{\mathfrak{Y}}_\tau$ as a time-dependent scaling of $\bar{\mathfrak{X}}_\tau$:
\begin{equation}
    \bar{\mathfrak{Y}}_\tau \;\DefinedAs\; \frac{1}{\alpha_\tau} \bar{\mathfrak{X}}_\tau \; =\; \bar{\mathfrak{X}}_1 + \gamma_\tau \bar{\mathfrak{X}}_0
\end{equation}
where $\gamma_\tau \DefinedAs \zeta_\tau/\alpha_\tau$. Denote by $\tilde{\mathfrak{p}}_{\tau}(\bar{\mathfrak{Y}}_\tau)$, the marginal density of $\bar{\mathfrak{Y}}_\tau$ when $\bar{\mathfrak{X}}_1\sim {\mathfrak{p}}_{1}$ and similarly $\tilde q_{t}(\bar{\mathfrak{Y}}_\tau)$, the marginal density of $\bar{\mathfrak{Y}}_\tau$ when ${\mathfrak{X}}_1\sim \mathfrak{q}_{1}$. Now we generalize Theorem 1 from~\cite{lyu2012interpretationgeneralizationscorematching} to show that~\eqref{eq: kl derivative} holds for $\tilde{\mathfrak{p}}_{\tau}, \tilde{\mathfrak{q}}_{\tau}$. Their Theorem is for the specific case of $\gamma_\tau = \sqrt{1 - t}$.\footnote{Just to avoid any confusion, in~\cite{lyu2012interpretationgeneralizationscorematching}, at $t = 0$ we have the data distribution and as $t$ increases the distributions converge to Gaussians. However in the current paper, the direction of time is the opposite, meaning $t = 0$ corresponds to the pure Gaussians and at $t = 1$ we have the data distributions.} 

We now present Lemmas~\ref{lem: heat equation} and~\ref{lem: vector calc identity} which we will need in the remainder of the proof.

\begin{lemma}\label{lem: heat equation}
    For density \( \tilde{\mathfrak{p}}_{\tau}(\bar{\mathfrak{Y}}_\tau) \) as defined in Theorem 1, the following identity holds:
\begin{equation}
    \frac{d}{dt} \tilde{\mathfrak{p}}_{\tau}(\bar{\mathfrak{Y}}_\tau) = \gamma_\tau \dot \gamma_\tau \Delta_{\bar{\mathfrak{Y}}_\tau} \tilde{\mathfrak{p}}_{\tau}(\bar{\mathfrak{Y}}_\tau).
\end{equation}
\end{lemma}
\begin{proof}
    We start with \( \tilde{\mathfrak{p}}_{\tau}(\bar{\mathfrak{Y}}_\tau) \) which is the convolution of a Gaussian distribution with $\mathfrak{p}_1(\bar{\mathfrak{X}}_1)$:

\begin{equation}
    \tilde{\mathfrak{p}}_{\tau}(\bar{\mathfrak{Y}}_\tau) \;=\; \int_{\bar{\mathfrak{X}}_1} \frac{1}{(2\pi \gamma_\tau^2)^{d/2}} \exp \left( -\frac{\norm{\bar{\mathfrak{Y}}_\tau - \bar{\mathfrak{X}}_1}^2}{2\gamma_\tau^2} \right) \mathfrak{p}_1(\bar{\mathfrak{X}}_1).
\end{equation}
Taking the derivative, we have
\begin{equation}\label{eq: derivative}
\begin{array}{rcl}
      \displaystyle \frac{d}{dt} \tilde{\mathfrak{p}}_{\tau}(\bar{\mathfrak{Y}}_\tau)&=&  \displaystyle \int_{\bar{\mathfrak{X}}_1} \frac{ \dot \gamma_\tau \norm{\bar{\mathfrak{Y}}_\tau - \bar{\mathfrak{X}}_1}^2}{\gamma_\tau^3} \frac{1}{(2\pi \gamma_\tau^2)^{d/2}} \exp \left( -\frac{\norm{\bar{\mathfrak{Y}}_\tau - \bar{\mathfrak{X}}_1}^2}{2\gamma_\tau^2} \right) \mathfrak{p}_1(\bar{\mathfrak{X}}_1)\\
     & & \displaystyle -\int_{\bar{\mathfrak{X}}_1} \frac{d}{\gamma_\tau} \frac{\dot \gamma_\tau }{(2\pi \gamma_\tau^2)^{d/2}} \exp \left( -\frac{\norm{\bar{\mathfrak{Y}}_\tau - \bar{\mathfrak{X}}_1}^2}{2\gamma_\tau^2} \right) \mathfrak{p}_1(\bar{\mathfrak{X}}_1).
\end{array}
\end{equation}

On the other hand, taking the gradient of $\tilde{\mathfrak{p}}_{\tau}(\bar{\mathfrak{Y}}_\tau)$ with respect to $\bar{\mathfrak{Y}}_\tau$ we get
\begin{equation}
    \nabla_{\bar{\mathfrak{Y}}_\tau} \tilde{\mathfrak{p}}_{\tau}(\bar{\mathfrak{Y}}_\tau) \;=\; -\int_{\bar{\mathfrak{X}}_1} \frac{\bar{\mathfrak{Y}}_\tau - \bar{\mathfrak{X}}_1}{\gamma_\tau^2} \frac{1}{(2\pi \gamma_\tau^2)^{d/2}} \exp \left( -\frac{\norm{\bar{\mathfrak{Y}}_\tau - \bar{\mathfrak{X}}_1}^2}{2\gamma_\tau^2} \right) \mathfrak{p}_1(\bar{\mathfrak{X}}_1).
\end{equation}

Taking the divergence of the gradient, we have
\begin{equation}\label{eq: laplacian}
    \begin{array}{rcl}
  \displaystyle  \Delta_{\bar{\mathfrak{Y}}_\tau} \tilde{\mathfrak{p}}_{\tau}(\bar{\mathfrak{Y}}_\tau) & = & \displaystyle \int_{\bar{\mathfrak{X}}_1} \frac{\norm{\bar{\mathfrak{Y}}_\tau - \bar{\mathfrak{X}}_1}^2}{\gamma_\tau^4} \frac{1}{(2\pi \gamma_\tau^2)^{d/2}} \exp \left( -\frac{\norm{\bar{\mathfrak{Y}}_\tau - \bar{\mathfrak{X}}_1}^2}{2\gamma_\tau^2} \right) \mathfrak{p}_1(\bar{\mathfrak{X}}_1) \\[0.3 cm]
     & & \displaystyle -\int_{\bar{\mathfrak{X}}_1} \frac{d}{\gamma_\tau^2} \frac{1}{(2\pi \gamma_\tau^2)^{d/2}} \exp \left( -\frac{\norm{\bar{\mathfrak{Y}}_\tau - \bar{\mathfrak{X}}_1}^2}{2\gamma_\tau^2} \right) \mathfrak{p}_1(\bar{\mathfrak{X}}_1).
\end{array}
\end{equation}

Comparing Equations~\eqref{eq: derivative} and~\eqref{eq: laplacian} proves the result.

\end{proof}

\begin{lemma}\label{lem: vector calc identity}
    For any positive valued function $f(x): \mathbb{R}^d \rightarrow \mathbb{R}$ whose gradient $\nabla_{x} f$ and Laplacian $\Delta_{x} f$ are well defined, we have the identity:

\begin{equation}
    \frac{\Delta_{x} f(x)}{f(x)} \;=\; \Delta_{x} \log f(x) + \norm{\nabla_{x} \log f(x)}^2.
\end{equation}

\end{lemma}

We now continue with the proof of Lemma~\ref{lem: kl derivative fisher}. We start with the definition of Fisher divergence for generic distributions $p, q$:
\begin{equation}\label{eq: fisher expansion}
\begin{array}{rcl}
    \displaystyle D_{\text{F}}(p \,\Vert\, q) & = &  \displaystyle\int_x p(x) \norm{ \nabla \log p(x) - \nabla \log q(x)}^2 dx \\[0.3 cm]
     &=  & \displaystyle\int_x p(x) \norm{ \frac{\nabla p(x)}{p(x)} - \frac{\nabla q(x)}{q(x)}}^2 dx \\[0.3 cm]
     &=  & \displaystyle\int_x p(x) \left( \norm{ \frac{\nabla p(x)}{p(x)}}^2 +  \norm{\frac{\nabla q(x)}{q(x)}}^2  - 2 \frac{\nabla p(x)^\top \nabla q(x)}{p(x) q(x)} \right) dx. 
\end{array}
\end{equation}

We apply integration by parts to the third term. For any open bounded subset $\Omega$ of $\mathbb{R}^d$ with a piecewise smooth boundary $\Gamma = \partial \Omega$, we have
\begin{equation}\label{eq: integration by parts}
\begin{array}{rcl}
    \displaystyle \int_{x \,\in\, \Omega} \nabla p(x)^\top \frac{ \nabla q(x)}{q(x)} dx & = & \displaystyle \int_{x \,\in\, \Omega} \nabla p(x)^\top (\nabla \log q(x)) dx \\[0.3 cm]
     & = & \displaystyle -\int_{x \,\in \,\Omega} p(x) \Delta \log q(x) dx + \int_\Gamma p(x) (\nabla \log q(x) ^\top \hat n) d\Gamma.
\end{array}
\end{equation}

Assuming that both $p(x)$ and $q(x)$ are smooth and fast-decaying, the boundary term in~\eqref{eq: integration by parts} vanishes. 

Then we can combine~\eqref{eq: fisher expansion} and~\eqref{eq: integration by parts} to write
\begin{equation}\label{eq: fisher final}
     D_{\text{F}}(p \,\Vert\, q) \; = \; \int_x p(x) \left( \norm{ \nabla \log p(x)}^2 +  \norm{ \nabla \log q(x) }^2  + 2 \Delta_x \log q(x) \right) dx.
\end{equation}

Returning to our distributions $\tilde{\mathfrak{p}}_{\tau}(\bar{\mathfrak{Y}}_\tau)$ and  $\tilde{\mathfrak{q}}_{\tau}(\bar{\mathfrak{Y}}_\tau)$ we can rewrite~\eqref{eq: fisher final} as
\begin{equation}\label{eq: fisher final final}
    D_{\text{F}}(\tilde{\mathfrak{p}}_{\tau}(\cdot)\,\Vert\,\tilde{\mathfrak{q}}_{\tau}(\cdot)) \;=\; \int_{\mathfrak{Y}_\tau} \tilde{\mathfrak{p}}_\tau(\mathfrak{Y}_\tau) \left( \norm{ \nabla \log \tilde{\mathfrak{p}}_\tau(\mathfrak{Y}_\tau)}^2 +  \norm{ \nabla \log \tilde{\mathfrak{q}}_{\tau}(\mathfrak{Y}_\tau) }^2  + 2 \Delta_{\mathfrak{Y}_\tau}\log \tilde{\mathfrak{q}}_{\tau}(\mathfrak{Y}_\tau) \right) d\mathfrak{Y}_\tau.
\end{equation}

For conciseness in notation, we drop references to variables $\bar{\mathfrak{Y}}_\tau$ and $\bar{\mathfrak{X}}_\tau$ in the integration, the density functions, and the operators whenever this does not lead to ambiguity. We start by applying Lemma~\ref{lem: vector calc identity} to Equation~\eqref{eq: fisher final}:
\begin{equation}\label{eq: tag10}
    \begin{array}{rcl}
      \displaystyle D_{\text{F}}(\tilde{\mathfrak{p}}_\tau \,\Vert\, \tilde{\mathfrak{q}}_{\tau})   
      & = & \displaystyle\int \tilde{\mathfrak{p}}_\tau \left( |\nabla \log \tilde{\mathfrak{p}}_\tau|^2 + |\nabla \log \tilde{\mathfrak{q}}_{\tau}|^2 + 2 \Delta \log \tilde{\mathfrak{q}}_{\tau} \right)
      \\[0.3 cm]
         & = & \displaystyle\int \tilde{\mathfrak{p}}_\tau \left( |\nabla \log \tilde{\mathfrak{p}}_\tau|^2 + \frac{\Delta \tilde{\mathfrak{q}}_{\tau}}{\tilde{\mathfrak{q}}_{\tau}} + \Delta \log \tilde{\mathfrak{q}}_{\tau} \right).
    \end{array}
\end{equation}

Next, we expand the derivative of the KL divergence:
\[
\frac{d}{d\tau} D_{\text{KL}}(\tilde{\mathfrak{p}}_\tau \,\Vert\, \tilde{\mathfrak{q}}_{\tau}) 
\;=\; 
\int \frac{d}{d\tau} \tilde{\mathfrak{p}}_\tau \log \frac{\tilde{\mathfrak{p}}_\tau}{\tilde{\mathfrak{q}}_{\tau}} + \int \tilde{\mathfrak{p}}_\tau \frac{d}{d\tau} \log \tilde{\mathfrak{p}}_\tau - \int \tilde{\mathfrak{p}}_\tau \frac{d}{d\tau} \log \tilde{\mathfrak{q}}_{\tau}.
\]

We can eliminate the second term by exchanging integration and differentiation of \( \tau \):
\[
\int \tilde{\mathfrak{p}}_\tau \frac{d}{d\tau} \log \tilde{\mathfrak{p}}_\tau 
\;=\; 
\int \frac{d \tilde{\mathfrak{p}}_\tau}{d\tau} 
\;=\; 
\frac{d}{d\tau} \int \tilde{\mathfrak{p}}_\tau 
\;=\; 
0.
\]

As a result, there are three remaining terms in computing \( \frac{d}{d\tau} D_{KL}(\tilde{\mathfrak{p}}_\tau \| \tilde{\mathfrak{q}}_{\tau}) \), which we can further substitute using Lemma~\ref{lem: heat equation}, as

\begin{equation}\label{eq: tag11}
    \begin{array}{rcl}
      \displaystyle\frac{d}{d\tau} D_{\text{KL}}(\tilde{\mathfrak{p}}_\tau \,\Vert\, \tilde{\mathfrak{q}}_{\tau})   & = &  \displaystyle\int \frac{d}{d\tau} \tilde{\mathfrak{p}}_\tau \log \tilde{\mathfrak{p}}_\tau - \int \frac{d}{d\tau} \tilde{\mathfrak{p}}_\tau \log \tilde{\mathfrak{q}}_{\tau} - \int \tilde{\mathfrak{p}}_\tau \frac{d}{d\tau} \log \tilde{\mathfrak{q}}_{\tau}
      \\[0.3 cm]
         & = & \displaystyle \gamma_\tau \dot \gamma_\tau \left( \int \Delta \tilde{\mathfrak{p}}_\tau \log \tilde{\mathfrak{p}}_\tau - \int \Delta \tilde{\mathfrak{p}}_\tau \log \tilde{\mathfrak{q}}_{\tau} - \int \tilde{\mathfrak{p}}_\tau \frac{\Delta \tilde{\mathfrak{q}}_{\tau}}{\tilde{\mathfrak{q}}_{\tau}} \right).
    \end{array}
\end{equation}

Using integration by parts, the first term in~\eqref{eq: tag11} is changed to
\[
\int \Delta \tilde{\mathfrak{p}}_\tau \log \tilde{\mathfrak{p}}_\tau \;=\; \sum_{i\,=\,1}^d \frac{\partial \tilde{\mathfrak{p}}_\tau}{\partial y_i} \log \tilde{\mathfrak{p}}_\tau(\vec{y}) \bigg|_{y_i = \infty}^{y_i = -\infty} - \int \nabla \tilde{\mathfrak{p}}_\tau^T \nabla \log \tilde{\mathfrak{p}}_\tau.
\]

The limits in the first term become zero given the smoothness and fast decay properties of \( \tilde{\mathfrak{p}}_\tau(\vec{y}) \). The remaining term can be further simplified as
\[
\int \nabla \tilde{\mathfrak{p}}_\tau^T \nabla \log \tilde{\mathfrak{p}}_\tau \;=\; \int \tilde{\mathfrak{p}}_\tau (\nabla \log \tilde{\mathfrak{p}}_\tau)^T \nabla \log \tilde{\mathfrak{p}}_\tau \;=\; \int \tilde{\mathfrak{p}}_\tau |\nabla \log \tilde{\mathfrak{p}}_\tau|^2.
\]

The second term in~\eqref{eq: tag11} can be manipulated similarly, by first using integration by parts to get
\[
\int \Delta \tilde{\mathfrak{p}}_\tau \log \tilde{\mathfrak{q}}_{\tau} \;=\; \sum_{i\,=\,1}^d \frac{\partial \tilde{\mathfrak{p}}_\tau}{\partial y_i} \log \tilde{\mathfrak{q}}_{\tau} \bigg|_{y_i = \infty}^{y_i = -\infty} - \int \nabla \tilde{\mathfrak{p}}_\tau^T \nabla \log \tilde{\mathfrak{q}}_{\tau}.
\]

Applying integration by parts again to \( \nabla \tilde{\mathfrak{p}}_\tau^T \nabla \log \tilde{\mathfrak{q}}_{\tau} \), we have
\[
\int \nabla \tilde{\mathfrak{p}}_\tau^T \nabla \log \tilde{\mathfrak{q}}_{\tau} \;=\; \sum_{i\,=\,1}^d \tilde{\mathfrak{p}}_\tau \frac{\partial \log \tilde{\mathfrak{q}}_{\tau}}{\partial y_i} \bigg|_{y_i = \infty}^{y_i = -\infty} - \int \tilde{\mathfrak{p}}_\tau \Delta \log \tilde{\mathfrak{q}}_{\tau}.
\]

The limits at the boundary values are all zero due to the smoothness and fast decay properties of \( \tilde{\mathfrak{p}}_\tau(\vec{y}) \). Now collecting all terms, we have \( \int \tilde{\mathfrak{p}}_\tau \log \tilde{\mathfrak{p}}_\tau \;=\; - \int \tilde{\mathfrak{p}}_\tau |\nabla \log \tilde{\mathfrak{p}}_\tau|^2 \) and \( \int \tilde{\mathfrak{p}}_\tau \log \tilde{\mathfrak{q}}_{\tau} = \int \tilde{\mathfrak{p}}_\tau \Delta \log \tilde{\mathfrak{q}}_{\tau} \). Thus~\eqref{eq: tag11} becomes
\[
\frac{d}{d\tau} D_{\text{KL}}(\tilde{\mathfrak{p}}_\tau \,\Vert\, \tilde{\mathfrak{q}}_{\tau}) \;=\; -\gamma_\tau \dot \gamma_\tau \int \tilde{\mathfrak{p}}_\tau \left( |\nabla \log \tilde{\mathfrak{p}}_\tau|^2 + \Delta \log \tilde{\mathfrak{q}}_{\tau} + \frac{\Delta \tilde{\mathfrak{q}}_{\tau}}{\tilde{\mathfrak{q}}_{\tau}} \right).
\]

Combining with~\eqref{eq: tag10}, this leads to 
\begin{equation}\label{eq: semifinal}
    \frac{d}{d\tau} D_{\text{\normalfont KL}} (\tilde{\mathfrak{p}}_\tau \,\Vert\, \tilde{\mathfrak{q}}_{\tau}) \;=\; - {\gamma_\tau} \dot \gamma_\tau D_{\text{\normalfont F}} (\tilde{\mathfrak{p}}_\tau \,\Vert\, \tilde{\mathfrak{q}}_{\tau}).
\end{equation}

Recall that $\tilde{\mathfrak{p}}_{\tau}(\cdot)$ and $\tilde{\mathfrak{q}}_{\tau}(\cdot)$ were the densities of the scaled random variable $\bar{\mathfrak{Y}}_\tau = \frac{1}{\alpha_\tau}\bar{\mathfrak{X}}_\tau$. This leads to ${\mathfrak{p}}_{\tau} (\bar{\mathfrak{X}}_\tau) d\bar{\mathfrak{X}}_\tau = \tilde{\mathfrak{p}}_{\tau} (\bar{\mathfrak{Y}}_\tau) d\bar{\mathfrak{Y}}_\tau$. Thus, it is straightforward to show that both KL divergence and Fisher divergence are invariant to the scaling of the underlying random variables. For KL divergence we have
\begin{equation}
    D_{\text{\normalfont KL}} (\tilde{\mathfrak{p}}_{\tau}(\cdot) \,\Vert\, \tilde{\mathfrak{q}}_{\tau}(\cdot)) = \int \tilde{\mathfrak{p}}_{\tau} (\bar{\mathfrak{Y}}_\tau) \log \frac{\tilde{\mathfrak{p}}_{\tau} (\bar{\mathfrak{Y}}_\tau)}{\tilde{\mathfrak{q}}_{\tau} (\bar{\mathfrak{Y}}_\tau)} d\bar{\mathfrak{Y}}_\tau = \int {\mathfrak{p}}_{\tau} (\bar{\mathfrak{X}}_\tau) \log \frac{{\mathfrak{p}}_{\tau} (\bar{\mathfrak{X}}_\tau)}{{\mathfrak{q}}_{\tau} (\bar{\mathfrak{X}}_\tau)} d\bar{\mathfrak{X}}_\tau = D_{\text{\normalfont KL}} ({\mathfrak{p}}_{\tau}(\cdot) \,\Vert\,{\mathfrak{q}}_{\tau}(\cdot)).
\end{equation}
And for Fisher Divergence, we can write
\begin{equation}
        \begin{array}{rcl}
         \displaystyle D_{\text{\normalfont F}} (\tilde{\mathfrak{p}}_{\tau}(\cdot) \,\Vert\, \tilde{\mathfrak{q}}_{\tau}(\cdot)) & = & \displaystyle\int  \tilde{\mathfrak{p}}_{\tau} (\bar{\mathfrak{Y}}_\tau) \norm{\frac{\nabla \tilde{\mathfrak{p}}_{\tau} (\bar{\mathfrak{Y}}_\tau)}{\tilde{\mathfrak{p}}_{\tau}(\bar{\mathfrak{Y}}_\tau)} - \frac{\nabla \tilde{\mathfrak{q}}_{\tau}(\bar{\mathfrak{Y}}_\tau)}{\tilde{\mathfrak{q}}_{\tau}(\bar{\mathfrak{Y}}_\tau)}} ^2 d\bar{\mathfrak{Y}}_\tau \\[0.3cm]
         & = & \displaystyle \int  {\mathfrak{p}}_{\tau} (\bar{\mathfrak{X}}_\tau) \norm{\frac{\nabla {\mathfrak{p}}_{\tau} (\bar{\mathfrak{X}}_\tau)}{{\mathfrak{p}}_{\tau}(\bar{\mathfrak{X}}_\tau)} - \frac{\nabla {\mathfrak{q}}_{\tau}(\bar{\mathfrak{X}}_\tau)}{{\mathfrak{q}}_{\tau}(\bar{\mathfrak{X}}_\tau)}} ^2 d\bar{\mathfrak{X}}_\tau\\[0.3 cm]
         & = & \displaystyle D_{\text{\normalfont F}} ({\mathfrak{p}}_{\tau}(\cdot) \,\Vert\, {\mathfrak{q}}_{\tau}(\cdot)).
    \end{array}
\end{equation}

Thus we can replace the divergences in~\eqref{eq: semifinal} with those of the non-scaled distribution, which concludes the proof.
\end{proof}

\subsection{Proof of Lemma~\ref{lem: marginal kl computation continuous}}
\begin{proof}
    We start with a direct application of Lemma~\ref{lem: kl derivative fisher}:
    \begin{equation}
        \begin{array}{rcl}
         \displaystyle D_{\text{\normalfont KL}} ({\mathfrak{p}}_1(\cdot) \,\Vert\, {\mathfrak{q}}_1(\cdot)) & = & \displaystyle D_{\text{\normalfont KL}} ({\mathfrak{p}}_0(\cdot) \,\Vert\, {\mathfrak{q}}_0(\cdot)) - \int_{\tau \,=\, 0}^1 \dot{\gamma_\tau} \gamma_\tau D_{\text{\normalfont F}} (\tilde{\mathfrak{p}}_\tau(\cdot) \,\Vert\, \tilde{\mathfrak{q}}_\tau(\cdot)) d\tau\\[0.3cm]
         & = & \displaystyle - \int_{\tau \,=\, 0}^1 \dot{\gamma_\tau} \gamma_\tau \mathbb{E}_{x \,\sim\, \tilde{\mathfrak{p}}_\tau} \left[  \norm{\nabla \log \tilde{\mathfrak{p}}_\tau(x) - \nabla \log \tilde{\mathfrak{q}}_\tau(x)}_2^2\right] d\tau\\[0.3 cm]
         & = & \displaystyle -\int_{\tau \,=\, 0}^1 \dot{\gamma_\tau} \gamma_\tau \mathbb{E}_{x \,\sim\, \tilde{\mathfrak{p}}_\tau} \left[  \norm{s_p(x, \tau) - s_q(x, \tau)}_2^2\right] d\tau\\[0.3 cm]
         & = & \displaystyle \int_{\tau \,=\, 0}^1 \frac{\dot {\alpha_\tau}}{\alpha_\tau^3} \mathbb{E}_{x \,\sim\, \tilde{\mathfrak{p}}_\tau} \left[  \norm{s_p(x, \tau) - s_q(x, \tau)}_2^2\right] d\tau.
    \end{array}
    \end{equation}

    In the second line we used the fact that ${\mathfrak{p}}_0(\cdot) = {\mathfrak{q}}_0(\cdot) = \mathcal{N}(0, I)$, therefore $D_{\text{\normalfont KL}} ({\mathfrak{p}}_0(\cdot) || {\mathfrak{q}}_0(\cdot)) = 0$. The third line follows from our definition of the score functions. Finally, in the last line we used the fact that $\dot{\gamma_\tau} \gamma_\tau = -\frac{\dot {\alpha_\tau}}{\alpha_\tau^3}$ which follows from $\gamma_\tau = \zeta_\tau/\alpha_\tau$ and $\alpha_\tau^2 + \zeta_\tau^2 = 1$:
    \begin{equation}
        \begin{array}{rcl}
            \displaystyle \dot{\gamma_\tau} \gamma_\tau & = & \displaystyle\frac{d}{d\tau}{(\frac{\zeta_\tau}{\alpha_\tau})} \frac{\zeta_\tau}{\alpha_\tau} \\[0.2cm]
             &= & \displaystyle\frac{\dot \zeta_\tau \zeta_\tau \alpha_\tau - \dot \alpha_\tau \zeta_\tau^2}{\alpha_\tau^3}\\[0.2cm]
             & = & \displaystyle\frac{-\dot \alpha_\tau \alpha_\tau^2 - \dot \alpha_\tau (1 - \alpha_\tau^2)}{\alpha_\tau^3} \\[0.2cm]
             & = & -\displaystyle\frac{\dot {\alpha_\tau}}{\alpha_\tau^3}\\[0.2cm]
        \end{array}
    \end{equation}
    by denoting $\tilde \omega_\tau := -\frac{\dot {\alpha_\tau}}{\alpha_\tau^3}$ we conclude the proof.
\end{proof}

\subsection{Proof of Lemma~\ref{lem: discrete continuos mou}}

\begin{proof}
In~\cite{mou2019improvedboundsdiscretizationlangevin}, they prove this result assuming a drift term that only depends on $x$. For our modification, we begin by defining the time-dependent drift  $b_\tau:\mathbb{R}^d \rightarrow \mathbb{R}^d$ of the diffusion process~\eqref{eq: ddim continuous} as
    \[
        b_\tau(x) 
        \; \DefinedAs \;
        \left(\frac{\dot{\alpha}_\tau}{2 \alpha_\tau}x + \left(\frac{\dot{\alpha}_\tau}{2 \alpha_\tau} + \frac{\sigma_\tau^2}{2}\right)s(x, \tau)\right).
    \]
    
\begin{assumption}\label{as: drift properties}
 The drift $b_\tau (\cdot)$ satisfies the following properties for all times $\tau \in [0,1]$,
 \begin{enumerate}
     \item \normalfont{\textbf{Lipschitz drift term.}}There is a finite constant $L_1$ such that
     \begin{equation}
         \|b_\tau(x) - b_\tau(y)\|_2 
         \; \leq \;
         L_1\, \|x - y\|_2 \quad \text{for all } x, y \in \mathbb{R}^d.
     \end{equation}
  \item  \normalfont{\textbf{Smooth drift term.}} There is a finite constant $L_2$ such that
\begin{equation}
    \|\nabla b_\tau(x) - \nabla b_\tau(y)\|_\infty 
    \; \leq \;
    L_2\, \|x - y\|_2 \quad \text{for all } x, y \in \mathbb{R}^d.
\end{equation}

\item  \normalfont{\textbf{Distant dissipativity.}} There exist strictly positive constants $\mu, \beta$ such that
\begin{equation}
    \langle b_\tau(x), x \rangle 
    \; \leq \;
    -\,\mu\, \|x\|_2^2 \,+\, \beta \quad \text{for all } x \in \mathbb{R}^d.
\end{equation}

\item  \normalfont{\textbf{Time-continuous drift term.}} There is a finite constant $L_3$ such that
\begin{equation}
    \norm{\frac{\partial b_\tau (x)}{\partial \tau}}_2 
    \; \leq \;
    L_3 \quad \text{for all } x \in \mathbb{R}^d.
\end{equation}

\end{enumerate}
    
\end{assumption}

There is an additional assumption in~\cite{mou2019improvedboundsdiscretizationlangevin} on the smoothness of the initial densities of the continuous and discrete processes. In our case both are the standard Gaussian which satisfies the assumption.
    We do not provide the whole proof here as it would consist of almost the entirety of~\cite{mou2019improvedboundsdiscretizationlangevin}. We focus on a small part of the proof, that is the only part that changes when we use a time-dependent drift $b_\tau(x)$ as opposed to~\cite{mou2019improvedboundsdiscretizationlangevin} where they assume a time-independent drift $b(x)$.

    Consistent with their notation, we define a continuous time diffusion process as
    \begin{equation}\label{eq: mou continuous}
        dX_\tau 
        \; = \;
        b_\tau(X_\tau) d\tau \, + \, dB_\tau
    \end{equation}
    and its Euler-Maruyama discretization parameterized by step size $\eta > 0$ (In our case $\eta = \frac{1}{T}$):
    \begin{equation}\label{eq: mou discrete}
        \hat X_{(k + 1)\eta} \; = \; 
        \hat X_{k \eta} \, + \, \eta b_{k\eta}(\hat X_{k\eta}) \, + \, \sqrt{\eta} \xi_k, \quad \quad \xi_k \sim \mathcal{N}(0, I).
    \end{equation}
    Furthermore, they construct a continuous time stochastic process over the interval $\tau \in [\eta, (k + 1)\eta]$ that interpolates~\eqref{eq: mou discrete}:
    \begin{equation}\label{eq: mou interpolation}
        \hat X_\tau 
        \; \DefinedAs \;
        \hat X_{k \eta} 
        \, + \, \int_0^{\tau - k\eta} b_{k\eta}(\hat X_{k\eta}) ds \, + \, \int_{k\eta}^\tau d\hat B_s.
    \end{equation}

    Then they prove that the densities of the two continuous time processes given by~\eqref{eq: mou continuous},\eqref{eq: mou interpolation} denoted as $\pi_\tau$ and $\hat \pi_\tau$ respectively satisfy the following (Lemma 2 from~\cite{mou2019improvedboundsdiscretizationlangevin}):
    \begin{equation}
        \frac{d}{d\tau} D_{\text{\normalfont KL}}(\hat \pi_\tau \,\Vert\, \pi_\tau) \; \leq \;
        \frac{1}{2} \int_{\mathbb{R}^d} \hat \pi_\tau(x) \norm{\hat b_\tau (x) - b_\tau(x)}_2^2 dx.
    \end{equation}
    where $\hat b_\tau(x) \DefinedAs \mathbb{E} \left[ b_{k \eta} (\hat X_{k \eta}) | \hat X_\tau = x \right]$ where the expectation is over the process~\eqref{eq: mou interpolation}. Then they proceed to bound the norm inside the integral. The next equation based on equation (18) from~\cite{mou2019improvedboundsdiscretizationlangevin} is where the time dependence of the drift term in our case enters the picture:
    \begin{equation}\label{eq: mou time dependence effect}
    \begin{array}{rcl}
         \hat b_\tau(x) - b_\tau(x)  & = & \displaystyle \mathbb{E}\left[ b_{k \eta}(\hat X_{k \eta}) \,|\, \hat X_\tau = x \right] - b_\tau(x)
         \\[0.4cm]
         & = & \displaystyle
         \mathbb{E}\left[ b_{k \eta}(\hat X_{k \eta}) | \hat X_\tau = x \right] - \left( b_{k \eta} (x) + \frac{\partial b_\tau(x)}{\partial \tau} (\tau - k\eta) + O((\tau - k\eta)^2)\right)
         \\[0.4cm]
         & = & \displaystyle
         \mathbb{E}\left[ b_{k \eta}(\hat X_{k \eta}) - b_{k \eta}(\hat X_\tau) | \hat X_\tau = x \right] - \frac{\partial b_\tau(x)}{\partial \tau}\bigg|_{\tau \,=\, k\eta} (\tau - k\eta) + O((\tau - k\eta)^2).
    \end{array}
    \end{equation}

They prove that the first term in~\eqref{eq: mou time dependence effect} is $O(\eta)$ and from our additional time-continuity requirement for the drift in Assumption~\ref{as: drift properties}, the second term is also $O(\eta)$. (Note that $\tau \in [k \eta, (k + 1) \eta]$ thus $(\tau - k \eta)$ can be at most $\eta$). With this, the rest of the proof from~\cite{mou2019improvedboundsdiscretizationlangevin} goes through.
\end{proof}

\subsection{Proof of Lemma~\ref{lem: kl perturbation}}

\begin{proof}

We first prove a similar relation for generic distributions $\pi(x), \rho(x)$ and their perturbations $\hat \pi(x), \hat{\rho} (x)$.

Whenever it is clear from the context, we omit the integration variables. Perturbing the first argument leads to
\begin{equation}\label{eq: perturb 1}
    \begin{aligned}
    |D_{\text{\normalfont KL}}(\hat \pi \,\Vert\, \rho) - D_{\text{\normalfont KL}}(\pi \,\Vert\, \rho)| 
        & \; = \; \int \hat \pi \log\left(\frac{\hat \pi}{\rho}\right) - \int \pi \log\left(\frac{\pi}{\rho}\right) + \int ( \hat \pi \log \pi - \hat \pi \log \pi) \\[0.2cm]
    & \; = \; D_{\text{\normalfont KL}}(\hat \pi \,\Vert\, \pi) + \int (\hat \pi - \pi) \log\left(\frac{\pi}{\rho}\right) \\[0.2cm]
    & \; \leq \; D_{\text{\normalfont KL}}(\hat \pi \,\Vert\, \pi) + \max\left(\left| \log\left(\frac{\pi}{\rho}\right) \right|\right) \int |\hat \pi - \pi|  \\[0.2cm]
    & \; = \; D_{\text{\normalfont KL}}(\hat \pi \,\Vert\, \pi) + 2\log M \, d_{\text{\normalfont TV}}(\hat \pi, \pi)
\end{aligned}
\end{equation}

where $\log M := \max_x\left| \log(\frac{\pi(x)}{\rho(x)}) \right|$ and $d_{\text{\normalfont TV}}$ denotes the total variation distance between distributions. Next, perturbing the second argument, we get
\begin{equation}\label{eq: perturb 2}
    \begin{aligned}
    \left| D_{\text{\normalfont KL}}(\hat \pi \,\Vert\, \hat \rho) - D_{\text{\normalfont KL}}(\hat \pi \,\Vert\, \rho) \right| 
        &\; = \; \left| \int \hat \pi \log\left(\frac{\hat \pi}{\hat \rho}\right) - \int \hat \pi \log\left(\frac{\hat \pi}{\rho}\right) \right| \\[0.2cm]
    &\; = \; -\int \hat \pi \log\left( \frac{\hat \rho}{\rho} \right) = -\int \hat \pi \log\left(1 + \frac{\hat \rho - \rho}{\rho}\right) \\[0.2cm]
    &\; \leq\;  \int \hat \pi \frac{\hat \rho - \rho}{\rho} = \int \frac{\hat \pi}{\pi} \frac{\pi}{\rho} (\hat \rho - \rho) \\[0.2cm]
    &\; \leq\;  \max\left( \frac{\pi}{\rho} \right) \int |\hat \rho - \rho| 
    \\[0.2cm]
    &\; =\; 2M \, d_{\text{\normalfont TV}}(\hat \rho, \rho).
\end{aligned}
\end{equation}
Using~\eqref{eq: perturb 1},~\eqref{eq: perturb 2} we have
\begin{equation}\label{eq: perturb both}
\begin{aligned}
    \left| D_{\text{\normalfont KL}}(\hat \pi \,\Vert\, \hat \rho) - D_{\text{\normalfont KL}}(\pi \,\Vert\, \rho) \right| & \; \leq \; \left| D_{\text{\normalfont KL}}(\hat \pi \,\Vert\, \hat \rho) - D_{\text{\normalfont KL}}(\hat \pi \,\Vert\, \rho) \right| + \left| D_{\text{\normalfont KL}}(\hat \pi \,\Vert\, \rho) - D_{\text{\normalfont KL}}(\pi \,\Vert\, \rho) \right| \\[0.2cm]
    & \; \leq \; D_{\text{\normalfont KL}}(\hat \pi \,\Vert\, \pi) + 2M \, d_{\text{\normalfont TV}}(\hat \rho, \rho) +  2\log M \, d_{\text{\normalfont TV}}(\hat \pi, \pi)\\[0.2cm]
    & \; \leq \; D_{\text{\normalfont KL}}(\hat \pi \,\Vert\, \pi) + 2M \, \sqrt{\frac{1}{2} D_{\text{\normalfont KL}}(\hat \rho \,\Vert\, \rho)} + 2\log M \, \sqrt{\frac{1}{2} D_{\text{\normalfont KL}}(\hat \pi \,\Vert\, \pi)}\\[0.2cm]
\end{aligned}
\end{equation}

where in the last line we utilized Pinsker's inequality to bound the TV distance with the square root of the KL divergence. Now we apply~\eqref{eq: perturb both} to diffusion models as follows
\begin{equation}\label{eq: perturb diff}
    \begin{aligned}
        \left|D_{\text{\normalfont KL}} (\mathfrak{p}_0(\cdot; s_p) \,\Vert\, \mathfrak{q}_0(\cdot; s_q)) - D_{\text{\normalfont KL}} ({p}_0(\cdot; s_p) \,\Vert\, {q}_0(\cdot; s_q))\right| \; \leq\; & D_{\text{\normalfont KL}}({p}_0(\cdot; s_p) \,\Vert\, \mathfrak{q}_0(\cdot; s_p))\\[0.2cm]
        &+ 2M \, \sqrt{\frac{1}{2} D_{\text{\normalfont KL}}({p}_0(\cdot; s_q) \,\Vert\, \mathfrak{q}_0(\cdot; s_q))}\\[0.2cm]
        &+ 2\log M \, \sqrt{\frac{1}{2} D_{\text{\normalfont KL}}({p}_0(\cdot; s_p) \,\Vert\, \mathfrak{q}_0(\cdot; s_p))}. 
    \end{aligned}
\end{equation}

Furthermore, from Lemma~\ref{lem: discrete continuos mou}, we have
\begin{equation}\label{eq: mou apply}
    D_{\text{\normalfont KL}}({p}_0(\cdot; s_p) \,\Vert\, \mathfrak{q}_0(\cdot; s_p)) \;\leq\; c/T^2, \quad D_{\text{\normalfont KL}}({p}_0(\cdot; s_q) \,\Vert\, \mathfrak{q}_0(\cdot; s_q)) \;\leq\; c/T^2
\end{equation}
Putting together~\eqref{eq: perturb diff} and~\eqref{eq: mou apply}, we obtain
\begin{equation}
    \left|D_{\text{\normalfont KL}} (\mathfrak{p}_0(\cdot; s_p) \,\Vert\, \mathfrak{q}_0(\cdot; s_q)) - D_{\text{\normalfont KL}} ({p}_0(\cdot; s_p) \,\Vert\, {q}_0(\cdot; s_q))\right| \; \leq \;\epsilon_1(T)
\end{equation}
where $\epsilon_1(T) \DefinedAs c/T^2 + (2M + 2\log M)\sqrt{c/T^2}$. The second term dominates therefore $\epsilon_1(T) = O(1/T)$ which concludes the proof.

\end{proof}

\subsection{Proof of Lemma~\ref{lem: cont to discrete}}

\begin{proof}
    There are two sources of error we need to consider. First, we bound the error in approximating an integral with a sum:
    \[
    \begin{array}{rcl}
         && \displaystyle 
         \!\!\!\!  \!\!\!\!
         \!\!
         \left| \int_{\tau \,=\, 0}^1 \tilde \omega_\tau \, \mathbb{E}_{x\,\sim\, \mathfrak{p}_\tau(\cdot\,; s_p)} \left[ \,
        \norm{s_p(x, \tau) - s_q(x, \tau)}_2^2
        \,\right]  - \sum_{t \,=\, 0}^T \frac{1}{T} \tilde \omega_{t/T} \, \mathbb{E}_{x\,\sim\,{\mathfrak{p}}_{t/T}(\cdot\,; s_p)} \left[ \,
        \norm{s_p(x, t) - s_q(x, t)}_2^2
        \,\right]  \right|
        \\[0.4cm]
         & = & \displaystyle \left| \int_{\tau \,=\, 0}^1 f(\tau) d\tau  - \sum_{t \,=\, 0}^T f(t/T) \cdot \frac{1}{T} \right|
         \\[0.4cm]
         & = & \displaystyle  \frac{1}{T} \sup_{\tau \,\in\, [0, 1]} \left| \frac{df}{d\tau}\right|
    \end{array}
    \]
    where we have defined $f(\tau) \DefinedAs \tilde \omega_\tau \mathbb{E}_{x\,\sim\, \mathfrak{p}_\tau(\cdot\,; s_p)} \left[ \,
        \norm{s_p(x, \tau) - s_q(x, \tau)}_2^2
        \,\right]$. We now upper bound the supremum to show that it is finite:
        \begin{equation}\label{eq: integral expansion}
            \begin{array}{rcl}
                 \displaystyle
                 \frac{df}{d\tau} & = &
                 \displaystyle
                 \frac{d}{d \tau} \left( \int \mathfrak{p}_\tau(x\,; s_p)  \norm{s_p(x, \tau) - s_q(x, \tau)}_2^2 dx \right)
                 \\[0.4cm]
                 & = & 
                 \displaystyle
                 \int \frac{d}{d \tau} (\mathfrak{p}_\tau(x\,; s_p))  \norm{s_p(x, \tau) - s_q(x, \tau)}_2^2 dx 
            \\[0.4cm]
                 &  & 
                 \displaystyle+ \int \mathfrak{p}_\tau(x\,; s_p) \frac{d}{d \tau}  (\norm{s_p(x, \tau) - s_q(x, \tau)}_2^2) dx.
            \end{array}
        \end{equation}

        We bound each term in~\eqref{eq: integral expansion} separately.  Then the first term in~\eqref{eq: integral expansion} is bounded because $ \frac{d}{d \tau} (\mathfrak{p}_\tau(x\,; s_p)) $ is finite as characterized in Lemma~\ref{lem: heat equation}. The second term in~\eqref{eq: integral expansion} we expand further:
        \[
        \begin{array}{rcl}
            && \!\!\!\! \!\!\!\!
            \!\! \displaystyle
            \int \mathfrak{p}_\tau(x\,; s_p) \frac{d}{d \tau}  (\norm{s_p(x, \tau) - s_q(x, \tau)}_2^2) dx  
            \\[0.4cm]
            & = & \displaystyle
            \int 2\mathfrak{p}_\tau(x\,; s_p) \left\langle s_p(x, \tau) - s_q(x, \tau), \, \frac{d s_p(x, \tau)}{d\tau} - \frac{d s_q(x, \tau)}{d\tau}\right\rangle dx
            \\[0.4cm]
            &\leq& \displaystyle 
            2 \sup_{x, \tau} \norm{s_p(x, \tau) - s_q(x, \tau)}_2 \norm{\frac{d}{d \tau}(s_p(x, \tau) - s_q(x, \tau))}_2 
            \\[0.4cm]
            &\leq& \displaystyle 2 B_1 B_2.
        \end{array}
        \]
        
    The second source of error is replacing expectation over the continuous time marginal $\mathfrak{p}_{t/T}(\cdot\,; s_p)$ with expectation over the discrete time marginal $p_t(\cdot\,; s_p)$ which we can bound by using the fact that the two aforementioned marginals are close to each other.
    \[
    \begin{array}{rcl}
         && \displaystyle
         \!\!\!\!  \!\!\!\!
         \!\!
         \left|\sum_{t \,=\, 0}^T \frac{1}{T} \tilde \omega_{t/T} \, \mathbb{E}_{x\,\sim\,{\mathfrak{p}}_{t/T}(\cdot\,; s_p)} \left[ \,
        \norm{s_p(x, t) - s_q(x, t)}_2^2
        \,\right] -  \sum_{t \,=\, 0}^T \frac{1}{T} \tilde \omega_{t/T} \, \mathbb{E}_{x\,\sim\, p_t(\cdot\,; s_p)} \left[ \,
        \norm{s_p(x, t) - s_q(x, t)}_2^2
        \,\right]     \right|
         \\[0.4cm]
         & \leq & \displaystyle
         \sum_{t \,=\, 0}^T \frac{1}{T} \tilde \omega_{t/T} d_{TV} (p_t(\cdot\,; s_p), \mathfrak{p}_{t/T}(\cdot\,; s_p)) \cdot \sup_{x} \norm{s_p(x, \tau) - s_q(x, \tau)}^2_2
         \\[0.4cm]
         & \leq & \displaystyle
         \sum_{t \,=\, 0}^T \frac{1}{T} \tilde \omega_{t/\tau} \sqrt{\frac{c}{T^2} }\cdot B_1^2 
         \\[0.4cm]
         & \leq & \displaystyle
         T \cdot \frac{1}{T} \cdot \sqrt{\frac{c}{T^2} }\cdot B_1^2 
         \\[0.4cm]
         & = & \displaystyle O\left(\frac{1}{T}\right)
    \end{array}
    \]
    where we used Lemma~\ref{lem: discrete continuos mou} to get the last line which concludes the proof.
\end{proof}

%% file: app4_mixture_algorithm.tex
\newpage
\section{Composition with Forward KL Divergences}\label{app: mixture comp}


We start with the constrained problem formulation using forward KL divergence~\eqref{eq: constrained composition forward KL} which we rewrite here:
\begin{equation}\label{eq: mixture comp}
    \begin{array}{rl}
        \displaystyle\minimize_{u,~p} & 
       u
        \\[0.2cm]
        \subject &  D_{\text{KL}}(q^i\,\Vert\, p)\; \leq \; u \;\; \text{ for } i = 1, \ldots, m.
    \end{array}
\end{equation}
In the case of diffusion models, the KL divergence in~\eqref{eq: mixture comp} becomes the forward path-wise KL between the processes:
\begin{equation}\label{eq: mixture comp diff}
    \begin{array}{rl}
        \displaystyle\minimize_{u,~p} & 
       u
        \\[0.2cm]
        \subject &  D_{\text{KL}}(q^i_{0:T}(\cdot)\,\Vert\, p_{0:T}(\cdot; s))\; \leq \; u \;\; \text{ for } i = 1, \ldots, m.
    \end{array}
\end{equation}

It is important to note here that using the forward KL as a constraint makes sense when $q^i$ represent forward diffusion processes obtained by adding noise to samples from some dataset. We can also solve this forward KL constrained problem to compose multiple models; In that case we treat samples generated by each model as a separate dataset with underlying distribution $q^i_0(x_0)$.

In summary, the two key differences of Problem~\eqref{eq: mixture comp diff} to Problem~\eqref{eq: constrained alignment reverse KL} are: (i) The closeness of a model $p$ to a pretrained model $q^i$ is measured by the forward KL divergence $D_{\text{KL}}(q^i\,\Vert\,p)$, instead of the reverse KL divergence $D_{\text{KL}}(p\,\Vert\,q^i)$; (ii) The distributions $\{q^i\}_{i\,=\,1}^m$ can be the distributions underlying $m$ datasets, not necessarily $m$ pretrained models. 

Regardless of whether the $q^i$ represent pretrained models or datasets, evaluating $D_{\text{KL}}(q^i_{0:T}(\cdot)\,\Vert\, p_{0:T}(\cdot; s))$ is intractable since it requires knowing $q^i_{0:T}(\cdot)$ which in turn requires knowing $q^i_0(\cdot)$ exactly. To get around this issue we formulate a closely related problem to~\eqref{eq: mixture comp diff} by replacing the KL with the Evidence Lower Bound (Elbo):
\begin{equation}\label{eq: constrained alignment elbo}
    \begin{array}{rl}
        \displaystyle\minimize_{u,~p} & 
       u
        \\[0.2cm]
        \subject &  \text{Elbo}(q^i_{0:T} ; p_{0:T})\; \leq \; u \;\; \text{ for } i = 1, \ldots, m

    \end{array}
\end{equation}
where the Elbo is defined as
\begin{equation}\label{eq: elbo definition}
    \text{Elbo}(q_{0:T} ; p_{0:T}) \;\DefinedAs\; \mathbb{E}_{x_0 \sim q_0} \mathbb{E}_{q(x_{1:T}|x_0)} \log \frac{p_{0:T}(x_{0:T})}{q(x_{1:T}|x_0)}.
\end{equation}
We note that the typical approach to train a diffusion model is minimizing the Elbo. Furthermore, minimizing $\text{Elbo}(q_{0:T} ; p_{0:T})$ over $p$ is equivalent to minimizing the KL divergence $D_{\text{KL}}(q^i_{0:T}(\cdot)\,\Vert\, p_{0:T}(\cdot; s))$  since they only differ by a constant that does not depend on $p$. (see~\cite{khalafi2024constrained} for more details on this)

For a given $\lambda$, we define a weighted mixture of distributions as
\begin{equation}\label{eq: mixture}
    q_{\text{mix}}^{(\lambda)} (\cdot)
    \; = \;
    \sum_{i = 1}^m \frac{\lambda_i}{\lambda^\top 1} q^i(\cdot)
\end{equation}
and we denote by $H(q)$ the differential entropy of a given distribution $q$,
\begin{equation}
     H(q) \;\DefinedAs\; - \mathbb{E}_{x \sim q} [\log q(x)].
\end{equation}

\begin{theorem}\label{thm: forward KL general}
    Problem~\eqref{eq: constrained alignment elbo} is equivalent to the following unconstrained problem:
    \begin{subequations}
    \begin{equation}\label{eq: unconstrained alignment forward KL}
    \minimize_{p}
    \; 
    D_{\text{\normalfont KL}}(q_{\text{\normalfont mix}}^{(\lambda^\star)}\,\Vert\, p)
    \end{equation}
    where $\lambda^\star$ is the optimal dual variable given by $\lambda^\star = \argmax_{\lambda\,\geq\,0} D(\lambda)$. The dual function has the explicit form,
    $D(\lambda) = H (q^{(\lambda)}_{\text{\normalfont mix}})$. Furthermore, the optimal solution of~\eqref{eq: unconstrained alignment reverse KL} is given by
    \begin{equation}\label{eq: unconstrained alignment reverse KL solution}
        p^\star = q_{\text{\normalfont mix}}^{(\lambda^\star)}.
    \end{equation}
    \end{subequations}
\end{theorem}
Unlike the reverse KL case, here we can characterize the optimal dual multipliers, and the optimal solution further; Note that the optimal dual multiplier $\lambda^\star = \argmax_{\lambda \geq\,0} D(\lambda) = \argmax_{\lambda \geq\,0} H(q_{\text{\normalfont mix}}(\cdot;\lambda^\star))$ is one that maximizes the differential entropy $H(\cdot)$ of the distribution of the corresponding mixture. This implies that the optimal solution is the most diverse mixture of the individual distributions.

There are many potential use cases where we may want to compose distributions that don't overlap in their supports; For example when combining distributions of multiple dissimilar classes of a dataset. The following characterizes the optimal solution in such settings.

\begin{corollary}
    For the special case where the distributions $q^i$ all have disjoint supports, the optimal dual multiplier $\lambda^\star$ of Problem~\eqref{eq: constrained alignment elbo} can be characterized explicitly as
    \[
        \lambda^\star_i 
        \; = \;
        \frac{e^{H(q^i)}}{\sum_{j = 1}^m e^{H(q^j)}}.
    \]
\end{corollary}

%% file: app8_algorithms.tex
\section{Algorithm Details}\label{app: algorithms}

\subsection{Alignment}

Recall from Section~\ref{subsec: Reward alignment for diffusion models} that the algorithm consists of two alternating steps:

\textbf{Primal minimization: } At iteration $n$, we obtain a new model $s^{(n+1)}$ via a Lagrangian maximization:
\[
    s^{(n + 1)} 
    \; \in  \;
    \argmin_{s \,\in\, \mathcal{S}}\; \bar{L}_{\text{ALI}}(s_{p}, \lambda^{(n)}).
\]
\textbf{Dual maximization: } Then, we use the model $s^{(n+1)}$ to estimate the constraint violation $\mathbb{E}_{x_0}[ r (x_0)] - b$, denoted as $r(s^{(n+1)}) - b$, and perform a dual sub-gradient ascent step:
\[
    \lambda^{(n + 1)} 
    \; = \;
    \left[
    \, \lambda^{(n)} 
    \, +\,
    \eta\, \left(r(s^{(n + 1)}) - b\right) \,\right]_+.
\]
In practice we replace minimization over $\mathcal{S}$ with minimization over a parametrized family of functions $\mathcal{S}_\theta$. The full algorithm is detailed in Algorithm~\ref{alg: alignment}.

\begin{algorithm}[H]
	\caption{Primal-Dual Algorithm for Reward Alignment of Diffusion Models}\label{alg: alignment}
	\begin{algorithmic}[1]
		\State \textbf{Input}: total diffusion steps $T$, diffusion parameter $\alpha_t$, total dual iterations $H$, number of primal steps per dual update $N$, dual step size $\eta_d$, primal step size $\eta_p$, initial model parameters $\theta(0)$.
		\State \textbf{Initialize}: $\lambda(1) = 1/m$.
		\For{$h = 1, \cdots, H$}
        \State Initialize $\theta_1  =  \theta{(h - 1)}$ 
		\For{$n = 1, \cdots, N$}

		\State Take a primal gradient descent step:
        \begin{equation}\label{eq: alignment grad descent}
            \theta_{n + 1} \;=\;   \displaystyle \theta_n \,-\, \eta_p \, \nabla_\theta  \bar{L}_{\text{ALI}}(\theta, \lambda^{(n)}).
        \end{equation}
		\EndFor
        \State Set the value of the parameters to be used for the next dual update: $\theta(h) \; = \; \theta_{N + 1}$.
		\State Update dual multipliers for $i = 1, \cdots, m$:
		
		\begin{equation}
		 \lambda_i (h + 1) 
		\; = \;
		\left[ \lambda_i(h) 
		\, + \,
		\eta_d \,
        (\mathbb{E}_{x_0\,\sim\,p_0(\cdot; s_\theta)} \big[ \,r_i(x_0)\, \big] - b_i)\right]_+.
		\end{equation}
        
		\EndFor
	\end{algorithmic}
\end{algorithm}

We now discuss the practicality of the primal gradient descent step~\eqref{eq: alignment grad descent} regarding the Lagrangian function,
\begin{equation}\label{eq: lagrangian alignment param}
    \bar{L}_{\text{ALI}}(\theta, \lambda) 
    \; = \;
    D_{\text{KL}} \big(\, p_{0:T}(\cdot; s_\theta)\,\Vert\, q_{0:T}(\cdot; s_q) \,\big) - \sum_i \lambda_i (\mathbb{E}_{x_0\,\sim\,p_0(\cdot; s_\theta)} \big[ \,r_i(x_0)\, \big] - b_i).
\end{equation}
To derive the gradient of $\bar{L}_{\text{ALI}}(\theta, \lambda)$, we first take the derivative of the expected reward terms by noting that the expectation is taken over a distribution that depends on the optimization variable $\theta$.   We can use the following result (Lemma 4.1 from~\cite{fan2023dpokreinforcementlearningfinetuning}) to take the gradient inside the expectation.

\begin{lemma}\label{lem: log trick rewards}
  If \( p_\theta(x_{0:T}) r(x_0) \) and \( \nabla_\theta p_\theta(x_{0:T}) r(x_0) \) are continuous functions of \( \theta \), then we can write the gradient of the reward function as
  \[
      \nabla_\theta \mathbb{E}_{x_0\,\sim\,p_0(\cdot; s_\theta)} \big[ \,r(x_0)\, \big] 
      \; = \; 
      \mathbb{E}_{x_{0:T}\sim p_{0:T}(\cdot; s_\theta)} \left[ r(x_0) \sum_{t=1}^{T} \nabla_\theta \log p(x_{t-1} \mid x_t;\, s_\theta) \right].
  \]
\end{lemma}

For the gradient of the KL divergence, we have
\[
    \begin{array}{rcl}
         &&
         \!\!\!\! \!\!\!\!
         \!\!
         \nabla_\theta D_{\text{KL}} \big(\, p_{0:T}(\cdot; s_\theta)\,\Vert\, q_{0:T}(\cdot; s_q) \,\big) 
    \\[0.2cm]
    & = &
    \displaystyle
    \nabla_\theta \left(\sum_{t\,=\,1}^{T} \mathbb{E}_{x_t \,\sim\, p_t(\cdot; s_\theta)}\left[ 
    \,\frac{1}{2 \sigma_t^2} \Vert s_\theta (x_t, t) - s_q (x_t, t) \Vert^2
    \,\right]\right)
         \\[0.2cm]
         & = & \displaystyle
          \nabla_\theta \left(\sum_{t\,=\,1}^T 
     \mathbb{E}_{x_t\,\sim\,p_t(\cdot; s_\theta)} 
     \left[ D_{\text{KL}}(p(x_{t - 1}\,\vert\,x_t; s_\theta)\,\Vert\, q(x_{t - 1}\,\vert\,x_t; s_q)) \right]\right) 
     \\[0.2cm]
     & = & \displaystyle 
     \sum_{t\,=\,1}^T 
     \mathbb{E}_{x_t\,\sim\,p_t(\cdot; s_\theta)} 
     \left[ \nabla_\theta D_{\text{KL}}(p(x_{t - 1}\,\vert\,x_t; s_\theta)\,\Vert\, q(x_{t - 1}\,\vert\,x_t; s_q)) \right]
    \end{array}
\]
\[
    + \sum_{t\,=\,1}^T  \mathbb{E}_{x_t\,\sim\,p_t(\cdot; s_\theta)} 
     \left[ \sum_{t' > t}^T \nabla_\theta \log p(x_{t'-1} \mid x_{t'};\, s_\theta) D_{\text{KL}}(p(x_{t - 1}\,\vert\,x_t; s_\theta)\,\Vert\, q(x_{t - 1}\,\vert\,x_t; s_q)) \right].
\]
For simplicity, we omit the second term in practice, as it has negligible effect on performance. See~\cite[Appendix A.3]{fan2023dpokreinforcementlearningfinetuning} for the derivation.

\subsection{Composition}\label{app: composition surrogate}
For composition, we take a similar approach to Algorithm~\ref{alg: alignment}. Recall from Lemma~\ref{lem: primal true score} that the Lagrangian minimizer for the constrained composition problem can be found by minimizing
\[
    \hat{L}_{\text{AND}}(\theta, \lambda) \;\DefinedAs\; \sum_{t \,=\, 0}^T \omega_t \, \mathbb{E}_{x_0 \, \sim \, q_{\text{\normalfont AND}}^{(\lambda)}(\cdot)} \mathbb{E}_{x_t \, \sim \, q(x_t\,|\,x_0)}\left[ \norm{s_\theta(x_t, t) - \nabla \log q(x_t | x_0)}^2 \right].
\]
Thus, we detail the algorithm for composition in Algorithm~\ref{alg: composition}.

\begin{algorithm}
	\caption{Primal-Dual Algorithm for Product Composition (AND) of Diffusion Models}\label{alg: composition}
	\begin{algorithmic}[1]
		\State \textbf{Input}: total diffusion steps $T$, diffusion parameter $\alpha_t$, total dual iterations $H$, number of primal steps per dual update $N$, dual step size $\eta_d$, primal step size $\eta_p$, initial model parameters $\theta(0)$.
		\State \textbf{Initialize}: $\lambda(1) = 1/m$.
		\For{$h = 1, \cdots, H$}
        \State Initialize $\theta_1  =  \theta{(h - 1)}$ 
		\For{$n = 1, \cdots, N$}

		\State Take a primal gradient descent step:
        \begin{equation}
            \theta_{n + 1}  \;= \;   \displaystyle \theta_n \,-\, \eta_p \, \nabla_\theta  \hat{L}_{\text{AND}}(\theta, \lambda^{(n)}).
        \end{equation}
		\EndFor
        \State Set the value of the parameters to be used for the next dual update: $\theta(h) \; = \; \theta_{N + 1}$.
		\State Update dual multipliers for $i = 1, \cdots, m$:
		
		\begin{equation}
		    \tilde \lambda_i (h + 1) 
		\; = \;
		\lambda_i(h) 
		\, + \,
		\eta_d \,
        D_{\text{\normalfont KL}} (p_0(\cdot\,; s_{\theta(h)}) \,\Vert\, q_0^i(\cdot\,; s^i)).
		\end{equation}
		
        \State $\lambda (h + 1) = \text{proj}\left(\tilde \lambda (h + 1)\right), \,$ where $\text{proj}(y)$ projects its input onto the simplex $\lambda^T 1 = 1$.
        
		\EndFor
	\end{algorithmic}
\end{algorithm}
The projection of the dual multiplier vector (line 10) ensures that $\lambda^\top \mathbf{1} = 1$, as required when maximizing the dual function (see the proof of Theorem~\ref{thm: reverse KL general}).

Note that Algorithm~\ref{alg: composition} implicitly requires samples from the weighted product distribution $q_{\text{\normalfont AND}}^{(\lambda)}(\cdot)$ in order to minimize the Lagrangian $\hat{L}_{\text{AND}}(\theta, \lambda)$. We obtain these samples using the Annealed MCMC sampling algorithm proposed in~\cite{du2024reducereuserecyclecompositional}.

\textbf{Skipping the primal.} As discussed in Section~\ref{sec: experiments}, both Annealed MCMC sampling and the minimization of the Lagrangian $\hat{L}_{\text{AND}}(\theta, \lambda)$ at each primal step—to match the true score $\nabla \log q_{\text{\normalfont AND}}^{(\lambda)}$—are challenging and computationally expensive. Therefore, for all settings except the low-dimensional case described in Appendix~\ref{app: gaussian experiments}, we employ Algorithm~\ref{alg: composition dual only}, which skips the primal step entirely.

In Algorithm~\ref{alg: composition dual only} we bypass the primal steps by using the surrogate product score, rather than the true score, to compute the point-wise KL used in the dual updates. The distinction between the true and surrogate scores is discussed in detail in~\cite{du2024reducereuserecyclecompositional}.

\begin{equation}\label{eq: true product score}
   \text{true score: \quad} \nabla \log q^{(\lambda)}_{\text{AND},\, t} (x_t) = \nabla \log \left( \int \sum_i (q_0(x_0))^{\lambda_i} q(x_t| x_0) dx_0\right) 
\end{equation}

\begin{equation}\label{eq: surrogate product score}
   \text{surrogate score: \quad} \nabla \log \hat q^{(\lambda)}_{\text{AND},\, t} (x_t) = \sum_i \lambda_i \nabla \log \left( \int q_0(x_0)q(x_t| x_0) dx_0\right)
\end{equation}

\begin{algorithm}
	\caption{Dual-Only Algorithm for Product Composition (AND) of Diffusion Models}\label{alg: composition dual only}
	\begin{algorithmic}[1]
		\State \textbf{Input}: total diffusion steps $T$, diffusion parameter $\alpha_t$, total dual iterations $H$, dual step size $\eta_d$.
		\State \textbf{Initialize}: $\lambda(1) = 1/m$.
		\For{$h = 1, \cdots, H$}
		\State Update dual multipliers for $i = 1, \cdots, m$:
		
		\begin{equation}
		    \tilde \lambda_i (h + 1) 
		\; = \;
		\lambda_i(h) 
		\, + \,
		\eta_d \, D_{\text{\normalfont KL}} \left(\hat q^{(\lambda(h))}_{\text{AND}, 0}(\cdot) \,\Vert\, p_0(\cdot\,; s^i) \right).
		\end{equation}
		
        \State $\lambda (h + 1)  \;= \; \text{proj}\left(\tilde \lambda (h + 1)\right), \,$ where $\text{proj}(y)$ projects its input onto the simplex $\lambda^T 1 = 1$.
        
		\EndFor
	\end{algorithmic}
\end{algorithm}
For a given $\lambda$, the surrogate score can be easily computed:
\[
    \begin{array}{rcl}
        \nabla \log \hat q^{(\lambda)}_{\text{AND},\, t} (x_t) 
        & = &  \displaystyle
        \sum_i \lambda_i \nabla \log \left( \int q_0(x_0)q(x_t| x_0) dx_0\right)
        \\[0.2cm]
        & = & \displaystyle 
        \sum_i \lambda_i \nabla \log p_t(x_t; s^i)
    \end{array}
\]
and thus we can use Lemma~\ref{lem: marginal kl computation} to compute the point-wise KLs needed for the dual update. As for the samples needed from the true product distribution, we also replace them with samples obtained by running DDIM using the surrogate score.

\newpage
\section{Additional Experiments and Experimental Details}\label{app: experiments}
\subsection{Related work}

Here we review related work and explain why these approaches are not directly applicable as baselines for our experiments.

In~\cite{skreta2025the} they propose a superposition method to sample from the mixture of diffusion models with arbitrary weights. However, they only use equal weight mixtures and don't discuss different weights. They also devise a method to sample points that have equal likelihood under different models which is fundamentally different to the product composition that we discuss in this work.

Existing works including~\cite{christopher2024constrained, liang2024multi, narasimhan2024constrained, zampini2025training} discuss constrained sampling from diffusion models, but the nature of their constraints is completely different from our work as it mainly involves sampling from a constrained set and they propose to do this through projection onto a feasible set at each diffusion time step. It is not clear how to apply these methods to reward constraints or how to use them to preserve distance to a model.

Other works like~\cite{giannone2023aligning, chen2024towards} enforce very specific constraints by adding additional losses with fixed weights to the objective which implicitly enforces the constraint. These methods are very specific to the constraints they are designed for and do not generalize to arbitrary reward functions and don't give us a way to constrain closeness to a model.

\subsection{Low-dimensional synthetic experiments}\label{app: gaussian experiments}
To visually illustrate the difference between the constrained and unconstrained approaches, we conduct experiments where the generated samples lie in $\mathbb{R}^2$. For the score predictor we used the same ResNet architecture as used in~\cite{du2024reducereuserecyclecompositional}.

\textbf{Product composition (AND).}
Unlike the image experiments, in this low-dimensional setting we use Algorithm~\ref{alg: composition} for product composition. See Figure~\ref{fig: gaussians AND} for visualization of the resulting distributions.

\textbf{Mixture composition (OR).}
For this experiment we used the same Algorithm as the one used in~\cite{khalafi2024constrained} for mixture of distributions. The only modification is doing an additional dual multiplier projection step similar to the last step of the product composition Algorithm~\ref{alg: composition}. See Figure~\ref{fig:gaussians OR} for visualization of the resulting distributions.

\subsection{Reward product composition (Section~\ref{subsec: composition experiments} (I))}
\label{app:exp reward prod comp}

\begin{figure}[h]
    \centering
    \includegraphics[width=0.5\linewidth]{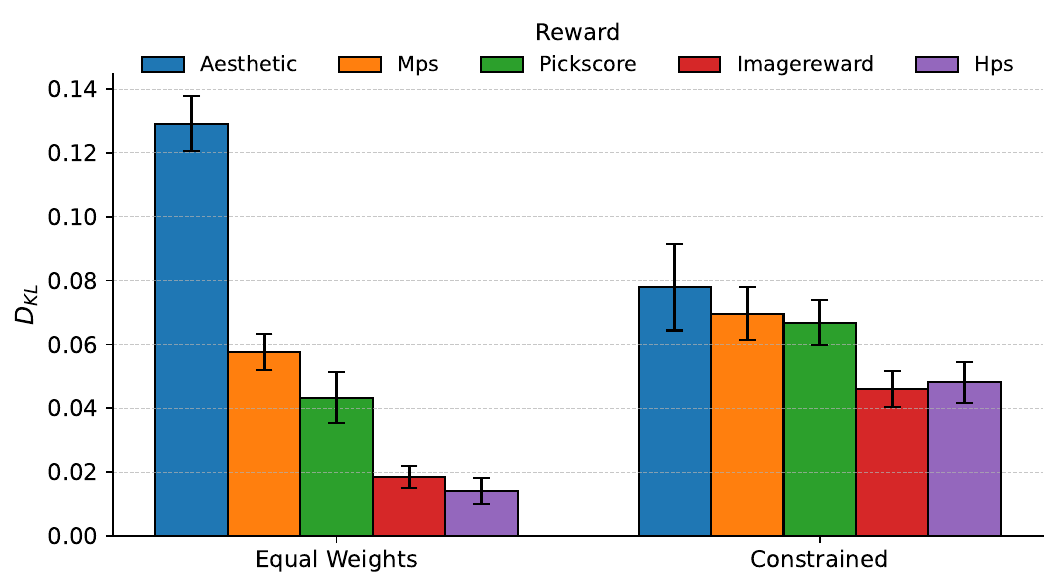}
    \caption{KL divergence for the product composition of 5 adapters pretrained with different rewards. Error bars  denote the standard deviation computed across 8 text prompts each with four samples.}
    \label{fig:kl-reward-prod}
\end{figure}

\textbf{Implementation details and hyperparameters}. We finetuned the model using the Alignprop~\cite{prabhudesai2024aligningtexttoimagediffusionmodels} official implementation~\footnote{https://github.com/mihirp1998/AlignProp} for each individual reward using the hyperparameters reported in Table~\ref{tab:single reward hyperparams}. We then composed the trained adapters running dual ascent using the surrogate score as described in section~\ref{app: composition surrogate}. We use the average of scores (denoted as ``Equal weights'') as a baseline. Hyperparameters are described in Table~\ref{tab:hyperparams reward prod}. The reward values reported in Figure~\ref{fig: all rewards alignment} were normalised so that 0\% corresponds to the reward obtained by the pretrained model, and 100\% the reward obtained by the model finetuned solely on the corresponding reward.

\textbf{Additional results}. As shown in Figure~\ref{fig:kl-reward-prod}, equal weighting leads to disparate KL divergences across adapters -- in particular high KL with respect to the adapter trained with the ``aesthetic'' reward -- while our constrained approach effectively reduces the worst case KL, equalizing divergences across adapters. Figure~\ref{fig: reward prod images} shows images sampled from these two compositions exhibit different characteristics, with our constrained approach producing smoother backgrounds, shallower depth of field and more painting-like images.

\begin{table}[H]
\centering
\begin{tabular}{ll}
\toprule
\textbf{Hyperparameter}        & \textbf{Value} \\
\midrule
Batch size          & $64$ \\
Samples per epoch              & 128 \\
Epochs                         & 10 \\
Sampling steps                 & 50 \\
Backpropagation sampling       & Gaussian \\
KL penalty                     & 0.1 \\
Learning rate                  & $1 \times 10^{-3}$ \\
LoRA rank                      & 4 \\
\bottomrule
\end{tabular}
\vspace{0.1cm}
\caption{Hyperparameters used to finetune models using individual rewards.}
\label{tab:single reward hyperparams}
\end{table}

\begin{table}[h]
\centering
\begin{tabular}{ll}
\toprule
\textbf{Hyperparameter} & \textbf{Value} \\
\midrule
Base model             & \texttt{runwayml/stable-diffusion-v1-5} \\
\multirow{2}{*}{Prompts}    & \texttt{\{"cheetah", "snail", "hippopotamus",} \\
                            & \texttt{"crocodile", "lobster", "octopus"\}} \\
Resolution             & 512 \\
Batch size             & 4 \\
Dual steps             & 5 \\
Dual learning rate     & 1.0 \\
Sampling steps    & 25 \\
Guidance scale         & 5.0 \\
Rewards                & \texttt{aesthetic, hps, pickscore, imagereward, mps} \\
\bottomrule
\end{tabular}
\vspace{0.1cm}
\caption{Hyperparameters for product composition of models finetuned with different rewards.}\label{tab:hyperparams reward prod}
\end{table}
\subsection{Concept composition (Section~\ref{subsec: composition experiments} (II))}

We present additional results for concept composition using three different concepts (as opposed to just 2 in the main paper and in~\cite{skreta2025the}) As seen in table~\ref{tab:clip_blip2}, our approach retains a clear advantage in both CLIP and BLIP scores. See Table~\ref{fig:example_table} for examples of images generated using each method. Images with the constrained method typically do a better job of representing all concepts.

\begin{table}[h]
  \centering
  \begin{tabular}{@{}lcc@{}}
    \toprule
    & \textbf{Min. CLIP (↑)} & \textbf{Min. BLIP (↑)} \\
    \midrule
    Combined Prompting & 21.52 & 0.206  \\
    Equal Weights      & 22.18 & 0.203  \\
    Constrained (Ours) & \textbf{22.45} & \textbf{0.221}  \\
    \bottomrule
  \end{tabular}
  \vspace{0.1cm}
  \caption{Comparing constrained approach to baselines on minimum CLIP and BLIP scores. The scores are averaged over 50 different prompt triplets sampled from a list of simple prompts.}
  \label{tab:clip_blip2}
\end{table}

\begin{table}[H]
\centering
\begin{tabular}{m{3cm} m{3cm} m{3cm}}
\multicolumn{1}{c}{\textbf{Combined Prompting}} & \multicolumn{1}{c}{\textbf{Equal Weights}} & \multicolumn{1}{c}{\textbf{Constrained}} \\[1ex]

\includegraphics[width=2.8cm]{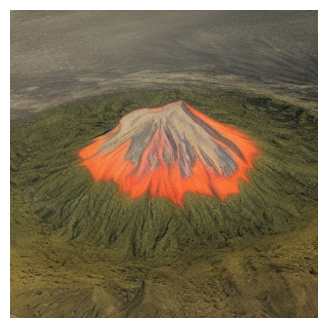} &
\includegraphics[width=2.8cm]{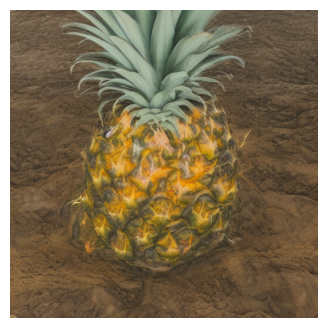} &
\includegraphics[width=2.8cm]{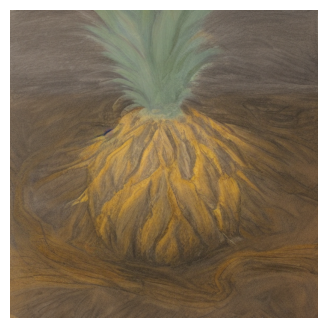} \\

\includegraphics[width=2.8cm]{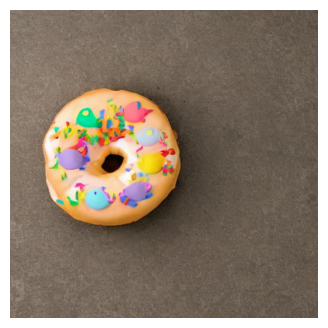} &
\includegraphics[width=2.8cm]{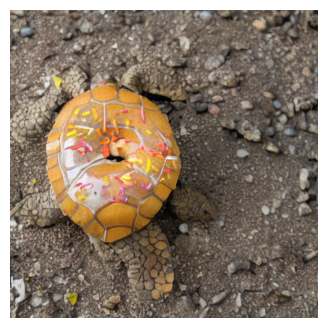} &
\includegraphics[width=2.8cm]{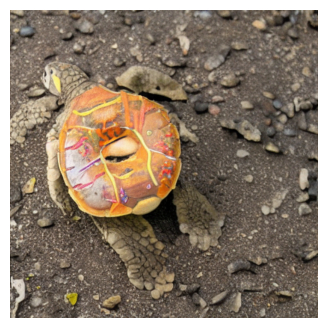} \\

\includegraphics[width=2.8cm]{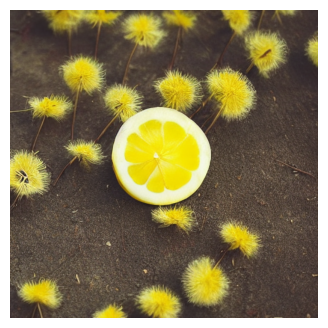} &
\includegraphics[width=2.8cm]{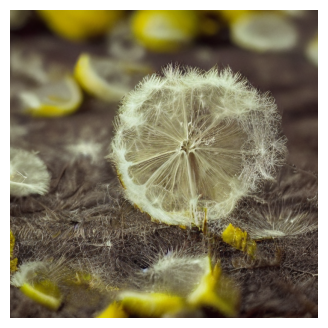} &
\includegraphics[width=2.8cm]{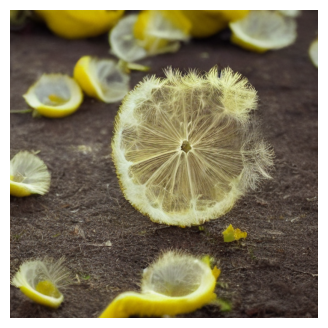} \\

\includegraphics[width=2.8cm]{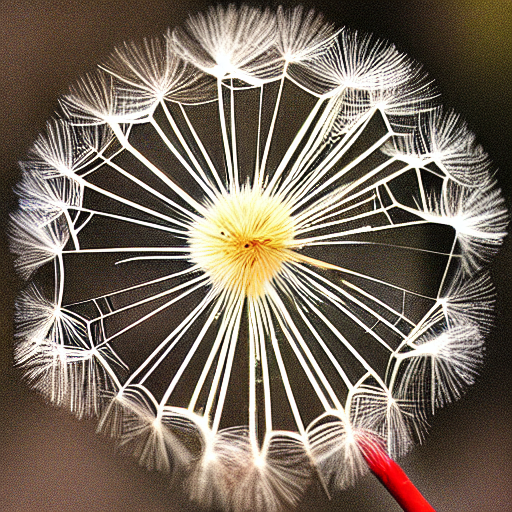} &
\includegraphics[width=2.8cm]{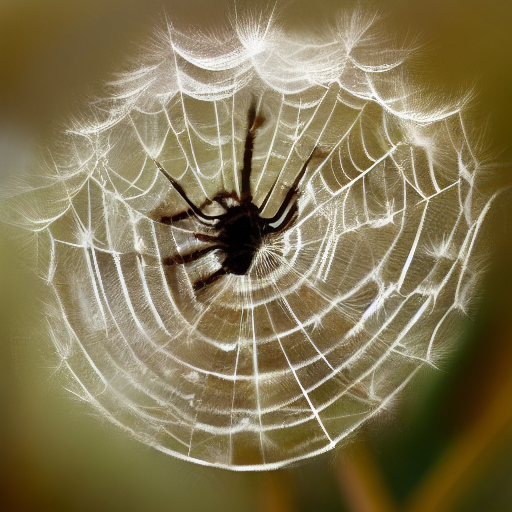} &
\includegraphics[width=2.8cm]{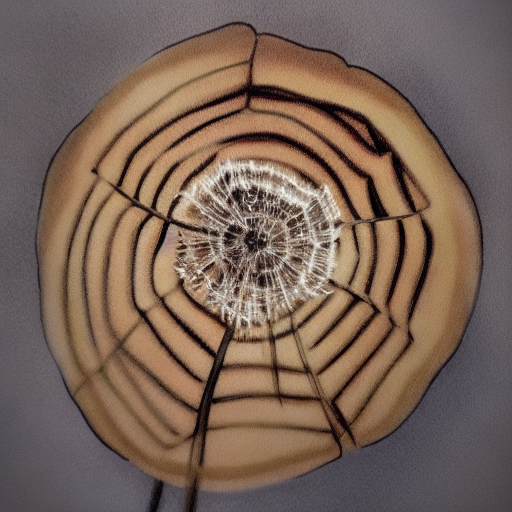} \\

\end{tabular}
\caption{Concept composition examples for each method. Prompts used for each row:\\ \textbf{Row 1: } "a pineapple", "a volcano". \textbf{Row 2: }  "a donut", "a turtle". \textbf{Row 3: }  "a lemon", "a dandelion". \textbf{Row 4: }  "a dandelion", "a spider web", "a cinammon roll". }
\label{fig:example_table}
\end{table}

\subsection{Concept composition for text-to-audio diffusion models}

We note that our proposed framework and theoretical analysis do not depend on any specific modality or task types. From our theoretical guarantees, we would expect experiments in other modalities to provide results similar to those presented for images. To validate this, we conduct concept composition experiments with a text-to-audio diffusion model as an example of another modality. We treat a text-to-audio diffusion model (in this case AudioLDM~\cite{liu2023audioldmtexttoaudiogenerationlatent}) conditioned on different inputs, each representing a concept, as the models to be composed. We apply our constrained learning to find the optimal weights to compose these two models, and use the CLAP score~\cite{elizalde2022claplearningaudioconcepts} to measure the similarity between the generated audio samples and the text prompts representing each model.

\begin{table}[h]
  \centering
  \begin{tabular}{@{}lc@{}}
    \toprule
    & \textbf{Min. CLAP Score(↑)}\\
    \midrule
    Combined Prompting & 0.816  \\
    Equal Weights      & 1.57 \\
    Constrained (Ours) & \textbf{1.92} \\
    \bottomrule
  \end{tabular}
  \vspace{0.1cm}
  \caption{Minimum CLAP scores across prompts for each method}
  \label{tab:clap}
\end{table}
Similar to concept composition for images, we observe in Table~\ref{tab:clap} that using our constrained approach, the minimum CLAP score across prompts increases compared to the two baselines. The constraints ensure closeness to each model, which in turn results in a more equal representation of the concepts.

\subsection{Alignment experiments}

\textbf{Reward normalization}. In practice, setting constraint levels for multiple rewards that are both feasible and sufficiently strict to enforce the desired behavior is challenging. Different rewards exhibit widely varying scales. This is illustrated in Table~\ref{tab:reward stats}, which shows the mean and standard deviation of reward values for the pretrained model. This issue can be exacerbated by the unknown interdependencies among constraints and the lack of prior knowledge about their relative difficulty or sensitivity. 

In order to tackle this, we propose normalizing rewards using the pretrained model statistics as a simple yet effective heuristic. This normalization facilitates the setting of constraint levels, enables direct comparisons across rewards and enhances interpretability. In all of our experiments, we apply this normalization before enforcing constraints. Explicitly, we set
\begin{align}
  \tilde{r}
  \; = \;
  \frac{r-\hat{\mu}_{\text{pre}}}{\hat{\sigma}_{\text{pre}}} 
\end{align}
where $r$ denotes the original reward and $\hat{\mu}_{pre}, \hat{\sigma}_{pre}$ the sample mean and standard deviation of the reward for the pretrained model. 
We find that, with this simple transformation, setting equal constraint levels can yield satisfactory results while forgoing extensive hyperparameter tuning.

\begin{table}[h]
\centering
\begin{tabular}{lcc}
\toprule
\textbf{Reward} & \textbf{Mean} & \textbf{Std} \\ 
\midrule
Aesthetic            & 5.1488  & 0.4390 \\
HPS                  & 0.2669  & 0.0057 \\
MPS                  & 5.2365  & 3.5449 \\
PickScore            & 21.1547 & 0.6551 \\
Local Contrast & 0.0086  & 0.0032 \\
Saturation     & 0.1060  & 0.0706 \\
\bottomrule
\end{tabular}
\vspace{0.1cm}
\caption{Mean and standard deviation of reward values for the pretrained model.}
\label{tab:reward stats}
\end{table}

\textbf{The effects of varying the constraint thresholds.} What we observed by varying the reward constraint thresholds in our experiments was that for thresholds up to 1.0 (i.e. $\hat \mu_{\text{pre}} + 1.0 \times \hat \sigma_{\text{pre}}$ for each reward) the model was typically able to satisfy the constraints with minimal violation. Another trend that we observed was that increasing thresholds usually leads to constraints that are harder to satisfy leading to higher Lagrange multipliers and resulting in higher KL to the pretrained model. See Tables~\ref{tab: mps threshold ablation} and~\ref{tab: pickscore threshold ablation}.

An advantage of our constrained approach is that Lagrange multipliers give information about the sensitivity of the objective with respect to relaxing the constraints i.e. if the multiplier for a certain reward ends up being much higher than the rest it means that constraint is particularly harder to satisfy. Consequently, even slightly relaxing the threshold for the corresponding reward can lead to much smaller KL objective.

\begin{table}[H]
\centering
\begin{tabular}{lcccc}
\hline
Constraint & Threshold & Slack & Dual Variable & $D_{KL}$ \\
\hline
contrast & 0.250 & -0.245 & 0.282 & 0.177 \\
contrast & 0.500 & -0.985 & 0.000 & 0.296 \\
contrast & 1.000 & -0.381 & 0.000 & 0.332 \\
saturation & 0.250 & -0.126 & 0.081 & 0.177 \\
saturation & 0.500 & 0.060 & 0.006 & 0.296 \\
saturation & 1.000 & 0.052 & 1.195 & 0.332 \\
\hline
\end{tabular}
\vspace{0.1cm}
\caption{MPS reward alignment with saturation and contrast constraints, for varying thresholds.}
\label{tab: mps threshold ablation}
\end{table}

\begin{table}[H]
\centering

\begin{tabular}{lcccc}
\hline
Constraint & Threshold & Slack & Dual Variable & $D_{KL}$ \\
\hline
contrast & 0.250 & -0.684 & 0.000 & 0.136 \\
contrast & 0.500 & -1.011 & 0.000 & 0.109 \\
contrast & 1.000 & 0.661 & 0.192 & 0.293 \\
saturation & 0.250 & -0.025 & 0.014 & 0.136 \\
saturation & 0.500 & -0.060 & 0.000 & 0.109 \\
saturation & 1.000 & 0.062 & 1.020 & 0.293 \\
\hline
\end{tabular}
\vspace{0.1cm}
\caption{Pickscore reward alignment with saturation and contrast constraints, for varying thresholds.}
\label{tab: pickscore threshold ablation}
\end{table}

\textbf{I.  MPS + local contrast, saturation.}

In this experiment, we augment a standard alignment loss—trained on user preferences—with two differentiable rewards that control specific image characteristics: local contrast and saturation. These rewards are computationally inexpensive to evaluate and offer direct interpretability in terms of their visual effect on the generated images. In addition, the unconstrained maximization of these features would lead to undesirable generations.  other potentially useful rewards not explored in this work are brightness, chroma energy, edge strength, white balancing and histogram matching.

\textbf{Local contrast reward}. In order to prevent images with excessive sharpness, we minimize the ``local contrast'', which we define as the mean absolute difference between the luminance of the image and a low-pass filtered version. Explicitly, let $Y$ denote the luminance, computed as $Y=0.2126 R+0.7152 G+0.0722 B$, and $G_\sigma * Y$ the luminance blurred with a  Gaussian kernel of standard deviation $\sigma=1.0$. We minimize the average per pixel difference by maximizing the reward 
$$
r_C
\; = \;
- \frac{1}{H W} \sum_{i, j}\left|Y_{ij}-\left(G_\sigma * Y\right)_{ij}\right|
$$
where $H, W$ denote image dimensions.

\textbf{Saturation reward}. To discourage overly saturated images, we simply penalize saturation, which we compute from $R,G,B$ pixel values as
\begin{align*}
  r_S
  \; = \;
  - \frac{1}{H W} \sum_{i, j} \frac{\max _{c \in\{R, G, B\}} x_{i,j}^{(c)}-\min _{c \in\{R, G, B\}} x_{i,j}^{(c)}}{\max _{c \in\{R, G, B\}} x_{i,j}^{(c)}+\varepsilon}
\end{align*}
where $\varepsilon=1\times10^{-8}$ is a small constant added for numerical stability.

\textbf{Implementation details and hyperparameters}. We implemented our primal-dual alignment approach (Algorithm~\ref{alg: alignment}) in the Alignprop framework. Following their experimental setting, we use different animal prompts for training and evaluation.
Hyperparameters are detailed in Table~\ref{tab:reward saturation hparams}.

\begin{table}[h]
\centering
\begin{tabular}{ll}
\toprule
\textbf{Hyperparameter}         & \textbf{Value} \\
\midrule
\multirow{2}{*}{Base model}     & \texttt{runwayml/stable-} \\
                                & \texttt{diffusion-v1-5} \\
Sampling steps                  & 15 \\
Dual learning rate              & 0.05 \\
Batch size (effective)          & $4 \times 16 = 64$ \\
Samples per epoch               & 128 \\
Epochs                          & 20 \\
KL penalty                      & 0.1 \\
LoRA rank                      & 4  \\
\multirow{3}{*}{Constraint level} 
    & MPS: 0.5 \\
    & Saturation: 0.5 \\
    & Local contrast: 0.25 \\
Equal weights & 0.2 \\
\bottomrule
\end{tabular}
\vspace{0.1cm}
\caption{Hyperparameters for reward alignment with contrast and saturation constraints. Constraint levels correspond to normalized rewards.}
\label{tab:reward saturation hparams}
\end{table}

\textbf{Additional results}. We include images sampled from the constrained model in Figure~\ref{fig:reward_prod_images_mps} for hps and aesthetic reward functions. Samples from a model trained with an equally weighted model are included for comparison. Constraints prevent overfitting to the saturation and smoothness penalties.




\textbf{II. Multiple aesthetic constraints}

\textbf{Implementation details and hyperparameters}. We modified the Alignprop framework to accomodate Algorithm~\ref{alg: alignment}. Following their setup, we use text conditioning on prompts of simple animals, using separate sets for training and evaluation. In this setting, due to the high variability of rewards throughout training, utilized an exponential moving average to reduce the variance in slack  estimates (and hence dual subgradients)~\cite{sohrabi2024pi}. Hyperparameters are detailed in Table~\ref{tab: multi reward hparams}.

\begin{table}[h]
\centering
\begin{tabular}{ll}
\toprule
\textbf{Hyperparameter}         & \textbf{Value} \\
\midrule
\multirow{2}{*}{Base model}     & \texttt{runwayml/stable-} \\
                                & \texttt{diffusion-v1-5} \\
Sampling steps                  & 15 \\
Dual learning rate              & 0.05 \\
Batch size (effective)          & $4 \times 16 = 64$ \\
Samples per epoch               & 128 \\
Epochs                          & 25 \\
KL penalty                      & 0.1 \\
LoRA rank                      & 4  \\
\multirow{5}{*}{Constraint level} 
    & MPS: 0.5 \\
    & HPS: 0.5 \\
    & Aesthetic: 0.5 \\
    & Pickscore : 0.5 \\
Equal weights & 0.2 \\
\multirow{8}{*}{Training Prompts} & \texttt{\{"cat", "dog", "horse", "monkey", "rabbit", "zebra"} \\
                                  & \texttt{"spider", "bird", "sheep", "deer", "cow", "goat"} \\
                                  & \texttt{"lion", "tiger", "bear", "raccoon", "fox", "wolf"} \\
                                  & \texttt{"lizard", "beetle", "ant", "butterfly", "fish", "shark"} \\
                                  & \texttt{"whale", "dolphin", "squirrel", "mouse", "rat", "snake"} \\
                                  & \texttt{"turtle", "frog", "chicken", "duck", "goose", "bee"} \\
                                  & \texttt{"pig", "turkey", "fly", "llama", "camel", "bat"} \\
                                  & \texttt{"gorilla", "hedgehog", "kangaroo"\}} \\

\multirow{2}{*}{Evaluation Prompts}    & \texttt{\{"cheetah", "snail", "hippopotamus",} \\
                            & \texttt{"crocodile", "lobster", "octopus"\}} \\
\bottomrule
\end{tabular}
\vspace{0.1cm}
\caption{Hyperparameters for reward alignment with multiple rewards. Constraint levels correspond to normalised rewards.}
\label{tab: multi reward hparams}
\end{table}

\textbf{Additional results}. We include two images per method and prompt in Figure~\ref{fig:reward_multiple_images}. These are sampled from the same latents for both models.

\subsection{Combining constrained alignment and composition}

As mentioned in Section~\ref{sec_composition_and_alignement}, The constrained alignment and composition problem formulations can  be combined. An example of this is composing reward-specialized models while enforcing a minimum aggregate reward level. To demonstrate the viability of this approach, we conducted a simple experiment in which we finetune a pretrained model using two KL constraints: one for pretrained stable diffusion and one for another model finetuned on the aesthetics reward, respectively, along with a constraint on the saturation reward. The finetuned model achieves less than 10\% reward constraint violation and similar KL divergences with respect to both pretrained models, as seen in Table~\ref{tab: combined experiment}. We leave in-depth exploration of the combined Alignment and Composition problem to future work. 

\begin{table}[H]
\centering
\begin{tabular}{lcccc}
\hline
\textbf{Constraint} & \textbf{Dual Variable} & \textbf{Initial Value} & \textbf{Final Value} & \textbf{Slack} \\
\hline
Saturation & 1.080 & 0.100 & 0.533 & 0.033 \\
KL (pretrained) & 0.493 & 0.000 & 0.101 & 0.004 \\
KL (aesthetics) & 0.507 & 0.260 & 0.097 & 0.000 \\
\hline
\end{tabular}
\vspace{0.1cm}
\caption{Results for finetuning a model with both expected reward and KL divergence constraints.}
\label{tab: combined experiment}
\end{table}

\subsection{Computational details}

All experiments were run on a single Nvidia A6000 GPU.

For alignment, there is little additional time overhead compared to baselines like AlignProp. For example, for the experiment in Figure~\ref{fig: mps alignment}, runtime is 33 minutes for both constrained and unconstrained methods, and for the experiments in Figure~\ref{fig: all rewards alignment}, constrained runtime is 64 minutes, unconstrained is 60 minutes. Existing approaches already estimate the KL and sample batches to evaluate and back-propagate through the reward. The only additional computation for our method is the dual updates which is negligible in terms of added time.

For composition, there is no meaningful comparison to the equal weights baseline since the weights are not learned in the equal weights baseline. For constrained composition, it takes around 5–10 dual updates for dual variables to converge which for composing the 5 finetuned stable diffusion models takes 9 minutes total and for concept composition it takes 2 minutes.

\newpage

\begin{table}[H]
\centering
\begin{tabular}{m{2.5cm} m{2.5cm} m{2.5cm} m{2.5cm}}
\multicolumn{2}{c}{\textbf{Equal Weights}} & \multicolumn{2}{c}{\textbf{Constrained}} \\[1ex]

\includegraphics[width=2.5cm]{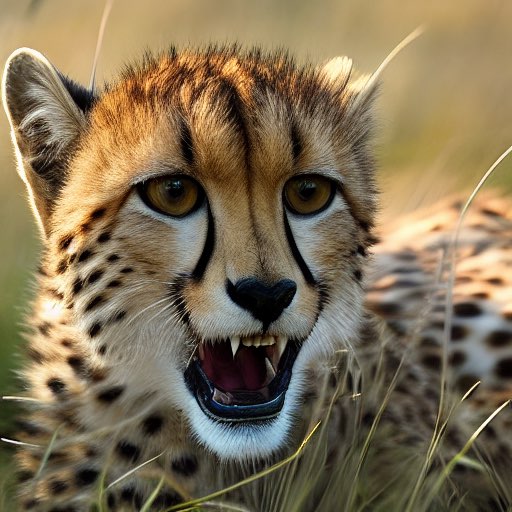} &
\includegraphics[width=2.5cm]{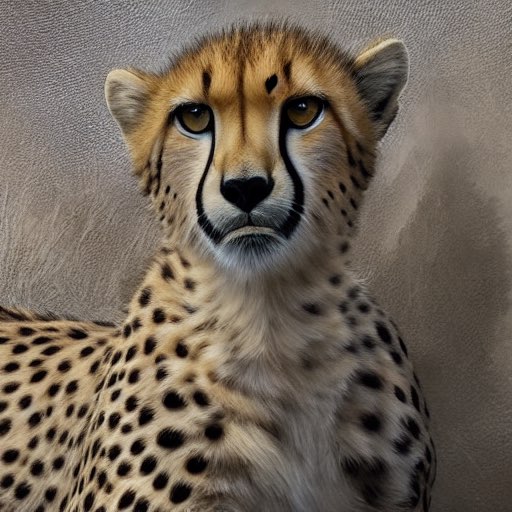} &
\includegraphics[width=2.5cm]{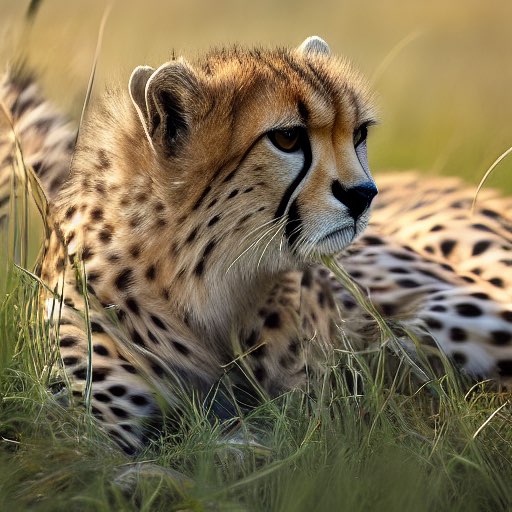} &
\includegraphics[width=2.5cm]{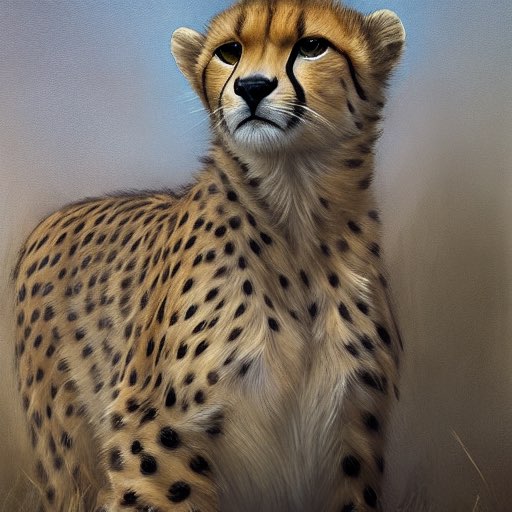} \\
\includegraphics[width=2.5cm]{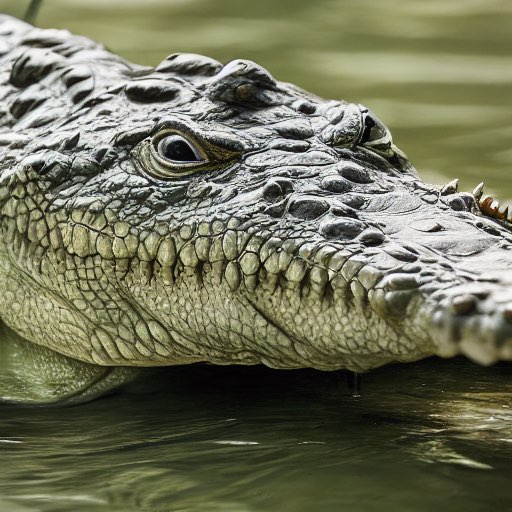} &
\includegraphics[width=2.5cm]{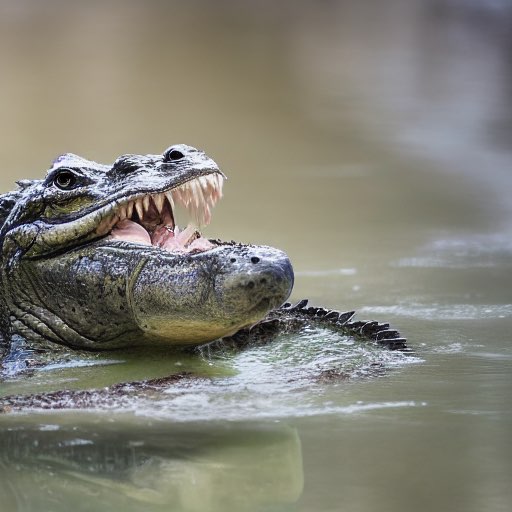} &
\includegraphics[width=2.5cm]{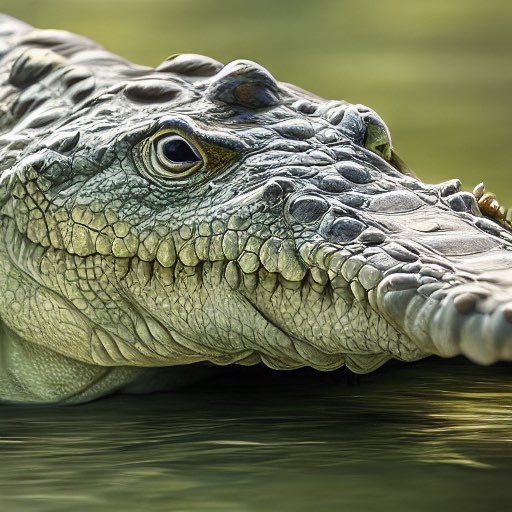} &
\includegraphics[width=2.5cm]{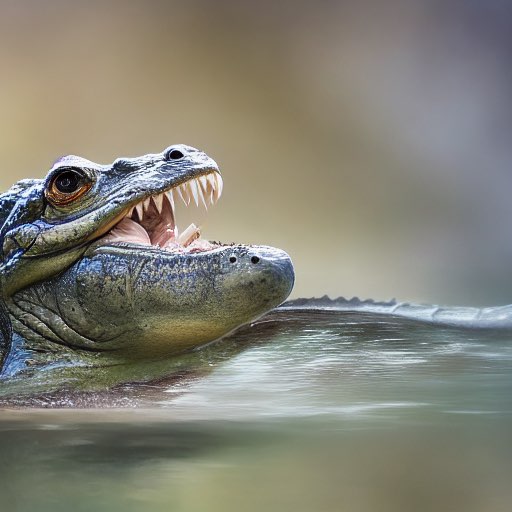} \\
\includegraphics[width=2.5cm]{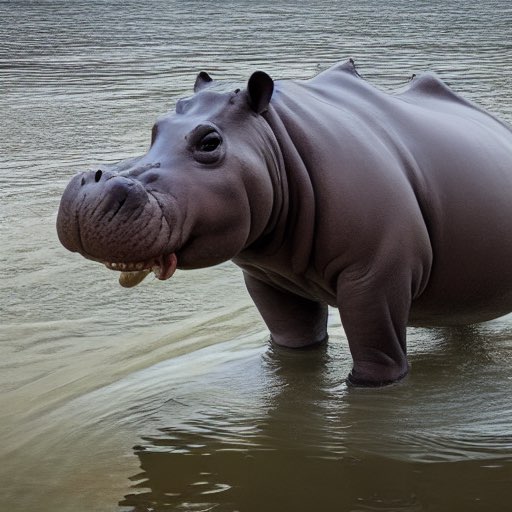} &
\includegraphics[width=2.5cm]{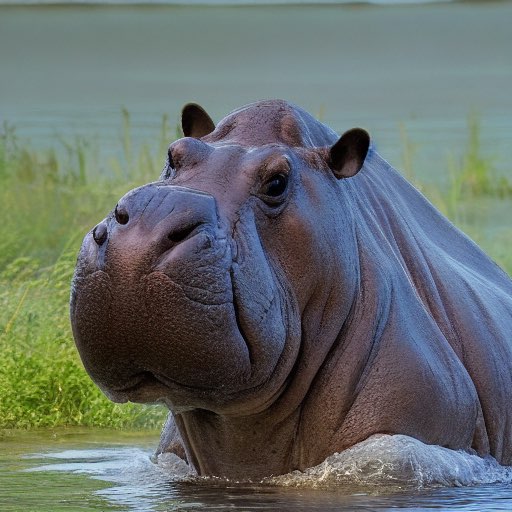} &
\includegraphics[width=2.5cm]{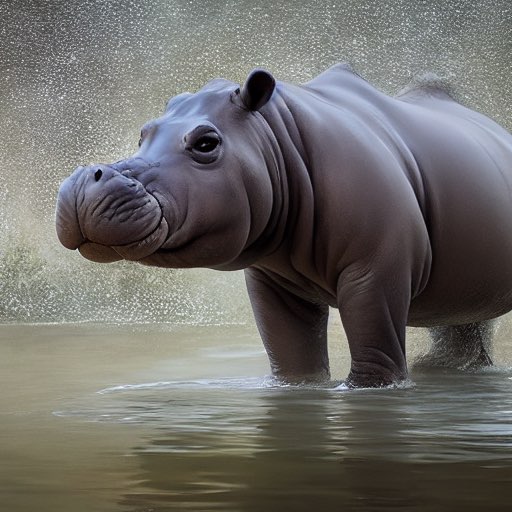} &
\includegraphics[width=2.5cm]{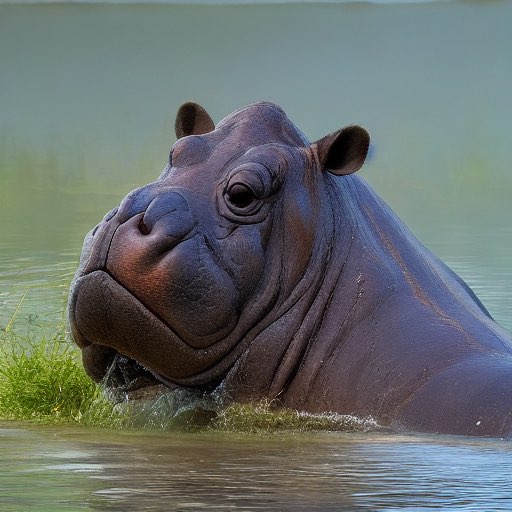} \\
\includegraphics[width=2.5cm]{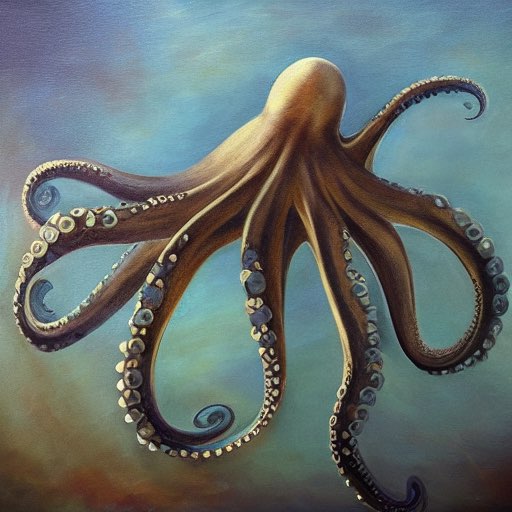} &
\includegraphics[width=2.5cm]{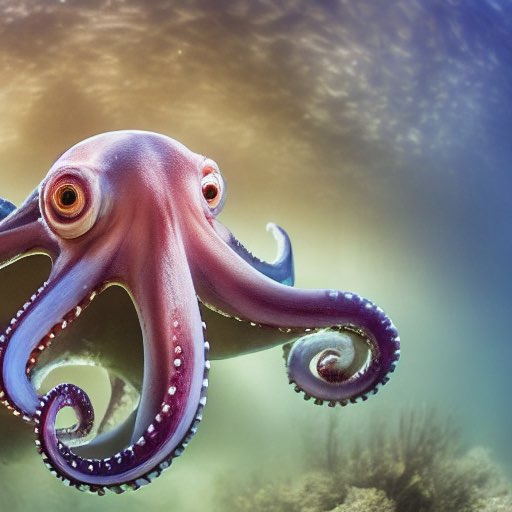} &
\includegraphics[width=2.5cm]{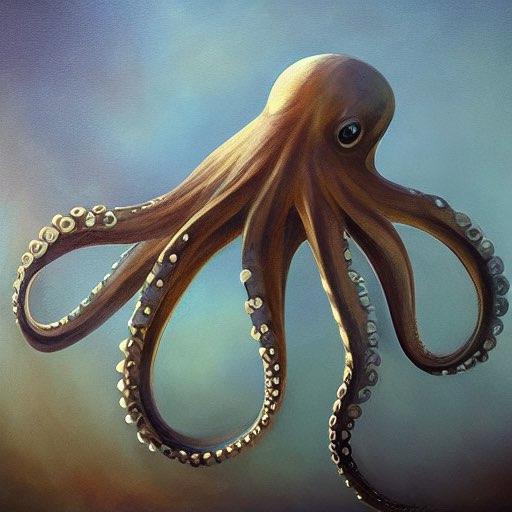} &
\includegraphics[width=2.5cm]{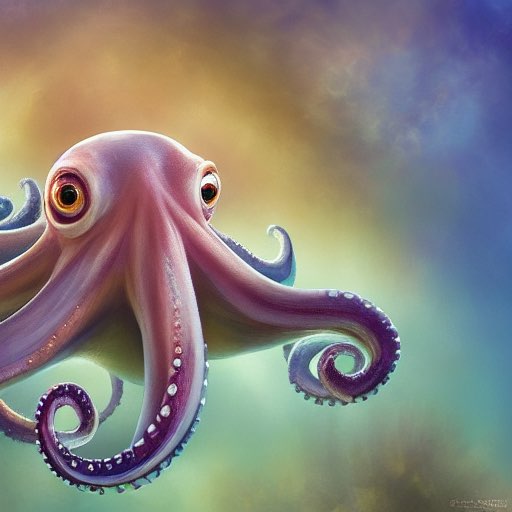} \\
\includegraphics[width=2.5cm]{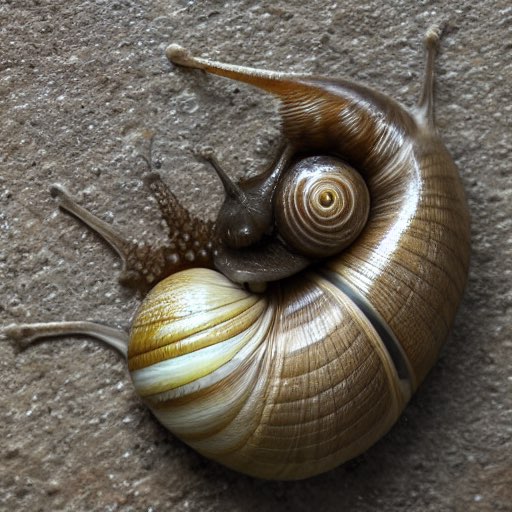} &
\includegraphics[width=2.5cm]{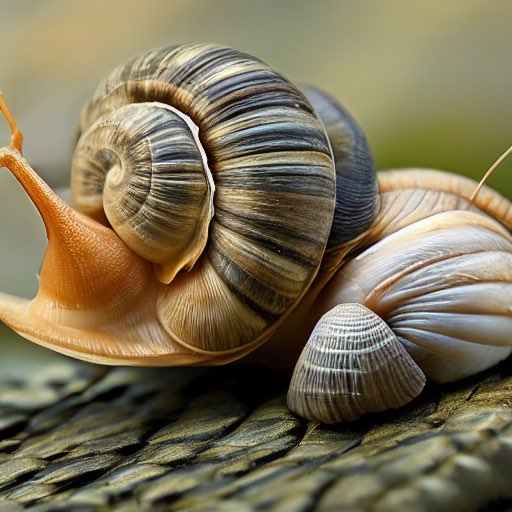}&
\includegraphics[width=2.5cm]{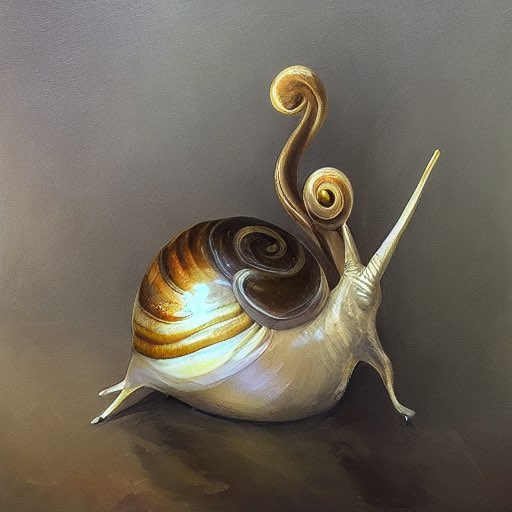} &
\includegraphics[width=2.5cm]{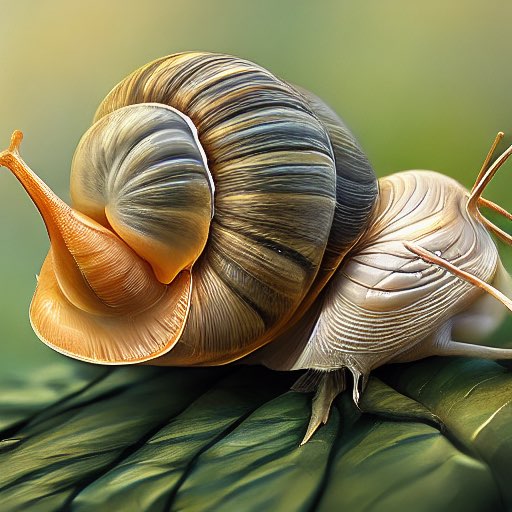} \\
\includegraphics[width=2.5cm]{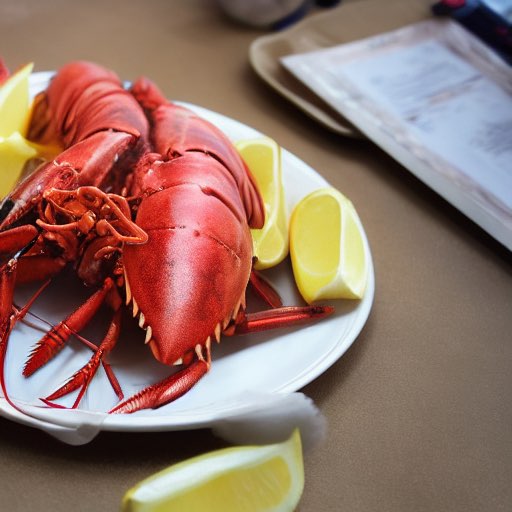} &
\includegraphics[width=2.5cm]{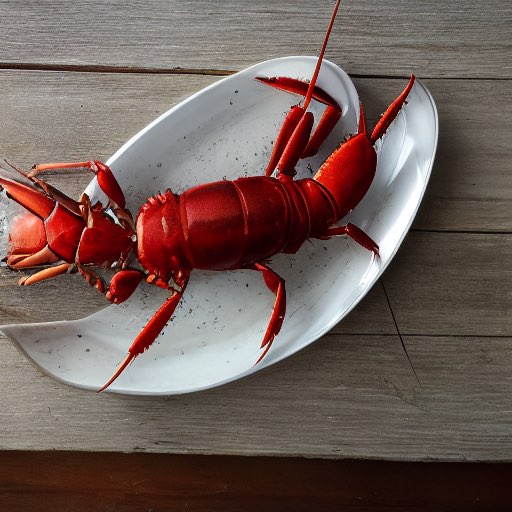} &
\includegraphics[width=2.5cm]{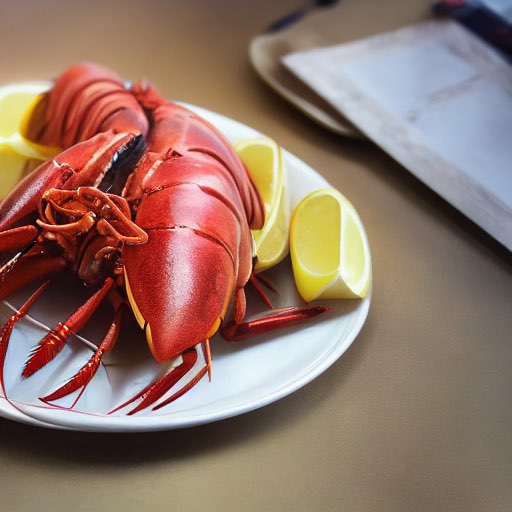} &
\includegraphics[width=2.5cm]{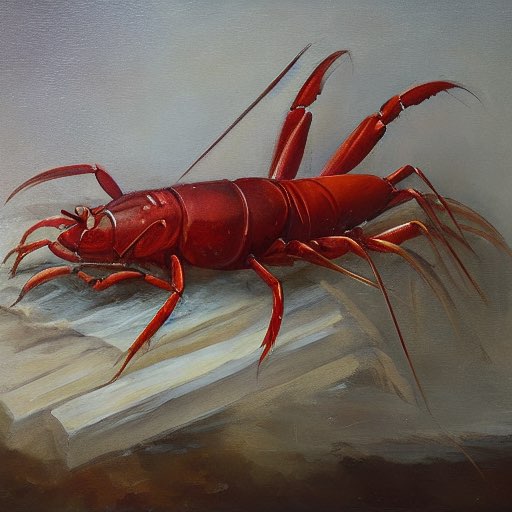} \\
\end{tabular}
\vspace{0.1cm}
\caption{Images sampled from the same latents for the product of adapters using the equal weights and when using the proposed KL-constrained reweighting scheme using 5 dual steps.}
\label{fig: reward prod images}
\end{table}

\begin{table}[H]
\centering
\begin{tabular}{m{2.5cm} m{2.5cm} m{2.5cm} m{2.5cm}}
\multicolumn{2}{c}{\textbf{Equal Weights}} & \multicolumn{2}{c}{\textbf{Constrained}} \\[1ex]

\includegraphics[width=2.5cm]{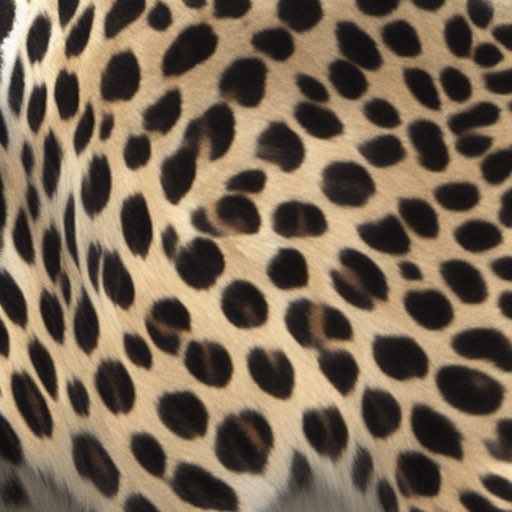} &
\includegraphics[width=2.5cm]{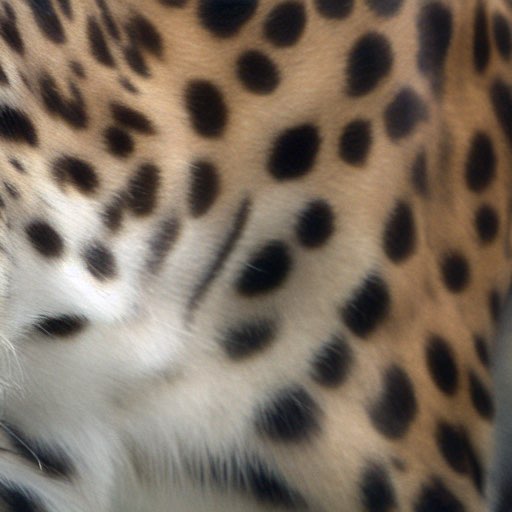} &
\includegraphics[width=2.5cm]{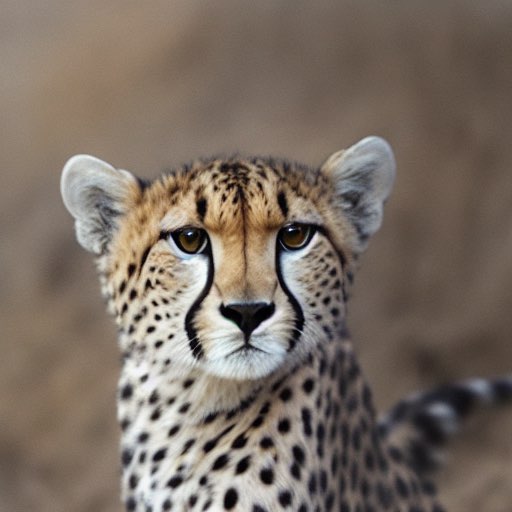} &
\includegraphics[width=2.5cm]{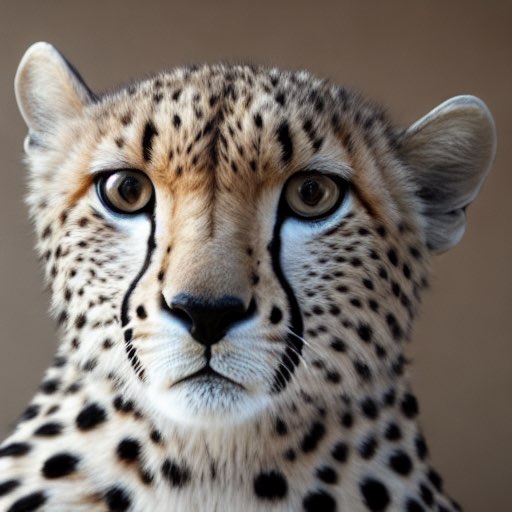} \\

\includegraphics[width=2.5cm]{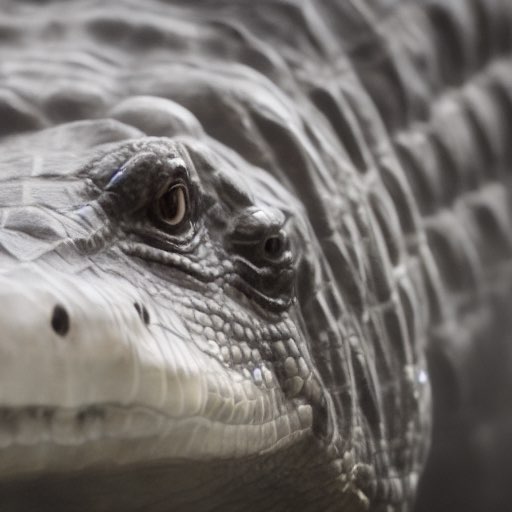} &
\includegraphics[width=2.5cm]{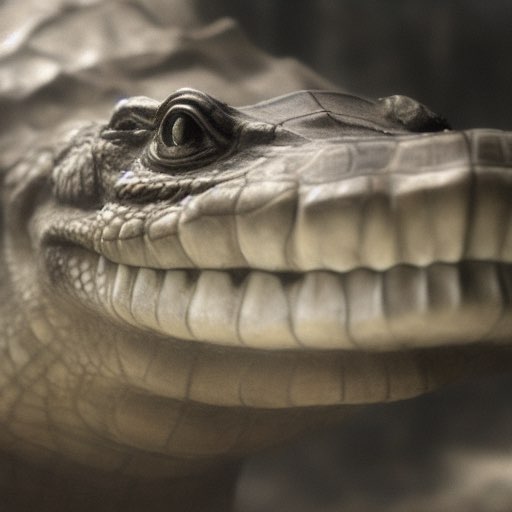} &
\includegraphics[width=2.5cm]{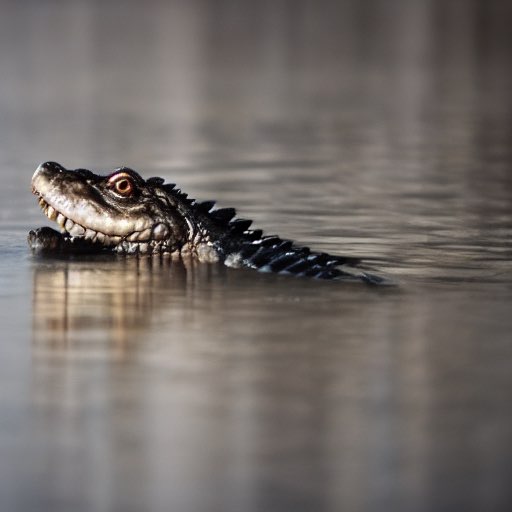} &
\includegraphics[width=2.5cm]{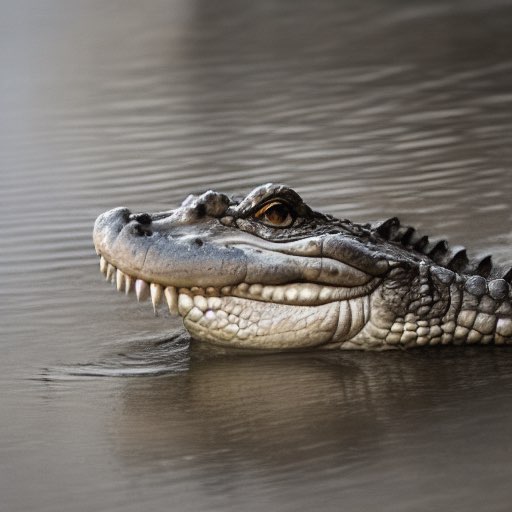} \\

\includegraphics[width=2.5cm]{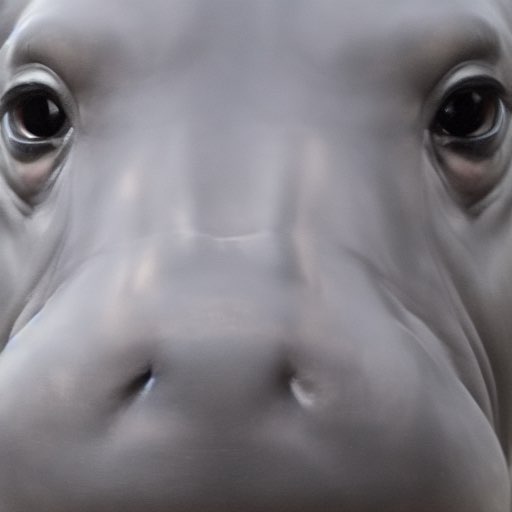} &
\includegraphics[width=2.5cm]{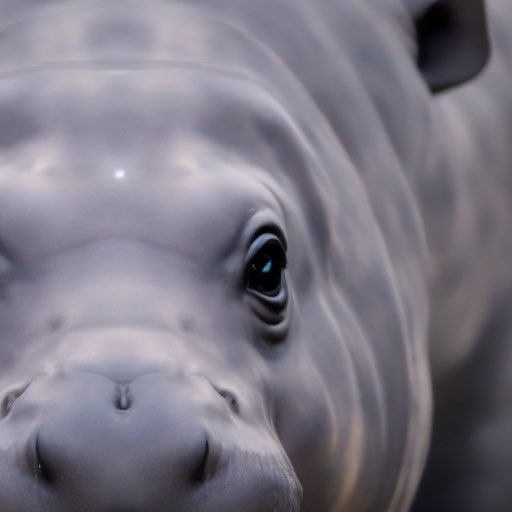} &
\includegraphics[width=2.5cm]{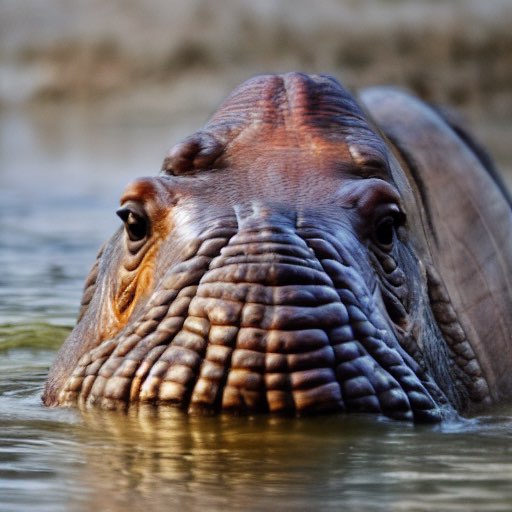} &
\includegraphics[width=2.5cm]{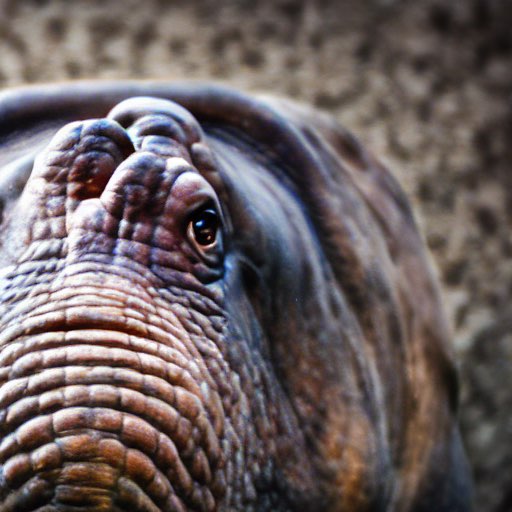} \\

\includegraphics[width=2.5cm]{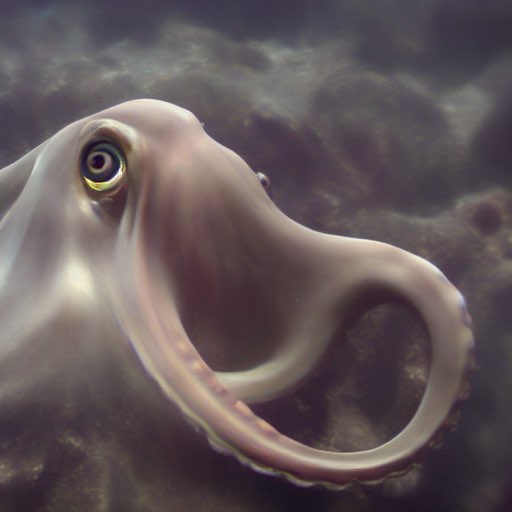} &
\includegraphics[width=2.5cm]{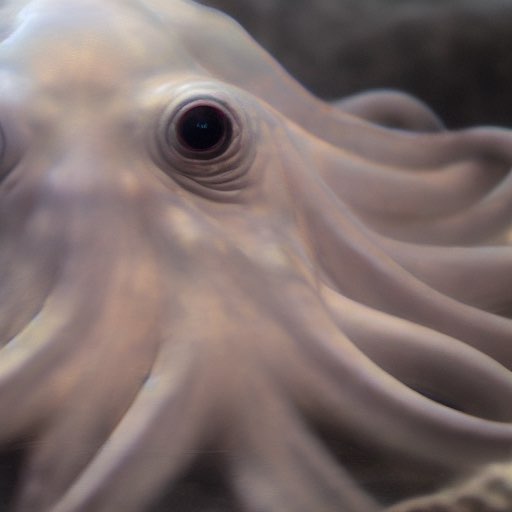} &
\includegraphics[width=2.5cm]{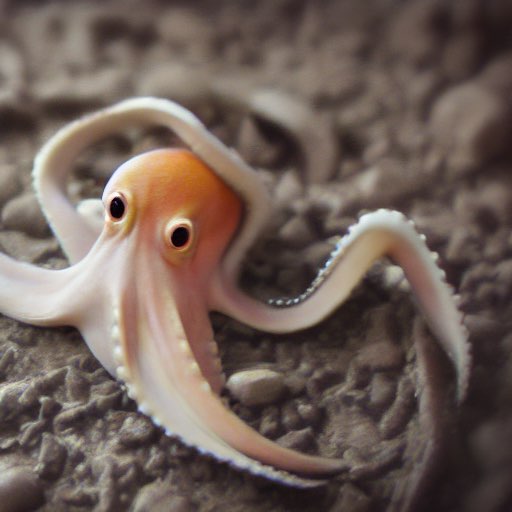} &
\includegraphics[width=2.5cm]{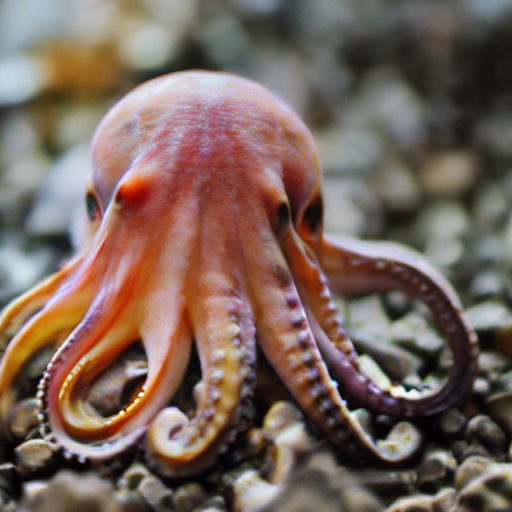} \\

\includegraphics[width=2.5cm]{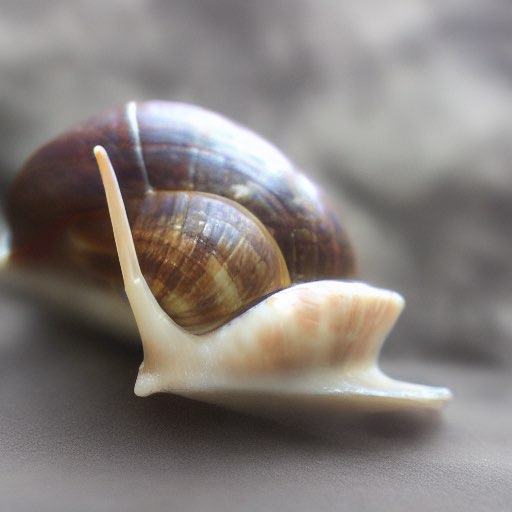} &
\includegraphics[width=2.5cm]{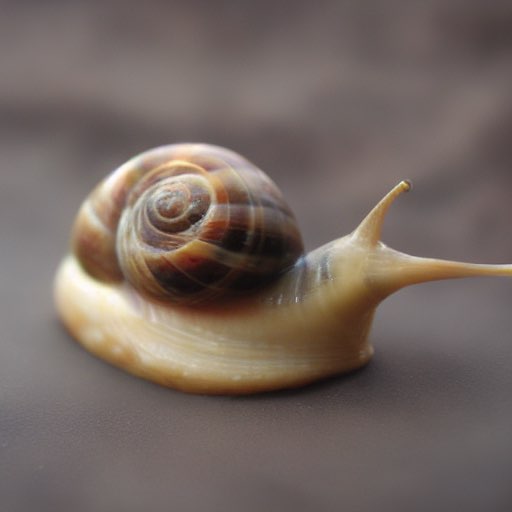} &
\includegraphics[width=2.5cm]{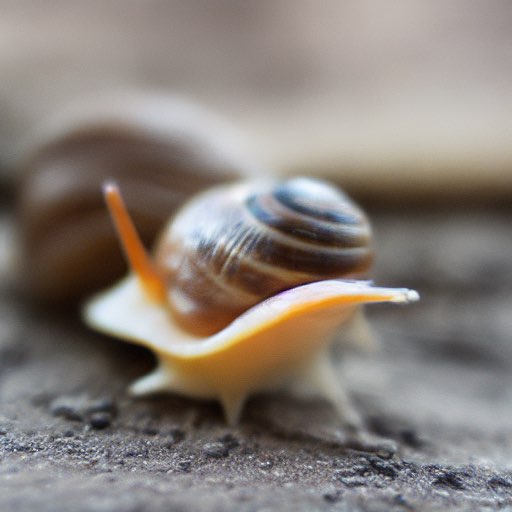} &
\includegraphics[width=2.5cm]{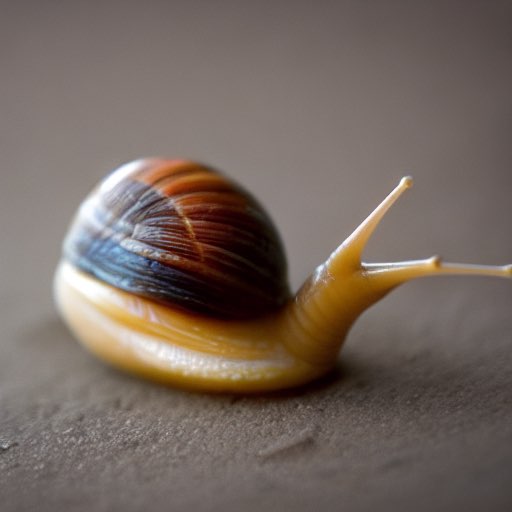} \\

\includegraphics[width=2.5cm]{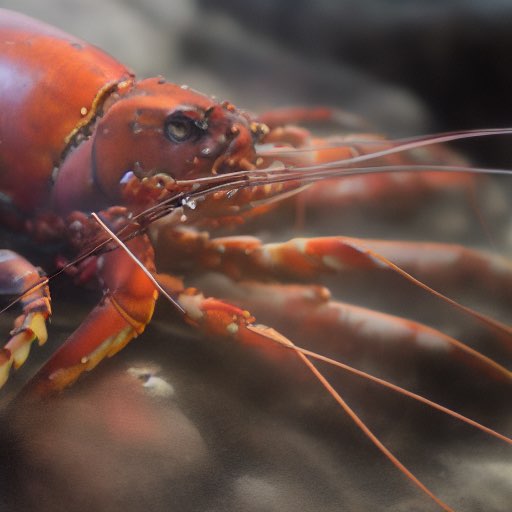} &
\includegraphics[width=2.5cm]{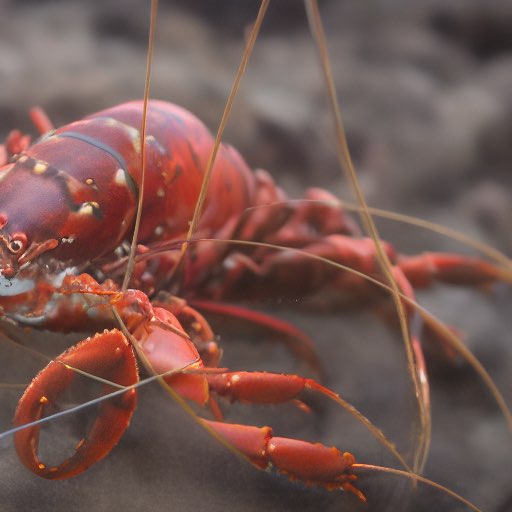} &
\includegraphics[width=2.5cm]{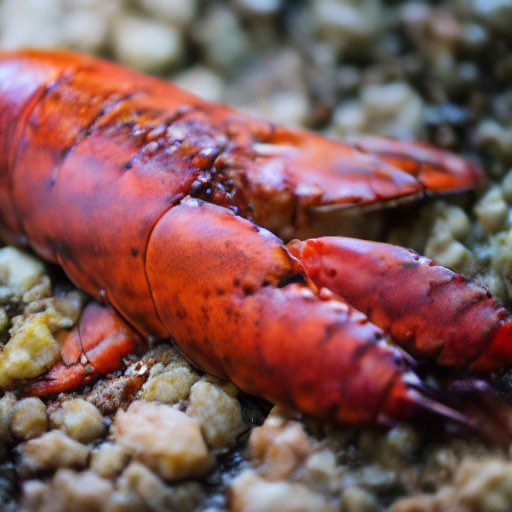} &
\includegraphics[width=2.5cm]{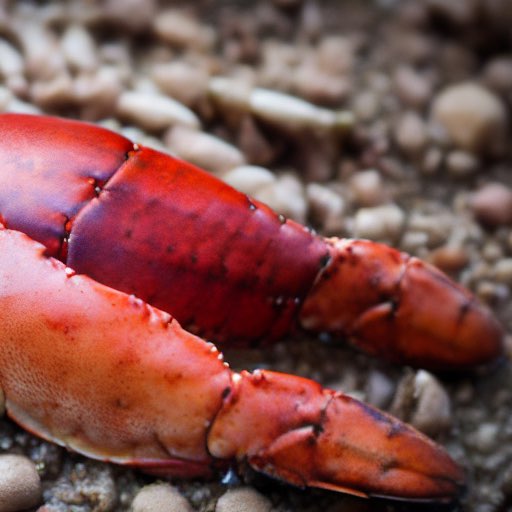}
\end{tabular}
\vspace{0.1cm}
\caption{Images sampled  from models finetuned to maximize MPS~\cite{mps}, along with sharpness and saturation penalizations. We compare optimizing an equally weighted objective against our constrained approach.}
\label{fig:reward_prod_images_mps}
\end{table}

\begin{table}[H]
\centering
\begin{tabular}{m{2.5cm} m{2.5cm} m{2.5cm} m{2.5cm}}
\multicolumn{2}{c}{\textbf{Equal Weights}} & \multicolumn{2}{c}{\textbf{Constrained}} \\[1ex]

\includegraphics[width=2.5cm]{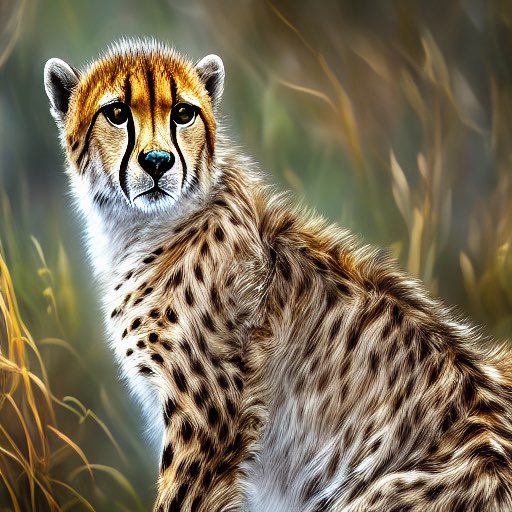} &
\includegraphics[width=2.5cm]{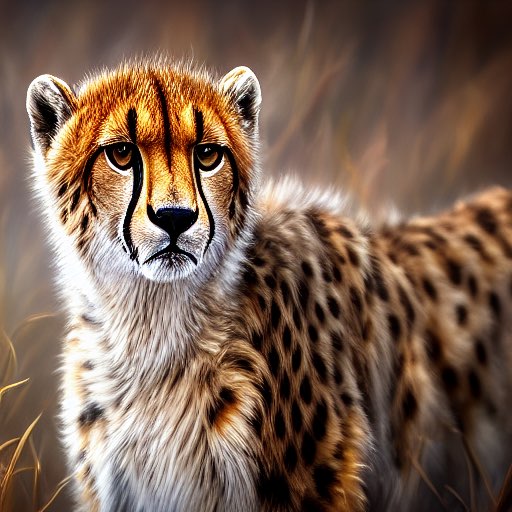} &
\includegraphics[width=2.5cm]{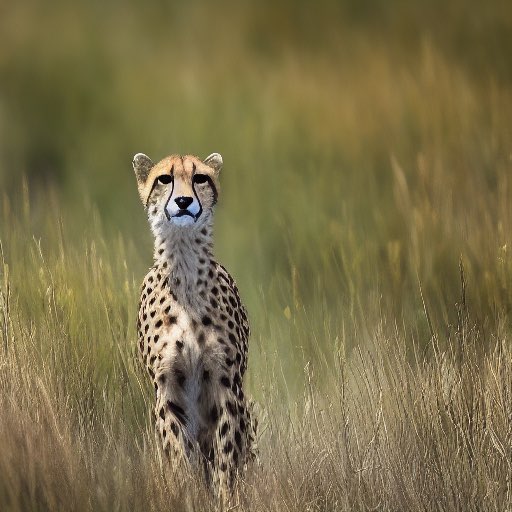} &
\includegraphics[width=2.5cm]{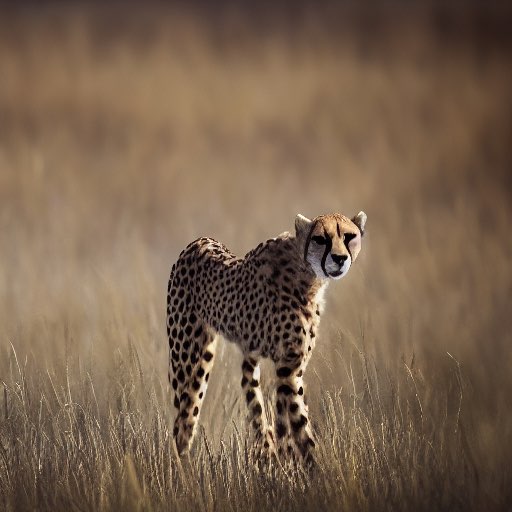} \\

\includegraphics[width=2.5cm]{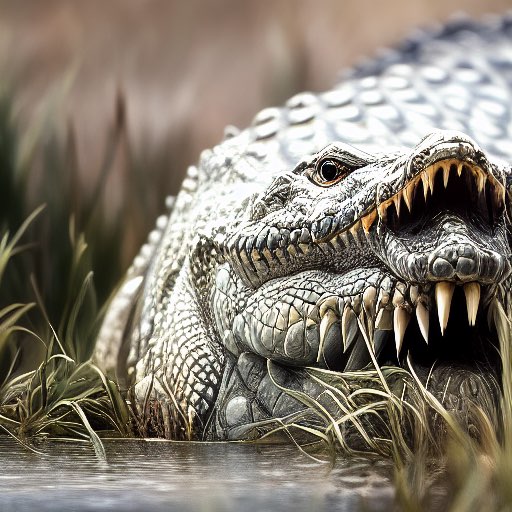} &
\includegraphics[width=2.5cm]{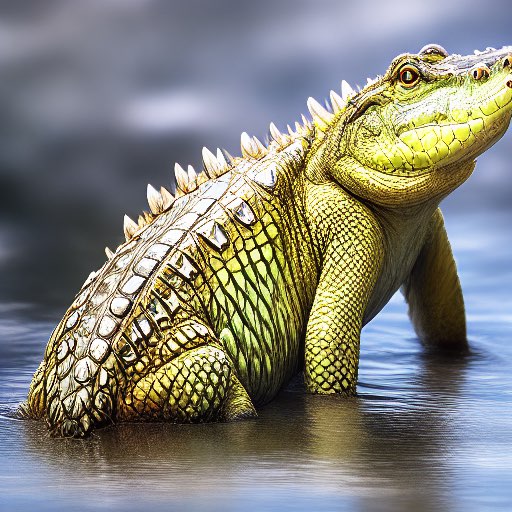} &
\includegraphics[width=2.5cm]{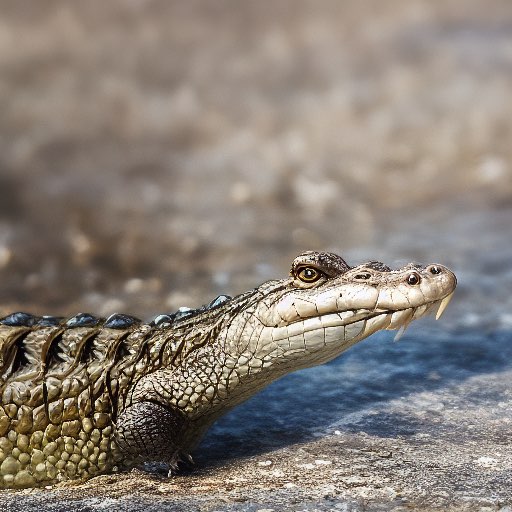} &
\includegraphics[width=2.5cm]{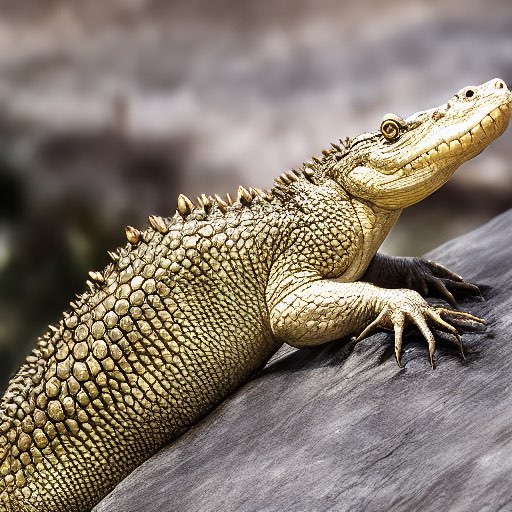} \\

\includegraphics[width=2.5cm]{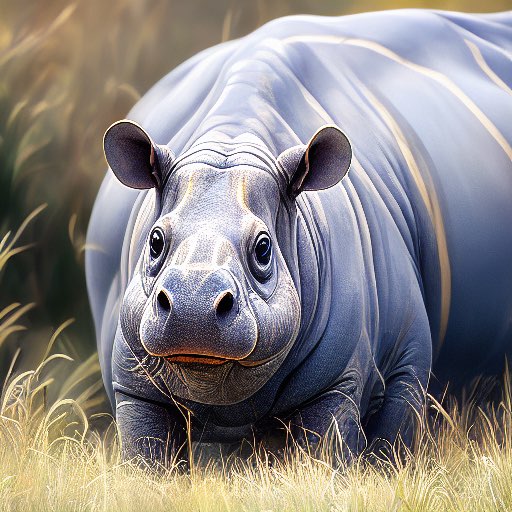} &
\includegraphics[width=2.5cm]{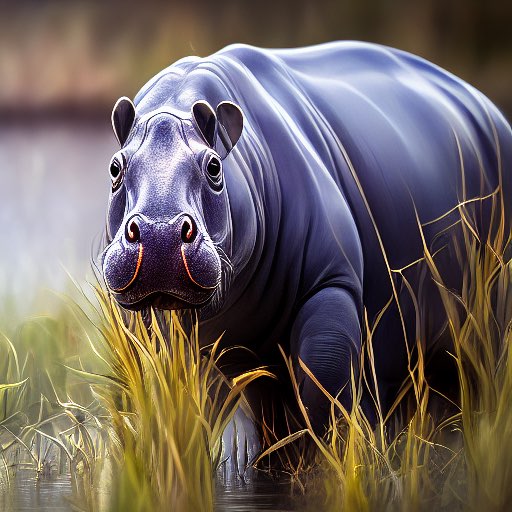} &
\includegraphics[width=2.5cm]{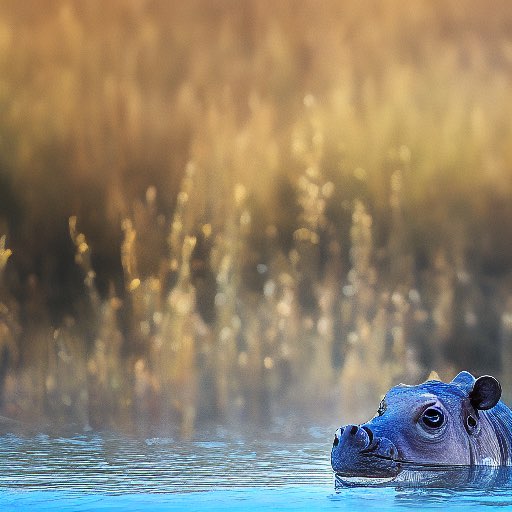} &
\includegraphics[width=2.5cm]{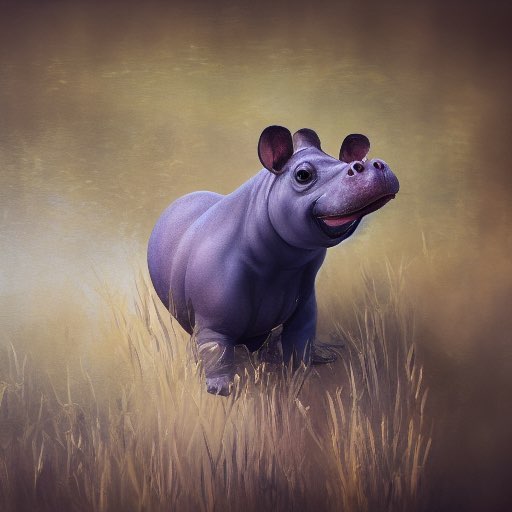} \\

\includegraphics[width=2.5cm]{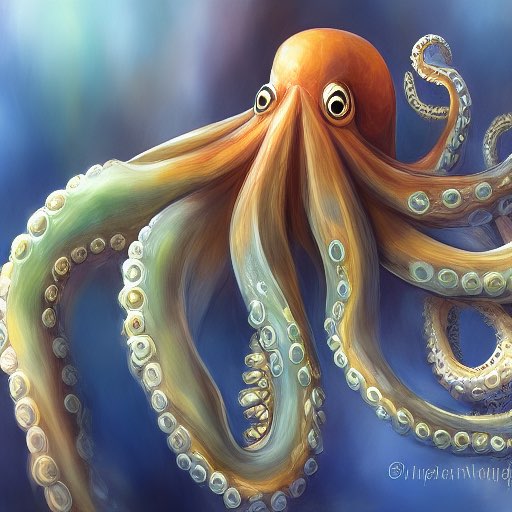} &
\includegraphics[width=2.5cm]{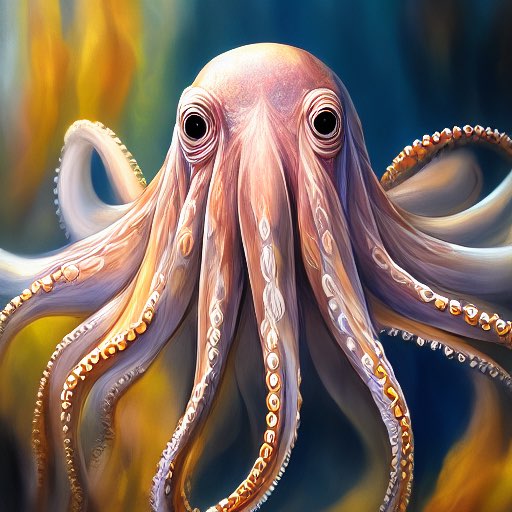} &
\includegraphics[width=2.5cm]{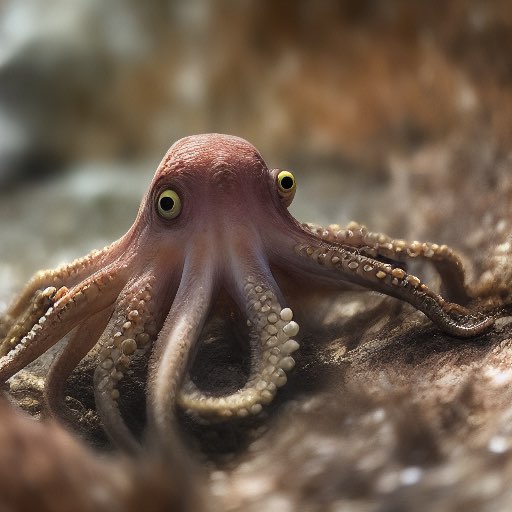} &
\includegraphics[width=2.5cm]{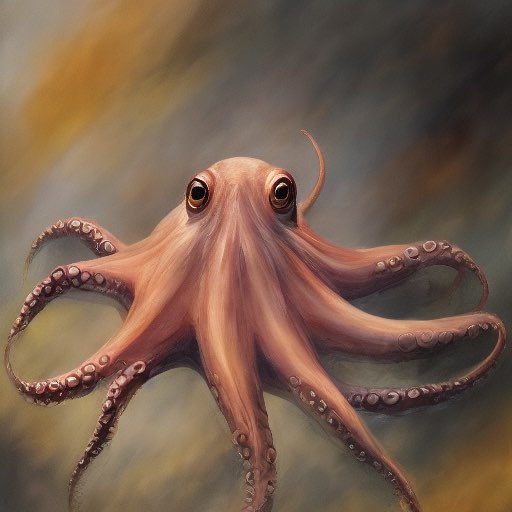} \\

\includegraphics[width=2.5cm]{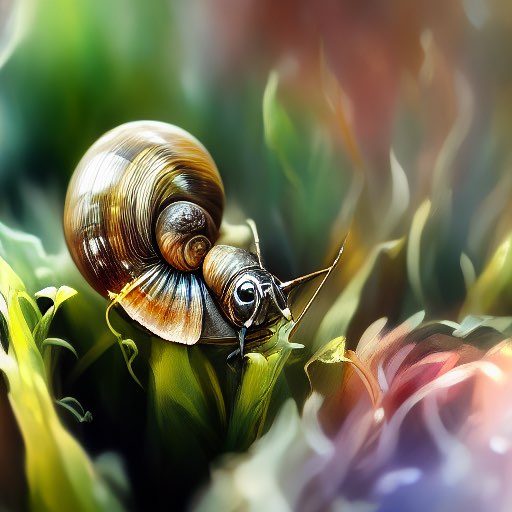} &
\includegraphics[width=2.5cm]{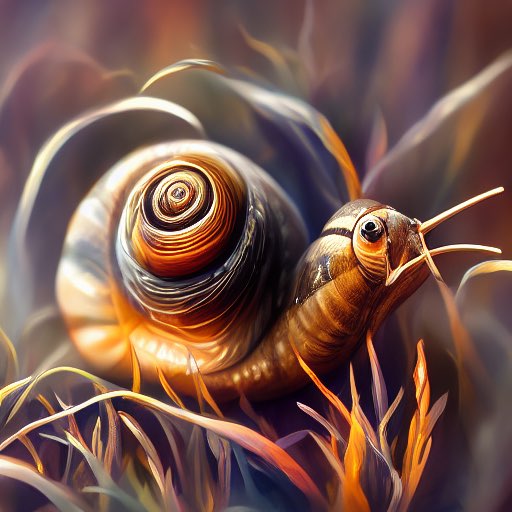} &
\includegraphics[width=2.5cm]{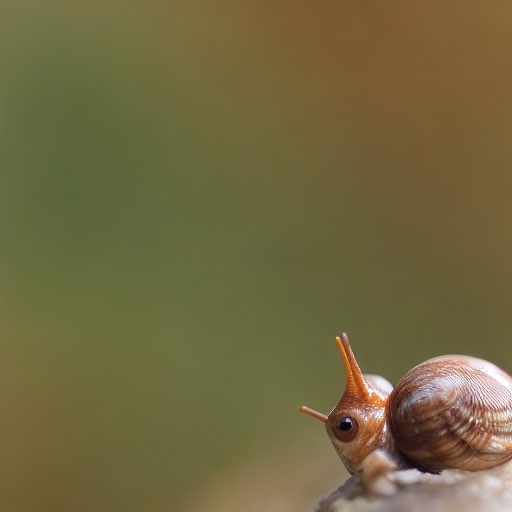} &
\includegraphics[width=2.5cm]{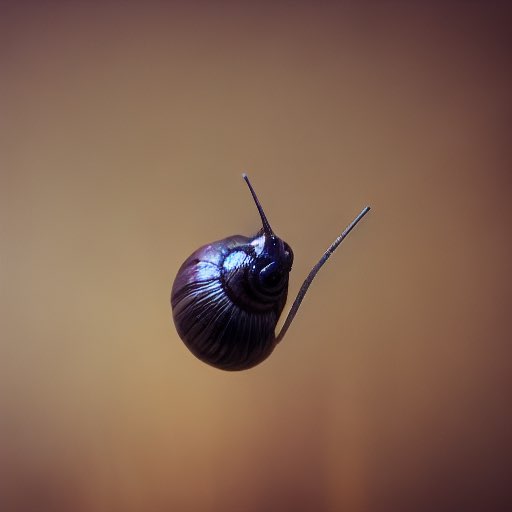} \\

\includegraphics[width=2.5cm]{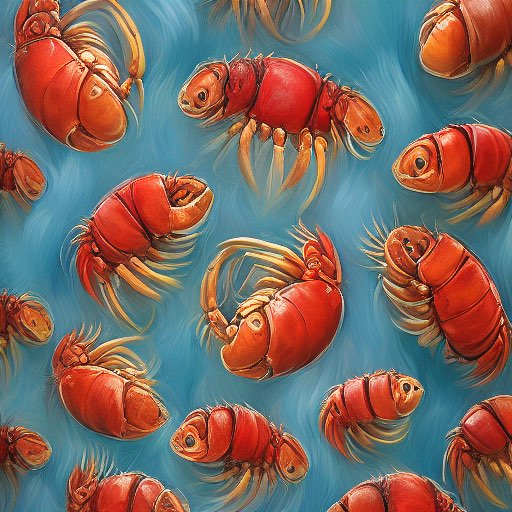} &
\includegraphics[width=2.5cm]{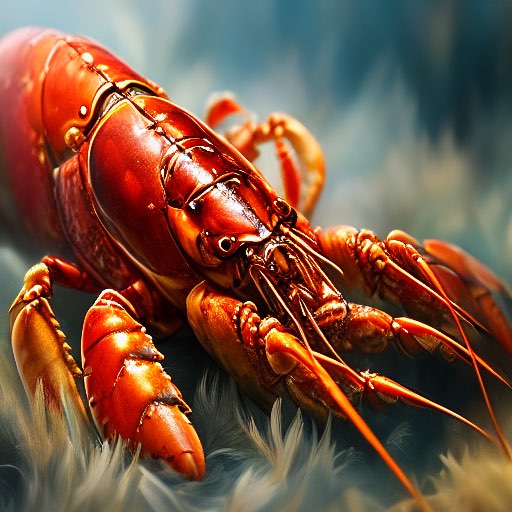} &
\includegraphics[width=2.5cm]{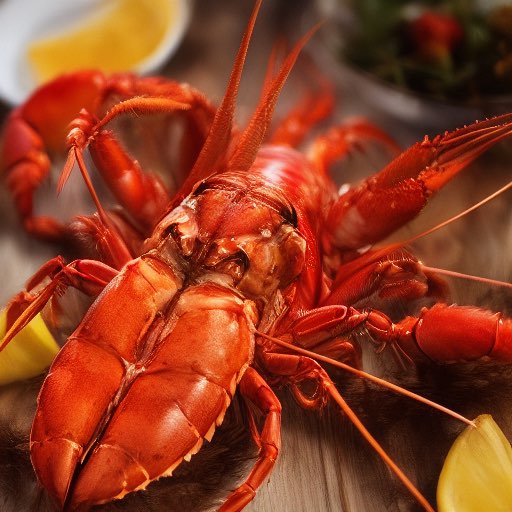} &
\includegraphics[width=2.5cm]{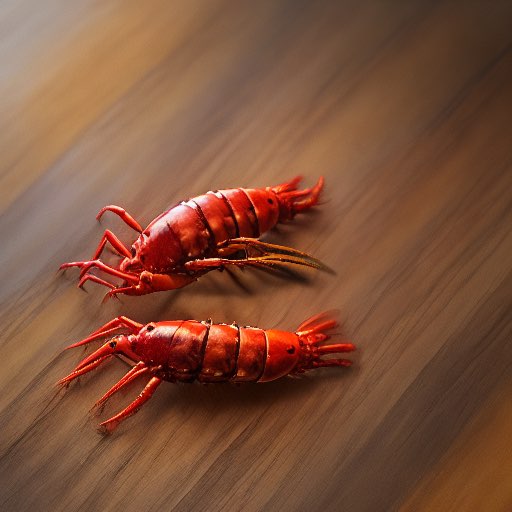}
\end{tabular}
\vspace{0.1cm}
\caption{Samples from models finetuned using  multiple rewards with equal weights and with our constrained alignment  method. }
\label{fig:reward_multiple_images}
\end{table}




%% file: dd-bib.bib
@article{bradley2024classifier,
  title={Classifier-free guidance is a predictor-corrector},
  author={Bradley, Arwen and Nakkiran, Preetum},
  journal={arXiv preprint arXiv:2408.09000},
  year={2024}
}

@article{chidambaram2024does,
  title={What does guidance do? a fine-grained analysis in a simple setting},
  author={Chidambaram, Muthu and Gatmiry, Khashayar and Chen, Sitan and Lee, Holden and Lu, Jianfeng},
  journal={arXiv preprint arXiv:2409.13074},
  year={2024}
}

@article{giannone2023aligning,
  title={Aligning optimization trajectories with diffusion models for constrained design generation},
  author={Giannone, Giorgio and Srivastava, Akash and Winther, Ole and Ahmed, Faez},
  journal={Advances in Neural Information Processing Systems},
  volume={36},
  pages={51830--51861},
  year={2023}
}

@inproceedings{chen2024towards,
  title={Towards Aligned Layout Generation via Diffusion Model with Aesthetic Constraints},
  author={Chen, Jian and Zhang, Ruiyi and Zhou, Yufan and Chen, Changyou},
  booktitle={The Twelfth International Conference on Learning Representations},
year={2024}
}

@misc{liu2023audioldmtexttoaudiogenerationlatent,
      title={AudioLDM: Text-to-Audio Generation with Latent Diffusion Models}, 
      author={Haohe Liu and Zehua Chen and Yi Yuan and Xinhao Mei and Xubo Liu and Danilo Mandic and Wenwu Wang and Mark D. Plumbley},
      year={2023},
      eprint={2301.12503},
      archivePrefix={arXiv},
      primaryClass={cs.SD},
      url={https://arxiv.org/abs/2301.12503}, 
}

@misc{elizalde2022claplearningaudioconcepts,
      title={CLAP: Learning Audio Concepts From Natural Language Supervision}, 
      author={Benjamin Elizalde and Soham Deshmukh and Mahmoud Al Ismail and Huaming Wang},
      year={2022},
      eprint={2206.04769},
      archivePrefix={arXiv},
      primaryClass={cs.SD},
      url={https://arxiv.org/abs/2206.04769}, 
}

@inproceedings{li2024aligning,
  title={Aligning Diffusion Models by Optimizing Human Utility},
  author={Li, Shufan and Kallidromitis, Konstantinos and Gokul, Akash and Kato, Yusuke and Kozuka, Kazuki},
  booktitle={The Thirty-eighth Annual Conference on Neural Information Processing Systems},
year={2024}
}

@inproceedings{yang2024using,
  title={Using human feedback to fine-tune diffusion models without any reward model},
  author={Yang, Kai and Tao, Jian and Lyu, Jiafei and Ge, Chunjiang and Chen, Jiaxin and Shen, Weihan and Zhu, Xiaolong and Li, Xiu},
  booktitle={Proceedings of the IEEE/CVF Conference on Computer Vision and Pattern Recognition},
  pages={8941--8951},
  year={2024}
}

@article{han2024stochastic,
  title={Stochastic Control for Fine-tuning Diffusion Models: Optimality, Regularity, and Convergence},
  author={Han, Yinbin and Razaviyayn, Meisam and Xu, Renyuan},
  journal={arXiv preprint arXiv:2412.18164},
  year={2024}
}

@inproceedings{fan2023optimizing,
  title={Optimizing {DDPM} Sampling with Shortcut Fine-Tuning},
  author={Fan, Ying and Lee, Kangwook},
  booktitle={International Conference on Machine Learning},
  pages={9623--9639},
  year={2023},
  organization={PMLR}
}

@misc{zhang2023controllable,
      title={Towards Controllable Diffusion Models via Reward-Guided Exploration}, 
      author={Hengtong Zhang and Tingyang Xu},
      year={2023},
      eprint={2304.07132}
}

@article{zhang2024aligning,
  title={Aligning Few-Step Diffusion Models with Dense Reward Difference Learning},
  author={Zhang, Ziyi and Shen, Li and Zhang, Sen and Ye, Deheng and Luo, Yong and Shi, Miaojing and Du, Bo and Tao, Dacheng},
  journal={arXiv preprint arXiv:2411.11727},
  year={2024}
}

@article{prabhudesai2024video,
  title={Video diffusion alignment via reward gradients},
  author={Prabhudesai, Mihir and Mendonca, Russell and Qin, Zheyang and Fragkiadaki, Katerina and Pathak, Deepak},
  journal={arXiv preprint arXiv:2407.08737},
  year={2024}
}

@misc{mou2019improvedboundsdiscretizationlangevin,
      title={Improved Bounds for Discretization of Langevin Diffusions: Near-Optimal Rates without Convexity}, 
      author={Wenlong Mou and Nicolas Flammarion and Martin J. Wainwright and Peter L. Bartlett},
      year={2019},
      eprint={1907.11331},
      archivePrefix={arXiv},
      primaryClass={math.PR},
      url={https://arxiv.org/abs/1907.11331}, 
}

@inproceedings{uehara2024feedback,
  title={Feedback Efficient Online Fine-Tuning of Diffusion Models},
  author={Uehara, Masatoshi and Zhao, Yulai and Black, Kevin and Hajiramezanali, Ehsan and Scalia, Gabriele and Diamant, Nathaniel Lee and Tseng, Alex M and Levine, Sergey and Biancalani, Tommaso},
  booktitle={Forty-first International Conference on Machine Learning},
  year={2024}
}

@article{uehara2024bridging,
  title={Bridging model-based optimization and generative modeling via conservative fine-tuning of diffusion models},
  author={Uehara, Masatoshi and Zhao, Yulai and Hajiramezanali, Ehsan and Scalia, Gabriele and Eraslan, Gokcen and Lal, Avantika and Levine, Sergey and Biancalani, Tommaso},
  journal={Advances in Neural Information Processing Systems},
  volume={37},
  pages={127511--127535},
  year={2024}
}

@article{zhao2024scores,
  title={Scores as Actions: a framework of fine-tuning diffusion models by continuous-time reinforcement learning},
  author={Zhao, Hanyang and Chen, Haoxian and Zhang, Ji and Yao, David D and Tang, Wenpin},
  journal={arXiv preprint arXiv:2409.08400},
  year={2024}
}

@article{uehara2024fine,
  title={Fine-tuning of continuous-time diffusion models as entropy-regularized control},
  author={Uehara, Masatoshi and Zhao, Yulai and Black, Kevin and Hajiramezanali, Ehsan and Scalia, Gabriele and Diamant, Nathaniel Lee and Tseng, Alex M and Biancalani, Tommaso and Levine, Sergey},
  journal={arXiv preprint arXiv:2402.15194},
  year={2024}
}

@inproceedings{wu2023human,
  title={Human preference score: Better aligning text-to-image models with human preference},
  author={Wu, Xiaoshi and Sun, Keqiang and Zhu, Feng and Zhao, Rui and Li, Hongsheng},
  booktitle={Proceedings of the IEEE/CVF International Conference on Computer Vision},
  pages={2096--2105},
  year={2023}
}

@article{lee2023aligning,
  title={Aligning text-to-image models using human feedback},
  author={Lee, Kimin and Liu, Hao and Ryu, Moonkyung and Watkins, Olivia and Du, Yuqing and Boutilier, Craig and Abbeel, Pieter and Ghavamzadeh, Mohammad and Gu, Shixiang Shane},
  journal={arXiv preprint arXiv:2302.12192},
  year={2023}
}

@inproceedings{wu2024deep,
  title={Deep reward supervisions for tuning text-to-image diffusion models},
  author={Wu, Xiaoshi and Hao, Yiming and Zhang, Manyuan and Sun, Keqiang and Huang, Zhaoyang and Song, Guanglu and Liu, Yu and Li, Hongsheng},
  booktitle={European Conference on Computer Vision},
  pages={108--124},
  year={2024}
}

@inproceedings{clark2024directly,
  title={Directly Fine-Tuning Diffusion Models on Differentiable Rewards},
  author={Clark, Kevin and Vicol, Paul and Swersky, Kevin and Fleet, David J},
  booktitle={The Twelfth International Conference on Learning Representations},
  year={2024}
}

@inproceedings{black2024training,
  title={Training Diffusion Models with Reinforcement Learning},
  author={Black, Kevin and Janner, Michael and Du, Yilun and Kostrikov, Ilya and Levine, Sergey},
  booktitle={The Twelfth International Conference on Learning Representations},
  year={2024}
}

@article{liang2024multi,
  title={Multi-Agent Path Finding in Continuous Spaces with Projected Diffusion Models},
  author={Liang, Jinhao and Christopher, Jacob K and Koenig, Sven and Fioretto, Ferdinando},
  journal={arXiv preprint arXiv:2412.17993},
  year={2024}
}

@article{liang2025simultaneous,
  title={Simultaneous Multi-Robot Motion Planning with Projected Diffusion Models},
  author={Liang, Jinhao and Christopher, Jacob K and Koenig, Sven and Fioretto, Ferdinando},
  journal={arXiv preprint arXiv:2502.03607},
  year={2025}
}

@article{christopher2024constrained,
  title={Constrained synthesis with projected diffusion models},
  author={Christopher, Jacob K and Baek, Stephen and Fioretto, Nando},
  journal={Advances in Neural Information Processing Systems},
  volume={37},
  pages={89307--89333},
  year={2024}
}

@article{zampini2025training,
  title={Training-Free Constrained Generation With Stable Diffusion Models},
  author={Zampini, Stefano and Christopher, Jacob and Oneto, Luca and Anguita, Davide and Fioretto, Ferdinando},
  journal={arXiv preprint arXiv:2502.05625},
  year={2025}
}

@article{narasimhan2024constrained,
  title={Constrained Posterior Sampling: Time Series Generation with Hard Constraints},
  author={Narasimhan, Sai Shankar and Agarwal, Shubhankar and Rout, Litu and Shakkottai, Sanjay and Chinchali, Sandeep P},
  journal={arXiv preprint arXiv:2410.12652},
  year={2024}
}

@inproceedings{khalafi2024constrained,
  title={Constrained Diffusion Models via Dual Training},
  author={Khalafi, Shervin and Ding, Dongsheng and Ribeiro, Alejandro},
  booktitle={The Thirty-eighth Annual Conference on Neural Information Processing Systems},
year={2024}
}

@article{liu2024alignment,
  title={Alignment of diffusion models: Fundamentals, challenges, and future},
  author={Liu, Buhua and Shao, Shitong and Li, Bao and Bai, Lichen and Xu, Zhiqiang and Xiong, Haoyi and Kwok, James and Helal, Sumi and Xie, Zeke},
  journal={arXiv preprint arXiv:2409.07253},
  year={2024}
}

@inproceedings{yan2024diffusion,
  title={Diffusion models without attention},
  author={Yan, Jing Nathan and Gu, Jiatao and Rush, Alexander M},
  booktitle={Proceedings of the IEEE/CVF Conference on Computer Vision and Pattern Recognition},
  pages={8239--8249},
  year={2024}
}

@article{ulhaq2022efficient,
  title={Efficient diffusion models for vision: A survey},
  author={Ulhaq, Anwaar and Akhtar, Naveed},
  journal={arXiv preprint arXiv:2210.09292},
  year={2022}
}

@article{chamon2022constrained,
  title={Constrained learning with non-convex losses},
  author={Chamon, Luiz FO and Paternain, Santiago and Calvo-Fullana, Miguel and Ribeiro, Alejandro},
  journal={IEEE Transactions on Information Theory},
  volume={69},
  number={3},
  pages={1739--1760},
  year={2022},
  publisher={IEEE}
}

@book{boyd2004convex,
  title={Convex optimization},
  author={Boyd, Stephen P and Vandenberghe, Lieven},
  year={2004},
  publisher={Cambridge university press}
}

@article{chi2023diffusion,
  title={Diffusion policy: Visuomotor policy learning via action diffusion},
  author={Chi, Cheng and Xu, Zhenjia and Feng, Siyuan and Cousineau, Eric and Du, Yilun and Burchfiel, Benjamin and Tedrake, Russ and Song, Shuran},
  journal={The International Journal of Robotics Research},
  pages={02783649241273668},
  year={2023},
  publisher={SAGE Publications Sage UK: London, England}
}

@misc{hessel2022clipscorereferencefreeevaluationmetric,
      title={CLIPScore: A Reference-free Evaluation Metric for Image Captioning}, 
      author={Jack Hessel and Ari Holtzman and Maxwell Forbes and Ronan Le Bras and Yejin Choi},
      year={2022},
      eprint={2104.08718},
      archivePrefix={arXiv},
      primaryClass={cs.CV},
      url={https://arxiv.org/abs/2104.08718}, 
}

@misc{li2022blipbootstrappinglanguageimagepretraining,
      title={BLIP: Bootstrapping Language-Image Pre-training for Unified Vision-Language Understanding and Generation}, 
      author={Junnan Li and Dongxu Li and Caiming Xiong and Steven Hoi},
      year={2022},
      eprint={2201.12086},
      archivePrefix={arXiv},
      primaryClass={cs.CV},
      url={https://arxiv.org/abs/2201.12086}, 
}

@article{wang2025diffusion,
  title={Diffusion Models for Molecules: A Survey of Methods and Tasks},
  author={Wang, Liang and Song, Chao and Liu, Zhiyuan and Rong, Yu and Liu, Qiang and Wu, Shu},
  journal={arXiv preprint arXiv:2502.09511},
  year={2025}
}

@inproceedings{blattmann2023align,
  title={Align your latents: High-resolution video synthesis with latent diffusion models},
  author={Blattmann, Andreas and Rombach, Robin and Ling, Huan and Dockhorn, Tim and Kim, Seung Wook and Fidler, Sanja and Kreis, Karsten},
  booktitle={Proceedings of the IEEE/CVF conference on computer vision and pattern recognition},
  pages={22563--22575},
  year={2023}
}

@inproceedings{
skreta2025the,
title={The Superposition of Diffusion Models Using the It\^o Density Estimator},
author={Marta Skreta and Lazar Atanackovic and Joey Bose and Alexander Tong and Kirill Neklyudov},
booktitle={The Thirteenth International Conference on Learning Representations},
year={2025},
url={https://openreview.net/forum?id=2o58Mbqkd2}
}

@article{saharia2022photorealistic,
  title={Photorealistic text-to-image diffusion models with deep language understanding},
  author={Saharia, Chitwan and Chan, William and Saxena, Saurabh and Li, Lala and Whang, Jay and Denton, Emily L and Ghasemipour, Kamyar and Gontijo Lopes, Raphael and Karagol Ayan, Burcu and Salimans, Tim and others},
  journal={Advances in neural information processing systems},
  volume={35},
  pages={36479--36494},
  year={2022}
}

@article{uehara2024understanding,
  title={Understanding reinforcement learning-based fine-tuning of diffusion models: A tutorial and review},
  author={Uehara, Masatoshi and Zhao, Yulai and Biancalani, Tommaso and Levine, Sergey},
  journal={arXiv preprint arXiv:2407.13734},
  year={2024}
}

@article{biggs2024diffusion,
  title={Diffusion Soup: Model Merging for Text-to-Image Diffusion Models},
  author={Biggs, Benjamin and Seshadri, Arjun and Zou, Yang and Jain, Achin and Golatkar, Aditya and Xie, Yusheng and Achille, Alessandro and Swaminathan, Ashwin and Soatto, Stefano},
  journal={arXiv preprint arXiv:2406.08431},
  year={2024}
}

@article{xu2023imagereward,
  title={Imagereward: Learning and evaluating human preferences for text-to-image generation},
  author={Xu, Jiazheng and Liu, Xiao and Wu, Yuchen and Tong, Yuxuan and Li, Qinkai and Ding, Ming and Tang, Jie and Dong, Yuxiao},
  journal={Advances in Neural Information Processing Systems},
  volume={36},
  year={2023}
}

@inproceedings{wallace2024diffusion,
  title={Diffusion model alignment using direct preference optimization},
  author={Wallace, Bram and Dang, Meihua and Rafailov, Rafael and Zhou, Linqi and Lou, Aaron and Purushwalkam, Senthil and Ermon, Stefano and Xiong, Caiming and Joty, Shafiq and Naik, Nikhil},
  booktitle={Proceedings of the IEEE/CVF Conference on Computer Vision and Pattern Recognition},
  pages={8228--8238},
  year={2024}
}

@misc{prabhudesai2024aligningtexttoimagediffusionmodels,
      title={Aligning Text-to-Image Diffusion Models with Reward Backpropagation}, 
      author={Mihir Prabhudesai and Anirudh Goyal and Deepak Pathak and Katerina Fragkiadaki},
      year={2024},
      eprint={2310.03739},
      archivePrefix={arXiv},
      primaryClass={cs.CV},
      url={https://arxiv.org/abs/2310.03739}, 
}

@inproceedings{liu2022compositional,
  title={Compositional visual generation with composable diffusion models},
  author={Liu, Nan and Li, Shuang and Du, Yilun and Torralba, Antonio and Tenenbaum, Joshua B},
  booktitle={European Conference on Computer Vision},
  pages={423--439},
  year={2022},
  organization={Springer}
}

@misc{rombach2022highresolutionimagesynthesislatent,
      title={High-Resolution Image Synthesis with Latent Diffusion Models}, 
      author={Robin Rombach and Andreas Blattmann and Dominik Lorenz and Patrick Esser and Björn Ommer},
      year={2022},
      eprint={2112.10752},
      archivePrefix={arXiv},
      primaryClass={cs.CV},
      url={https://arxiv.org/abs/2112.10752}, 
}

@misc{song2021scorebasedgenerativemodelingstochastic,
      title={Score-Based Generative Modeling through Stochastic Differential Equations}, 
      author={Yang Song and Jascha Sohl-Dickstein and Diederik P. Kingma and Abhishek Kumar and Stefano Ermon and Ben Poole},
      year={2021},
      eprint={2011.13456},
      archivePrefix={arXiv},
      primaryClass={cs.LG},
      url={https://arxiv.org/abs/2011.13456}, 
}

@misc{song2022denoisingdiffusionimplicitmodels,
      title={Denoising Diffusion Implicit Models}, 
      author={Jiaming Song and Chenlin Meng and Stefano Ermon},
      year={2022},
      eprint={2010.02502},
      archivePrefix={arXiv},
      primaryClass={cs.LG},
      url={https://arxiv.org/abs/2010.02502}, 
}

@misc{chamon2021probablyapproximatelycorrectconstrained,
      title={Probably Approximately Correct Constrained Learning}, 
      author={Luiz F. O. Chamon and Alejandro Ribeiro},
      year={2021},
      eprint={2006.05487},
      archivePrefix={arXiv},
      primaryClass={cs.LG},
      url={https://arxiv.org/abs/2006.05487}, 
}

@misc{lyu2012interpretationgeneralizationscorematching,
      title={Interpretation and Generalization of Score Matching}, 
      author={Siwei Lyu},
      year={2012},
      eprint={1205.2629},
      archivePrefix={arXiv},
      primaryClass={cs.LG},
      url={https://arxiv.org/abs/1205.2629}, 
}

@misc{du2024reducereuserecyclecompositional,
      title={Reduce, Reuse, Recycle: Compositional Generation with Energy-Based Diffusion Models and MCMC}, 
      author={Yilun Du and Conor Durkan and Robin Strudel and Joshua B. Tenenbaum and Sander Dieleman and Rob Fergus and Jascha Sohl-Dickstein and Arnaud Doucet and Will Grathwohl},
      year={2024},
      eprint={2302.11552},
      archivePrefix={arXiv},
      primaryClass={cs.LG},
      url={https://arxiv.org/abs/2302.11552}, 
}

@misc{domingoenrich2025adjointmatchingfinetuningflow,
      title={Adjoint Matching: Fine-tuning Flow and Diffusion Generative Models with Memoryless Stochastic Optimal Control}, 
      author={Carles Domingo-Enrich and Michal Drozdzal and Brian Karrer and Ricky T. Q. Chen},
      year={2025},
      eprint={2409.08861},
      archivePrefix={arXiv},
      primaryClass={cs.LG},
      url={https://arxiv.org/abs/2409.08861}, 
}

@misc{fan2023dpokreinforcementlearningfinetuning,
      title={{DPOK}: {R}einforcement Learning for Fine-tuning Text-to-Image Diffusion Models}, 
      author={Ying Fan and Olivia Watkins and Yuqing Du and Hao Liu and Moonkyung Ryu and Craig Boutilier and Pieter Abbeel and Mohammad Ghavamzadeh and Kangwook Lee and Kimin Lee},
      year={2023},
      eprint={2305.16381},
      archivePrefix={arXiv},
      primaryClass={cs.LG},
      url={https://arxiv.org/abs/2305.16381}
}

@article{aesthetic,
  title={Laion-5b: An open large-scale dataset for training next generation image-text models},
  author={Schuhmann, Christoph and Beaumont, Romain and Vencu, Richard and Gordon, Cade and Wightman, Ross and Cherti, Mehdi and Coombes, Theo and Katta, Aarush and Mullis, Clayton and Wortsman, Mitchell and others},
  journal={Advances in neural information processing systems},
  volume={35},
  pages={25278--25294},
  year={2022}
}

@article{hps,
  title={Better aligning text-to-image models with human preference},
  author={Wu, Xiaoshi and Sun, Keqiang and Zhu, Feng and Zhao, Rui and Li, Hongsheng},
  journal={arXiv preprint arXiv:2303.14420},
  volume={1},
  number={3},
  year={2023}
}

@article{pickscore,
  title={Pick-a-pic: An open dataset of user preferences for text-to-image generation},
  author={Kirstain, Yuval and Polyak, Adam and Singer, Uriel and Matiana, Shahbuland and Penna, Joe and Levy, Omer},
  journal={Advances in Neural Information Processing Systems},
  volume={36},
  pages={36652--36663},
  year={2023}
}

@article{imagereward,
  title={Imagereward: Learning and evaluating human preferences for text-to-image generation},
  author={Xu, Jiazheng and Liu, Xiao and Wu, Yuchen and Tong, Yuxuan and Li, Qinkai and Ding, Ming and Tang, Jie and Dong, Yuxiao},
  journal={Advances in Neural Information Processing Systems},
  volume={36},
  pages={15903--15935},
  year={2023}
}

@inproceedings{mps,
  title={Learning multi-dimensional human preference for text-to-image generation},
  author={Zhang, Sixian and Wang, Bohan and Wu, Junqiang and Li, Yan and Gao, Tingting and Zhang, Di and Wang, Zhongyuan},
  booktitle={Proceedings of the IEEE/CVF Conference on Computer Vision and Pattern Recognition},
  pages={8018--8027},
  year={2024}
}

@article{hu2022lora,
  title={Lora: Low-rank adaptation of large language models.},
  author={Hu, Edward J and Shen, Yelong and Wallis, Phillip and Allen-Zhu, Zeyuan and Li, Yuanzhi and Wang, Shean and Wang, Lu and Chen, Weizhu and others},
  journal={ICLR},
  volume={1},
  number={2},
  pages={3},
  year={2022}
}

@inproceedings{sohrabi2024pi,
  title={On PI Controllers for Updating Lagrange Multipliers in Constrained Optimization},
  author={Sohrabi, Motahareh and Ramirez, Juan and Zhang, Tianyue H and Lacoste-Julien, Simon and Gallego-Posada, Jose},
  booktitle={International Conference on Machine Learning},
  pages={45922--45954},
  year={2024},
  organization={PMLR}
}
